\documentclass[%
preprint,
superscriptaddress,
 amsmath,amssymb,
aip,jcp
]{revtex4-1}

\usepackage{graphicx}% Include figure files
\usepackage{dcolumn}% Align table columns on decimal point
\usepackage{bm}% bold math
\usepackage{xcolor}
\usepackage{hyperref}
\usepackage{subfigure}

\usepackage{multirow}
\usepackage{rotating}
\usepackage{colortbl}

\usepackage{tikz}
\usetikzlibrary{shapes,arrows,
positioning, % e.g. below left= of node
calc, % for coordinate positioning
bayesnet % for graphical models
}

\usepackage{algorithm}%,algcompatible}
\usepackage{algcompatible}

\usepackage{siunitx}

\begin{document}

%\preprint{APS/123-QED}

\title{Predictive Collective Variable Discovery with Deep Bayesian Models}

\author{Markus Sch\"oberl}
\email{mschoeberl@gmail.com}
\affiliation{Center for Informatics and Computational Science, University of Notre Dame, 311 Cushing Hall, Notre Dame, IN 46556, USA.}
\affiliation{Continuum Mechanics Group,
Technical University of Munich, Boltzmannstra{\ss}e 15, 85748 Garching, Germany.}
%\affiliation{Scholar of Hanns-Seidel-Foundation, Lazarettstra{\ss}e 33, 80636 Munich, Germany.}

\author{Nicholas Zabaras}
\homepage{https://cics.nd.edu/}
\email{nzabaras@gmail.com}
\affiliation{Center for Informatics and Computational Science, University of Notre Dame, 311 Cushing Hall, Notre Dame, IN 46556, USA.}

\author{Phaedon-Stelios Koutsourelakis}%
\homepage{http://www.contmech.mw.tum.de}
\email{p.s.koutsourelakis@tum.de}
\affiliation{Continuum Mechanics Group,
Technical University of Munich, Boltzmannstra{\ss}e 15, 85748 Garching, Germany.}

\newcommand\be{\begin{equation}}
\newcommand\ee{\end{equation}}

\newcommand{\refeqq}[1]{Eq.~(\ref{#1})}
\newcommand{\reffig}[1]{Fig.~\ref{#1}}

\newcommand\bTheta{\boldsymbol{\Theta}}
\newcommand\btheta{\boldsymbol{\theta}}
\newcommand\boeta{\boldsymbol{\eta}}
\newcommand\bphi{\boldsymbol{\phi}}
\newcommand\beps{\boldsymbol{\epsilon}}
\newcommand\bSig{\boldsymbol{\Sigma}}
\newcommand\bsig{\boldsymbol{\sigma}}
\newcommand\bz{\boldsymbol{z}}
\newcommand\bS{\boldsymbol{S}}
\newcommand\bK{\boldsymbol{K}}
\newcommand\bA{\boldsymbol{A}}
\newcommand\bB{\boldsymbol{B}}
\newcommand\bC{\boldsymbol{C}}
\newcommand\bW{\boldsymbol{W}}
\newcommand\bX{\boldsymbol{X}}
\newcommand\bx{\boldsymbol{x}}
\newcommand\bXi{\boldsymbol{X}^{(i)}}
\newcommand\bxi{\boldsymbol{x}^{(i)}}
\newcommand\bzi{\boldsymbol{z}^{(i)}}
\newcommand\bw{\boldsymbol{w}}
\newcommand\bmu{\boldsymbol{\mu}}
\newcommand\bmui{\boldsymbol{\mu}^{(i)}}
\newcommand\diag{\textbf{diag}}

\newcommand\cf{_{\text{cf}}}
\newcommand\cg{_{\text{c}}}
\newcommand\indf{_{\text{f}}}
\newcommand\inth{_{\btheta}}
\newcommand\inphi{_{\bphi}}
\newcommand\indcv{_{\text{CV}}}
\newcommand{\indtarg}{_{\text{target}}}
%%% ALGORITHM
\algnewcommand\algorithmicinput{\textbf{Input}}
\algnewcommand\INPUT{\item[\algorithmicinput]}
\algnewcommand\algorithmicinit{\textbf{Initialize}}
\algnewcommand\Init{\item[\algorithmicinit]}
\algnewcommand\algorithmicreturn{\textbf{Return}}
\algnewcommand\Return{\item[\algorithmicreturn]}

%\author{Markus Sch\"oberl}
%\affiliation{
% Center for Informatics and Computational Science\\
%\affiliation{
%\author{Nicholas Zabaras}
%\affiliation{%

%\author{Phaedon-Stelios Koutsourelakis}
%\affiliation{%


\date{\today}% It is always \today, today,

% --------------------  ABSTRACT -------------------------
\begin{abstract}

\vspace{1cm}
\emph{
\noindent
Preprint - accepted and published in The Journal of Chemical Physics 2019 150:2; \url{https://doi.org/10.1063/1.5058063}.
} \vspace{1cm} \\
	Extending spatio-temporal scale limitations of models for complex atomistic systems considered in biochemistry and materials science necessitates the development of enhanced sampling methods. The potential acceleration in exploring the configurational space by enhanced sampling methods depends on the choice of collective variables (CVs). In this work, we formulate the discovery of CVs as a Bayesian inference problem and consider the CVs as hidden generators of the full-atomistic trajectory. The ability to generate samples of the fine-scale atomistic configurations using limited training data allows us to compute estimates of observables as well as our probabilistic confidence on them. The methodology is based on emerging methodological advances in machine learning and variational inference. 
	The discovered CVs are related to physicochemical properties which are essential for understanding mechanisms especially in unexplored complex systems. We provide a quantitative assessment of the CVs in terms of their predictive ability for alanine dipeptide (ALA-2) and ALA-15 peptide.
\end{abstract}
% --------------------------------------------------------

\maketitle

% --------------------- INTRODUCTION ---------------------

\section{Introduction}
\label{sec:introduction}
Molecular dynamics (MD) simulations, in combination with prevalent algorithmic enhancements and tremendous progress in computational resources, have contributed to new insights into mechanisms and processes present in physics, chemistry, biology and engineering. 
However, their applicability in systems of practical relevance poses  insurmountable computational difficulties~\cite{perilla2015, koutsourelakis2016}. For example, the simulation of $M=10^5$ atoms over a time horizon of a mere $T\approx 10^{-4}~\textnormal{s}$ with a time step of $\Delta t=10^{-15}~\textnormal{s}$ implies a computational time of one year~\cite{barducci2011}. A rugged free-energy surface and configurations separated by high free-energy barriers lead to unobserved conformations even in very long simulations. 

Enhanced sampling methods~\cite{pietrucci2011} provide a framework for accelerating the exploration of the configurational space~\cite{ferguson2011, zheng2013b, valsson2014, chen2017, chen2018, mitsutake2013, bierig2016}. Those methods rely on the existence of a lower-dimensional representation of the atomistic detail.
Lower-dimensional system variables (reaction coordinates), capture the characteristics of the system, allow us to understand relevant processes and conformational changes~\cite{luque2012}, and can enable guided and enhanced MD simulations. \emph{Reaction coordinates} provide quantitative understanding of macromolecular motion, whereas \emph{order parameters} are of qualitative nature as discussed in~[\onlinecite{rohrdanzclementi2013}]. In the following, we use the term \emph{collective variables} (CVs), combining the quantitative and qualitative properties of reaction coordinates and order parameters, respectively. Refs.~[\onlinecite{pietrucci2011, rohrdanzclementi2013}] review the challenges in the exploration of the free-energy landscape and the identification of ``good'' collective variables. 

Adding an appropriate biasing potential or force, based on CVs, results into an accelerated exploration of the configurational space~\cite{rohrdanzclementi2013}.
Such algorithms might employ a constant bias term (e.g. umbrella sampling~\cite{torrie1977}, hyperdynamics~\cite{voter1997}, accelerated MD~\cite{hamelberg2004}, etc.) or a time-dependent one (e.g. local elevation~\cite{huber1994}, conformational flooding~\cite{grubmueller1995}, metadynamics~\cite{laio2002, barducci2008, barducci2011}, adaptive biasing force~\cite{darve2008, jerome2010}, etc.).
The crucial ingredient for almost all of the aforementioned algorithms is the \emph{right choice} of the \emph{collective variables}. The potential benefit and justification of enhanced sampling algorithms strongly depend on the quality of the collective variables as comprehensively elaborated in~[\onlinecite{pietrucci2017,pan2014,fu2017}]. Physical intuition, experience gathered from previous simulation as well as quantitative methods for dimensionality reduction (e.g. by utilizing principal component analysis~\cite{hotelling1933} (PCA)), potentially support the choice of reasonable collective variables. For complex materials-design problems and large-scale biochemical processes, complexity exceeds our intuition and the question of ``good'' collective variables remains unanswered. Enhanced sampling methods employing inappropriate collective variables can be outperformed by brute force MD simulations~\cite{pande2017}. Thus, the identification of collective variables or reaction coordinates poses an important and difficult problem. 

A systematic, robust, and general approach is needed for the discovery of lower-dimensional representations.
Recent developments in dimensionality reduction methods provide a systematic strategy for discovering CVs~\cite{rohrdanzclementi2013}. For completeness, we give a brief overview of significant tools addressing CV discovery and dimensionality reduction in the context of molecular systems. An early study~\cite{amadei1993} found a steep decay in the eigenvalues of peptide trajectories indicating the existence of a low-dimensional representation that is capable of capturing essential physics. This study is based on PCA~\cite{pearson1901,hotelling1933} which identifies a linear coordinate transformation for best capturing the variance. However, the linear coordinate transformations employed merely describe local fluctuations in the context of peptide trajectories. Multidimensional scaling (MDS)~\cite{troyer1995, haerdle2007} identifies a lower-dimensional embedding such that pairwise distances (e.g. root-mean-square deviation (RMSD)) between atomistic configurations are best preserved. Sketch-map~\cite{ceriotti2011} focuses on preserving ``middle'' ranged RMSD between trajectory pairs. Middle ranged RMSD pairs are the most relevant for observing pertinent behavior of the system~\cite{ceriotti2011}.
Isometric feature map or ISOMAP~\cite{tenenbaum2000} follows a similar idea of preserving geodesic distances.
The aforementioned methods require dense sampling and encounter problems if the training data is non-uniformly distributed~\cite{rohrdanz2011, balasubramanian2002, donoho2003}. Furthermore, we note that those methods involve a mapping from the atomistic configurations to the CVs whereas  predictive tasks require a generative mapping from the CVs to the atomistic configuration.

Another group of non-linear dimensionality reduction methods follows the idea of approximating the eigenfunctions of the backward Fokker-Plank operator~\cite{risken1996} by identifying eigenvalues and eigenvectors of transition kernels. The employed kernels resemble transition probabilities between configurations that we aim to preserve.
For example, the diffusion map~\cite{coifman2005, coifman2005b, ferguson2011b} retains the diffusion distance by the identified coordinates for dynamic~\cite{nadler2006} and stochastic systems~\cite{nadler2008}.
A variation of diffusion maps exploits locally scaled diffusion maps (LSDMap)~\cite{rohrdanz2011} which calculate the transition probabilities between two configurations, utilizing the RMSD instead of an Euclidean distance. An additional local scale parameter, indicating the distance around a specific configuration presumably could be well approximated by a low-dimensional hyperplane tangent. LSDMap is applied in~[\onlinecite{rohrdanz2014}] and enhances the exploration of the configurational space as shown in~[\onlinecite{zheng2013}].
More recent approaches to collective variable discovery work under a common variational approach for conformation dynamics (VAC)~\cite{noe2013} and employ a combination of basis functions for defining the eigenfunctions to the backward Fokker-Planck operator.
One approach under VAC was developed in the context of metadynamics~\cite{laio2002} combining ideas from time-lagged independent component analysis and well-tempered metadynamics~\cite{mccarty2017}. Further developments have focused  on alternate distance metrics, relying either on a kinetic distance which measures  how slowly configurations interconvert~\cite{noe2015}, or on the commute distance~\cite{noe2016} which provides an extension (arising by integration) of the former.

Several  methods rely on the estimation of the eigenvectors of transitions matrices which is an expensive task in terms of computational cost. The need for ``large'' training datasets (e.g. $\num{10000}$ datapoints are required for robustness of the results ~\cite{rohrdanzclementi2013}) limits the applicability of these methods to less complex systems. We refer to~[\onlinecite{duan2013}] for a critical review and comparison of the various methodologies mentioned before.

In this work, we propose a data-driven reformulation of the identification of CVs under the paradigm of probabilistic (Bayesian) inference.
The methodology implies a generative model, considering CVs as lower-dimensional (latent) generators~\cite{jordan1999} of the full atomistic trajectory.
The focus furthermore is on problems where limited atomistic training data are available that prohibit the accurate calculation of statistics for quantities of interest. Our approach is to compute an approximation of the underlying probabilistic distribution of the data. We then use this approximate distribution in a generative manner to perform accurate Monte Carlo estimation of the quantities of interest. To account for the limited information provided by small size training datasets, epistemic uncertainties on quantities of interest are also computed within the Bayesian paradigm. 

In the context of coarse-graining atomistic systems, latent variable models have been introduced in~[\onlinecite{schoeberl2017, felsberger_physics-constrained_2018}].
We optimize a flexible non-linear mapping between CVs and atomistic coordinates which implicitly specifies the meaning of the CVs. The identified CVs provide physical/chemical insight into the characteristics of the considered system. In the proposed model, the posterior distribution of the CVs for a given   atomistic data point is computed. This posterior provides a pre-image of the atomistic representation in the lower-dimensional latent space.
We utilize recent developments in machine learning and deep Bayesian modeling (Auto-Encoding Variational Bayes\cite{kingma2014, rezendre2014}). While typically  deep learning models rely on huge amounts on data, we demonstrate the robustness of the proposed methodology considering only small and highly-variable datasets (e.g. $50$ data points compared to $\num{10000}$ as required in the aforementioned methods). The proposed strategy requires significantly less data as compared to MDS~\cite{troyer1995, haerdle2007}, ISOMAP~\cite{tenenbaum2000}, and diffusion map~\cite{coifman2005, coifman2005b, nadler2006} and  simultaneously enables the quantification of uncertainties arising from limited data.
We also discuss how additional datapoints can be readily incorporated by efficiently updating the previously trained model.

Apart from  the possibility of utilizing the discovered CVs for dimensionality reduction and enhanced sampling, we exploit them for \emph{predictive} purposes i.e. for generating new atomistic configurations and estimating macroscopic observables.
One could draw similarities between the identification of CVs and the
problem of identifying a good coarse-grained representation~\cite{kmiecik2016, noid2007, shell2008, kremer2009, trashorras2010, kalligiannaki2012, harmandaris2016, bilionis2013, dama2013, noid2013, foley2015, schoeberl2017, langenberg2018}.
In addition, rather than solely obtaining point estimates of observables,  the Bayesian framework adopted provides whole distributions which capture the epistemic uncertainty. This uncertainty propagates in the form of error bars around the predicted observables.

Several recent publications focus on similar problems~\cite{hernandez2017, noe2018, mohamed2018}. The present work clearly differs from~[\onlinecite{mohamed2018}] where the data is provided in a pre-processed form of sine and cosine backbone dihedral angles, i.e. not as the full-atom configurations. 
The approach in~[\onlinecite{noe2018}] utilizes a pre-reduced representation of heavy atom positions as training data. While this is valid,  it necessitates physical insight which might be not available for unexplored complex chemical compounds.
In contrast, we rely on training data represented as Cartesian coordinates comprising all atoms of the considered system. We do not consider any physically- or chemically-motivated transformation nor do we perform any preprocessing of the dataset.
Instead, we reveal, given the dimensionality of the CVs, important characteristics (i.e. dihedral angles, heavy atom positions) or less relevant fluctuations (noise) from the full atomistic picture.
This work is also distinguished by following throughout a formalism based on Bayesian Learning. Instead of adopting or designing optimization objectives or loss functions, we consistently work within a Bayesian framework where the objective naturally arises. Furthermore, this readily allows us to make use of sparsisty-inducing priors which reveal parsimonious features.
The work of~[\onlinecite{chen2017}] is based on auto-associative artificial neural networks (autoencoders) which allow the encoding and reconstruction of atomistic configurations given an input datum. 
Ref.~[\onlinecite{chen2017}] relies on reduced Cartesian coordinates in the form of backbone atoms which induces information loss. In addition, the focus in~[\onlinecite{chen2017}] is on CV discovery and enhanced sampling whereas we focus on CV discovery and obtaining a predictive model accounting for epistemic uncertainty.

The structure of the rest of the paper is as follows. Section~\ref{sec:methodology} presents the basic model components, the use of  Variational Autoencoders ~(VAEs\cite{kingma2014}) in the CV discovery, and provides details on the learning algorithms employed. Numerical evidence of the capabilities of the proposed framework is provided in Section~\ref{sec:numericalillustrations}. We identify CVs for alanine dipeptide and show the correlation between the discovered CVs and the dihedral angles. We furthermore assess the predictive quality of the discovered CVs and estimate observables augmented by credible intervals. We show the dependence of credible intervals on the  amount of training data. We  also present the results of  a similar analysis for a more complex and higher-dimensional  molecule, i.e. the  ALA-15 peptide.  Finally, Section~\ref{sec:conclusions} summarizes the key findings of this paper and provides a brief discussion on potential extensions.

% ------------------ METHODOLOGY -------------------------

\section{Methods}
\label{sec:methodology}

After introducing the main notational convention in the context of equilibrium statistical mechanics, this section is devoted to the key concepts of generative latent variable models and variational inference~\cite{beal2003} with emphasis on the identification of collective variables in atomistic systems.

\subsection{Equilibrium statistical mechanics}
\label{sec:esm}
We denote the coordinates of atoms of a molecular ensemble as $\bx \in \mathcal M\indf \subset \mathbb R^{n\indf}$, with $n\indf = \dim(\bx)$. The coordinates $\bx$ follow the Boltzmann-Gibbs density,
\be
\label{eqn:pf}
p_{\text{target}}(\bx) = \frac{1}{Z(\beta)} e^{-\beta U(\bx)},
\ee
with the interatomic potential $U(\bx)$, $\beta = \frac{1}{k_b T}$ where $k_b$ is the Boltzmann constant and $T$ the temperature. The normalization constant is given as $Z(\beta) = \int_{\mathcal M\indf} \exp \{ -\beta U(\bx) \}~d\bx$.
MD simulations~\cite{adler1959}, or Monte-Carlo-based methods~\cite{binder2010} allow us to obtain samples from the distribution defined in~\refeqq{eqn:pf}. In the following, we assume that a dataset, $\bX = \{\bxi\}_{i=1}^N$, has been collected, where  $\bxi \sim p_{\text{target}}(\bx)$. $N$ denotes the amount of data points considered. The dataset $\bX$ will be used for training the generative model to be introduced in the sequel. The underlying assumption in this work is that the size of the available training dataset $\bX$ is small and not sufficient to compute directly statistics of observables. Our focus is thus on deriving an approximation to the distribution in~\refeqq{eqn:pf} from which, in a computationally inexpensive manner, one can sample sufficient realizations of $\bx$ to allow probabilistic estimates of observables.  

As elaborated in~[\onlinecite{rohrdanzclementi2013}], the collection of a dataset $\bX$ that sufficiently captures the configurational space constitutes a difficult problem of its own. Hampered by free-energy barriers, a MD simulation is not guaranteed to visit all conformations of an atomistic system within a finite simulation time. The discovery of CVs can facilitate the development of  enhanced sampling methods~\cite{pietrucci2017,laio2002, barducci2011} to address the efficient exploration of the configurational space.

This study considers systems in equilibrium for a given constant temperature $T$ and consequently constant $\beta$. Optimally, the CVs discovered should be  suitable for a range of temperatures~\cite{fu2017}.  

\subsection{Probabilistic generative models}
\label{sec:pgm}

Deep learning~\cite{lecun2015} integrated with probabilistic modeling~\cite{ghahramani2015} has impacted many research areas~\cite{linden2014}. In this paper, we emphasize a subset of these models referred to as  \emph{probabilistic generative models}~\cite{jordan1999, ng2002}.

The objective is to identify CVs associated with relevant configurational changes of the system of interest. We consider CVs as hidden (low-dimensional) generators, giving rise to the observed atomistic configurations $\bx$~\cite{mackay2003}.
Extending the variable space of atomistic coordinates $\bx$ by latent CVs denoted as  $\bz \in \mathcal M\indcv \subset \mathbb R^{n\indcv}$, with $n\indcv = \dim(\bz)$ and $\dim(\bz) \ll \dim(\bx)$, allows us to define a joint distribution over the observed data $\bx$ and latent CVs~\cite{jordan1999,bishop1999} $ p(\bx,\bz)$.
The joint distribution $ p(\bx,\bz)$ is written as,
\be
\label{eqn:joint}
p(\bx,\bz) = p(\bx|\bz)~p(\bz).
\ee
In~\refeqq{eqn:joint}, $p(\bz)$ prescribes the distribution of the CVs and $p(\bx|\bz)$ represents the conditional probability of the full atomistic coordinates $\bx$ given their latent representation $\bz$. The probabilistic connection between the latent CVs $\bz$ and the atomistic representation $\bx$ implicitly defines the meaning of the CVs.

Marginalizing the joint representation of \refeqq{eqn:joint} with respect to the CVs leads to  $ p(\bx)$,
\be
\label{eqn:predictive}
 p(\bx) = \int_{\mathcal M\indcv} p(\bx,\bz)~d\bz =  \int_{\mathcal M\indcv}  p(\bx|\bz)~p(\bz)~d\bz.
\ee
Equation~(\ref{eqn:predictive}) provides a generative model for the atomistic configurations $\bx$
and will be utilized as an efficient estimator for observables of the atomistic system.
Standard autoencoders in the context of CV discovery~\cite{chen2017} do not yield a probabilistic, predictive model which is the focus of this work.
With appropriate selection of  $p(\bz)$ and $p(\bx|\bz)$, the resulting predictive distribution $p(\bx)$ should resemble the atomistic reference $p_{\text{target}}(\bx)$ in~\refeqq{eqn:pf}. In order to quantify the closeness of the approximating distribution $p(\bx)$ and the actual distribution $p_{\text{target}}(\bx)$, a distance measure is employed. The KL-divergence is one possibility out of the family of $\alpha$-divergences\cite{adrzej2010, cha2017}\footnote{Inference on the generalized $\alpha$-divergence is addressed in Ref.~[\onlinecite{lobato2016}].} measuring the similarity between ${{red}p_{\text{target}}}(\bx)$ and $p(\bx)$. The non-negative  valued KL-divergence is zero if and only if the two distributions coincide, which leads to the minimization objective with respect to $p(\bx)$ of the following form:
\begin{align}
	\label{eqn:kl}
	D_{KL}(p_{\text{target}}(\bx) ||  p(\bx)) = &-\int_{\mathcal M\indf} p_{\text{target}}(\bx) \log \frac{ p(\bx)}{p_{\text{target}}(\bx)}~d\bx \nonumber \\
	= &-\int_{\mathcal M\indf} p_{\text{target}}(\bx) \log ~  p(\bx)~d\bx \nonumber \\
	&+\int_{\mathcal M\indf} p_{\text{target}}(\bx) \log ~ p_{\text{target}}(\bx)~d\bx.
\end{align}
We introduce a parametrization $\btheta$ of the approximating distribution as $p(\bx | \btheta) = \int_{\mathcal M\indcv} p\inth(\bx|\bz) p\inth(\bz)~d\bz$.
Instead of minimizing the KL-divergence with respect to $p(\bx)$, one can optimize the objective with respect to the parameters $\btheta$. We note that the minimization of \refeqq{eqn:kl} is equivalent to maximizing the expression $\int_{\mathcal M\indf} p_{\text{target}}(\bx) \log ~  p(\bx)~d\bx$. If we consider a data-driven approach where $p_{\text{target}}(\bx)$ is approximated by a finite-sized dataset $\bX$,  we can write the problem as the maximization of the marginal log-likelihood $\log  p\inth(\bx^{(i)},\cdots
, \bx^{(N)})$:
\begin{align}
	\label{eqn:loglikelihood}
	\log  p(\bX| \btheta) = &\sum_{i=1}^N \log p(\bxi | \btheta) \nonumber \\
	= &\sum_{i=1}^N \log \left( \int_{\mathcal M\indcv} p\inth(\bxi|\bzi) ~ p\inth(\bzi) ~d\bzi \right).
\end{align}
Maximizing~\refeqq{eqn:loglikelihood} with respect to the model parameters $\btheta$ results into the maximum likelihood estimate (MLE) $\btheta_{\text{MLE}}$. By introducing a prior $p(\btheta)$ on the parameters, one can augment this optimization problem to compute the Maximum a Posteriori (MAP) estimate~\cite{mackay1992, gelman1995, jaynes2005} of $\btheta$ as follows:
\be
\label{eqn:map_estimate}
\arg \max_{\btheta} \left\{  \log  p(\bX| \btheta) + \log p(\btheta) \right\}.
\ee
The full posterior of the model parameters $\btheta$ could also be obtained by applying Bayes' rule,
\be
p(\btheta | \bX) = \frac{ p(\bX| \btheta) p(\btheta) }{p(\bX)}.
\ee
Quantifying uncertainties in $\btheta$ enables us to capture the epistemic uncertainty introduced from the limited training data.
The discovery of CVs through Bayesian inference is elaborated in the sequel.

\subsection{Inference and learning}
\label{sec:inference}

This section focuses on the details of inference and parameter learning for the generative model introduced in~\refeqq{eqn:predictive}. Both tasks are facilitated by approximate variational inference~\cite{hoffmann2013} and stochastic backpropagation~\cite{ranganath2014,rezendre2014,paisley2012} which we discuss next.

Direct optimization of the marginal likelihood $ p(\bx | \btheta)$ requires the evaluation of  $p(\bx | \btheta) = \int_{\mathcal M\indcv} p\inth (\bx|\bz) p\inth(\bz)~d\bz$ which constitutes an intractable integration over ${\mathcal M\indcv}$. The  posterior over the latent CVs,  $p\inth(\bz|\bx) = p\inth(\bx|\bz) p\inth(\bz) /  p(\bx | \btheta)$,  is also computationally intractable. Therefore, direct application of  Expectation-Maximization~\cite{dempster1977, neal1999} is not feasible.
To that end, we reformulate the marginal log-likelihood for the dataset $\bX = \{\bxi \}_{i=1}^N$ by introducing auxiliary densities $q\inphi(\bzi|\bxi)$  parametrized by $\bphi$.
The meaning of $q\inphi(\bzi|\bxi)$ will be specified later in the text.
The marginal log-likelihood follows,
\begin{align}
	\label{eqn:lowerbound}
	\log \ p(\bX| \btheta) &= \sum_{i=1}^N \log p(\bxi | \btheta) \nonumber \\
	&= \sum_{i=1}^N \log \int_{\mathcal M\indcv} p\inth(\bxi|\bzi) p\inth(\bzi)~d\bzi \nonumber \\
	&= \sum_{i=1}^N \log \int_{\mathcal M\indcv} q\inphi(\bzi|\bxi) \frac{p\inth(\bxi|\bzi)  p\inth(\bzi)}{q\inphi(\bzi|\bxi)}~d\bzi  \nonumber \\
	%&= \sum_{i=1}^N \log \int_{\mathcal M\indcv} q\inphi(\bzi|\bxi) \frac{p\inth(\bxi|\bzi)  p\inth(\bzi)}{q\inphi(\bzi|\bxi)}~d\bzi  \nonumber \\
	&\geq \sum_{i=1}^N \underbrace{\int_{\mathcal M\indcv}  q\inphi(\bzi|\bxi) \log \frac{p\inth(\bxi|\bzi)  p\inth(\bzi)}{q\inphi(\bzi|\bxi)}~d\bzi }_{\mathcal L (\btheta,\bphi; \bxi)}, %\nonumber \\
\end{align}
where in the last step we have made use of Jensen's inequality. Note that for each data point $\bxi$, one latent CV $\bzi$ is assigned.
 The lower-bound of  the marginal log-likelihood is:
\be
\mathcal L (\btheta,\bphi;\bX) = \sum_{i=1}^N \mathcal L (\btheta,\bphi;\bxi),
\ee
and implicitly depends on $\bphi$  through the parametrization of $q\inphi(\bz|\bx)$.
For each data point $\bxi$ and from  the definition of $\mathcal L (\btheta, \bphi; \bxi)$, one can rewrite the marginal log-likelihood $\log  p(\bxi | \btheta)$ as,
\begin{align}
	\log  p (\bxi | \btheta) = D_{KL}\left(q\inphi(\bzi|\bxi) || p\inth(\bzi|\bxi) \right) + \mathcal L (\btheta, \bphi; \bxi) \ge \mathcal L (\btheta, \bphi; \bxi).
	\label{eqn:decomposedloglikeli}
\end{align}
Since the  KL-divergence is always non-negative, the inequalities in~\refeqq{eqn:lowerbound} and~\refeqq{eqn:decomposedloglikeli}  become  equalities  if and only if $q\inth(\bzi|\bxi) = p\inth(\bzi|\bxi)$  as in this case   $D_{KL}\left(q\inphi(\bzi|\bxi) || p\inth(\bzi|\bxi)\right ) = 0$.
Thus $q\inphi(\bzi|\bxi)$ can be thought of as  an approximation of the true  posterior over the latent variables $\bz$. If the lower-bound gets tight, $q\inphi(\bzi|\bxi)$ equals the exact posterior $p\inth(\bz|\bxi)$.

Equation~(\ref{eqn:lowerbound}) can also be written as follows,
\begin{align}
	\mathcal L(\btheta, \bphi; \bX) = &\sum_{i=1}^N \mathbb E_{q\inphi(\bzi|\bxi)} [ - \log q\inphi(\bzi|\bxi) \nonumber + \log p\inth(\bxi,\bzi)] \nonumber \\
	= &-\sum_{i=1}^N D_{KL}\left(q\inphi(\bzi|\bxi) || p\inth(\bzi)\right)  + \sum_{i=1}^N \mathbb E_{q\inphi(\bzi|\bxi)} [\log p\inth(\bxi|\bzi) ].
	\label{eqn:decomposedlowerbound}
\end{align}
It is clear from~\refeqq{eqn:decomposedlowerbound} that the lower-bound balances the optimization of the following two objectives~\cite{kingma2014}:
\begin{enumerate}
	\item Minimizing $\sum_{i=1}^N D_{KL}\left(q\inphi(\bzi|\bxi) || p\inth(\bz)\right)$ regularizes the approximate posterior $q\inphi(\bzi|\bxi)$ such that, \emph{on average} over all data points $\bxi$, it  resembles  $p\inth(\bz)$.
	We expect highly probable atomistic configurations $\bxi$  to be encoded to CVs $\bzi$ located in regions with high probability mass in $p\inth(\bz)$. The approximate posterior $q\inphi(\bzi|\bxi)$ over the latent CVs $\bz$ accounts for this and supports findings presented in~[\onlinecite{noe2018}].
	\item $\mathbb E_{q\inphi(\bzi|\bxi)} [\log p\inth(\bxi|\bzi) ]$ is the negative expected reconstruction error employing the encoded pre-image of the atomistic configuration $\bxi$ in the latent CV space.
	For example assuming $p\inth(\bxi|\bzi)$ to be a Gaussian with  mean $\bmu(\bzi)$ and variance $\bsig^2$, one can rewrite $\mathbb E_{q\inphi(\bzi|\bxi)} [\log p\inth(\bxi|\bzi) ]$ as,
	\begin{align}
		\mathbb E_{q\inphi(\bzi|\bxi)} [\log p\inth(\bxi|\bzi) ] &= \mathbb E_{q\inphi(\bzi|\bxi)} \left[ -\frac{1}{2} \frac{\left(\bxi - \bmu(\bzi)\right)^2}{\bsig^2} \right] + \text{const.} \nonumber \\
		&\propto - \mathbb E_{q\inphi(\bzi|\bxi)} \left[ \left( \bxi - \bmu(\bzi)\right)^2 \right] \nonumber \\
		&= - \int_{\mathcal M\indcv} q\inphi(\bzi|\bxi) ~ \left( \bxi - \bmu(\bzi)\right)^2  ~d\bzi .
		\label{eqn:rec_error}
	\end{align}
	The second line of~\refeqq{eqn:rec_error} is the negative expected error of reconstructing the atomistic configuration $\bxi$ through the decoder $p\inth(\bxi|\bzi)$. The expectation (see last line in~\refeqq{eqn:rec_error}) is evaluated with respect to $q\inphi(\bzi|\bxi)$ and therefore with respect to all CVs $\bzi$ probabilistically assigned to $\bxi$.
\end{enumerate}

The approximate posterior $q\inphi$ of the latent variables $\bz$ serves as a recognition model and is called the encoder~\cite{kingma2014}. Atomistic configurations $\bx$ can be mapped via $q\inphi(\bz|\bx)$ to their lower-dimensional representation $\bz$ in the CV space. Hence, each $\bz$ could be interpreted as a (latent) \emph{encoding} of an $\bx$.
Its counterpart, the decoder   $p\inth(\bx|\bz)$, probabilistically maps CVs $\bz$  to atomistic configurations $\bx$. As it will be demonstrated in the sequel, $\bz$ sampled from $p\inth(\bz)$ will be used to  reconstruct  atomistic configurations via  $p\inth(\bx|\bz)$. The corresponding graphical model is presented in~\reffig{fig:pgm_vae}.
Note that we do not require any physicochemical meaning assigned to the latent CVs that are identified implicitly during the training process.
\begin{figure}
\centering
\resizebox{0.2\textwidth}{!}{%
      \tikz{ %
        \node[obs] (x) {$\bxi$} ; %
        \node[latent, left=of x] (z) {$\bzi$};
        \node[below=of z] (phi) {$\bphi$};
        \node (zxmiddle) at ($(z)!0.5!(x)$) {};
        \path let \p1 = (zxmiddle), \p2 = (phi) in node (theta) at (\x1,-\y2) {$\btheta$};
        \plate[inner sep=0.25cm, xshift=-0.12cm, yshift=0.12cm] {plate3} {(x) (z)} {$N$}; %
        \edge {z} {x};
        \edge {theta} {z};
        \edge {theta} {x};
        \edge[dashed] {phi} {z};
        \edge[dashed, ->, out=240, in=300] {x} {z}
}
}
\caption{Probabilistic graphical model representation following~[\onlinecite{kingma2014}] with the latent CV representation  $\bzi$ of each configuration $\bxi$ obtained by the approximate variational posterior $q\inphi(\bzi|\bxi)$ using the parametrization $\bphi$. The variational approximation is indicated with dashed edges and the generative model $p\inth(\bx|\bz)p(\bz)$ with solid edges. $\btheta$ is the parametrization of the generative model.}
\label{fig:pgm_vae}
\end{figure}

%%%%%%%%%%%%%%%%%%%%%%%%%%%%%%%
% Uncomment the following lines for using a pdf as input for the graphical model and comment the figure above
%%%%%%%%%%%%%%%%%%%%%%%%%%%%%%%

%\begin{figure}
%    \centering
%    \includegraphics{figures/pgm_vae.pdf}
%\caption{{\color{blue}Probabilistic graphical model representation following~[\onlinecite{kingma2014}] with the latent CV representation  $\bzi$ of each configuration $\bxi$ obtained by the approximate variational posterior $q\inphi(\bzi|\bxi)$ using the parametrization $\bphi$. The variational approximation is indicated with dashed edges and the generative model $p\inth(\bx|\bz)p(\bz)$ with solid edges. $\btheta$ is the parametrization of the generative model.}}
%\label{fig:pgm_vae}
%\end{figure}

The (approximate) inference task of $q\inphi(\bz|\bx)$ has been re-formulated as an optimization problem with respect to the parameters $\bphi$. These will be updated in combination with the parameters  $\btheta$ as described in the following.
At this point, we emphasize that the lower-bound $\mathcal L(\btheta, \bphi; \bxi)$ on the marginal log-likelihood (unobserved CVs are marginalized out) of~\refeqq{eqn:decomposedlowerbound} has been used as a negative ``loss'' function in non-Bayesian applications of  autoencoders in the context of atomistic simulations as in~[\onlinecite{hernandez2017, mohamed2018}].

In order to carry out the optimization $\mathcal L (\bphi, \btheta;\bX)$ with respect to $\{ \bphi, \btheta \}$, first-order derivatives are needed of terms involving expectations with respect to $q\inphi$ as it can be seen in~\refeqq{eqn:decomposedlowerbound}. Consider in general a function $f(\bz)$ and the corresponding expectation $\mathbb E_{q\inphi(\bz| \bx)}[f(\bz)]$. Its gradient with respect to $\bphi$ can be expressed as 
\be
\nabla_{\bphi} \mathbb E_{q\inphi(\bz| \bx)}[f(\bz)] = \mathbb E_{q\inphi(\bz|\bx)}\left[ f(\bz) \nabla_{q\inphi(\bz|\bx)} \log  q\inphi(\bz|\bx) \right],
\label{eqn:aevb_est_highvar}
\ee
and the expectation $\mathbb E_{q\inphi(\bz|\bx)}[\cdot ]$ on the right hand-side can be approximated via a Monte-Carlo (MC) estimate using samples of $\bz$ drawn from $q\inphi(\bz|\bx)$. It is however known~\cite{ranganath2014} that the variance of such estimators can be very high which adversely affects the optimization process.
The high variance of the estimator in~\refeqq{eqn:aevb_est_highvar} can be addressed with the so-called reparametrization trick~\cite{kingma2014, rezendre2014}.
It is based on expressing $\bz$ by auxiliary random variables $\boldsymbol{\epsilon}$ and a differentiable transformation $g\inphi(\boldsymbol{\epsilon}; \bx)$ as
\begin{equation}
\bz = g\inphi(\beps; \bx) ~ \text{with} ~ \beps \sim p(\beps).
\label{eqn:repar_method}
\end{equation}
Using the mapping, $g\inphi : \beps \rightarrow \bz$, we can write the following for the densities $p(\beps)$ and  $q\inphi(\bz|\bx)$:
\be
q\inphi(\bz|\bx) = p\big(g\inphi^{-1}(\bz; \bx)\big) \bigg\lvert \frac{\partial g\inphi^{-1}(\bz; \bx)}{\partial \bz}\bigg\rvert.
\label{eqn:change_of_variables}
\ee
In \refeqq{eqn:change_of_variables}, $g\inphi^{-1}: \bz \rightarrow \beps$ denotes the inverse function of $g\inphi$ which gives rise to $\beps = g\inphi^{-1}(\bz; \bx)$.
Several such transformations have been documented for typical densities (e.g. Gaussians)~\cite{ruiz2016generalized}.
The change of variables leads to the following expression for the gradient,
\begin{align}
\nabla_{\bphi} \mathbb E_{q\inphi(\bz|\bx)}[f(\bz)] 
%&=  \nabla_{\bphi} \int p(\beps) \bigg\lvert \frac{\partial \beps}{\partial g\inphi(\beps; \bx)} \bigg\rvert f(g\inphi(\beps; \bx)) ~\bigg\lvert \frac{\partial g\inphi(\beps; \bx)}{\partial \beps} \bigg\rvert d\beps \nonumber \\ 
&= \mathbb E_{p(\beps)} [\nabla_{\bphi} f(g\inphi(\beps; \bx)) ] \nonumber \\ 
&= \mathbb E_{p(\beps)} \left[ \frac{ \partial f(g\inphi(\beps; \bx))}{\partial \bz} \frac{\partial g\inphi(\beps; \bx))}{\partial \bphi} \right],
\end{align}
which can in turn be calculated by Monte Carlo using  samples of $\beps$ drawn from $p(\beps)$.
Based on this, we  define the following modified estimator for the lower-bound~\cite{kingma2014},
\begin{align}
	&\tilde{\mathcal L } (\bphi, \btheta;\bxi) = - D_{KL}\left(q\inphi(\bzi|\bxi)||p\inth(\bz)\right) + \frac{1}{L}\sum_{l=1}^L \log p\inth(\bxi|\bz^{(i,l)} )   \nonumber \\
	&\text{with} ~\bz^{(i,l)} = g\inphi(\beps^{(l)}; \bxi) ~\text{and}~ \beps^{(l)} \sim p (\beps).
	\label{eqn:lowerbound_redvar}
\end{align}
Note that for the particular forms of  $q\inphi(\bzi|\bxi)$ and $p\inth(\bz)$ selected in Section~\ref{sec:numill_sim_ala2_model_spec},  $D_{KL}\left(q\inphi(\bzi|\bxi)||p\inth(\bz)\right)$  becomes an analytically tractable expression.
In order  to increase the computational efficiency, we work with a sub-sampled minibatch $\bX^M$ comprising $M$ datapoints from $\bX$, with $M<N$.
This leads to $\lfloor N/M \rfloor$ minibatches, each uniformly sampled from $\bX$.
The corresponding estimator of the lower-bound on the marginal log-likelihood is then given as,
\be
\mathcal L (\bphi, \btheta; \bX) \simeq \tilde{\mathcal L }^M(\btheta, \bphi; \bX^M) = \frac{N}{M} \sum_{i=1}^M \tilde{ \mathcal L} (\btheta,\bphi;\bxi),
\label{eqn:aevb_lowerbound_minibatch}
\ee
with $\tilde{ \mathcal L} (\btheta,\bphi;\bxi)$ computed  in~\refeqq{eqn:lowerbound_redvar}.
The factor $N/M$ in~\refeqq{eqn:aevb_lowerbound_minibatch} rescales $\sum_{i=1}^M \tilde{ \mathcal L} (\btheta,\bphi;\bxi)$ such that the lower-bound $\tilde{\mathcal L }^M(\btheta, \bphi; \bX^M)$ computed by $M<N$ datapoints  approximates the actual lower-bound $\mathcal L (\bphi, \btheta; \bX)$ computed with $N$ datapoints~\cite{kingma2014}.
However, note that using a subset of the datapoints unavoidably increases the variance in the stochastic gradient estimator~\refeqq{eqn:lowerbound_redvar}. Strategies compensating this increase are presented in~[\onlinecite{zhang2014, nocedal2018}] and a rigorous study of optimization techniques with enhancements in the context of coarse-graining is given in~[\onlinecite{bilionis2013}].
The overall inference procedure is summarized in Algorithm~\ref{alg:aevb}. 

\begin{algorithm}[H]
	\caption{\small Stochastic Variational Inference Algorithm.}
	\begin{algorithmic}%[1]
		%\algsetup{linenosize=\tiny}
		\small
		\STATE $\{\btheta, \bphi \} \leftarrow$ Initialize parameters.
		\REPEAT
		\STATE $\bX^{M} \leftarrow$ Random minibatch of $M$ datapoints drawn from dataset $\bX$.
		\STATE $\beps \leftarrow$ Random sample(s) from noise distribution $p(\beps)$.
		\STATE $\boldsymbol g \leftarrow \nabla_{\bphi, \btheta} \tilde{\mathcal L}^M(\bphi, \btheta; \bX^M)$ Calculate gradients with the estimator in  \refeqq{eqn:aevb_lowerbound_minibatch}.
		\STATE $\{\bphi, \btheta\} \leftarrow$ Update parameters with gradient $\boldsymbol{g}$ (e.g. employing ADAM\cite{kingmaAdam2014}).
		\UNTIL{Convergence of $\{\btheta, \bphi\}$.}
		\State \textbf{return} $\{\btheta, \bphi \}$.
	\end{algorithmic}
	\label{alg:aevb}
\end{algorithm}

We finally  note that new data  can be readily incorporated by augmenting accordingly the objective and initializing the algorithm with the optimal parameter values found up to that point. In fact this strategy was adopted in the results presented in the Section~\ref{sec:numericalillustrations} and led to significant efficiency gains. One can envision running an all-atom simulation which sequentially generates new training data that are automatically and quickly  ingested by the  proposed coarse-grained model which is in turn used  to produce predictive estimates as will be described in the sequel.  In contrast, other dimensionality reduction methods based on the solution of an eigenvalue problem are required to solve a new system   for the whole dataset when new data is presented.

\subsection{Predicting atomistic configurations - Leveraging the exact likelihood}
\label{sec:pred}

After training the model as described in Section~\ref{sec:inference}, we are interested in obtaining the predictive distribution $p(\bx | \btheta) = \int_{\mathcal M\indcv} p\inth(\bx|\bz) p\inth(\bz)~d\bz$ (see \refeqq{eqn:predictive}) which poses a demanding computational task. One approach for predicting configurations $\bx$ distributed according to $p(\bx | \btheta)$ is ancestral sampling. Firstly, one can generate a sample $\bz^l$ from $p\inth(\bz)$ and secondly sample $\bx^{(k,l)} \sim p\inth(\bx|\bz^l)$. The variance of such estimators significantly increases with increasing $\dim(\bz)$. Ancestral sampling does not account for training the model by employing an \emph{approximate} posterior $q\inphi(\bz|\bx)$ instead of the actual posterior $p\inth(\bz|\bx)$ of the CVs $\bz$.
The \emph{Metropolis-within-Gibbs} sampling scheme~\cite{mattei2018}  accounts for grounding the optimization of the objective in~\refeqq{eqn:decomposedlowerbound} on a variational approximation.
This approach builds upon findings in~[\onlinecite{rezendre2014}] and proposes that generated samples $\bar \bx$ follow a Markov chain $(\bz_t, \bar \bx_t)$ for steps $t\geq 1$.
Ref.~[\onlinecite{mattei2018}] proposes employing the following Metropolis~\cite{metropolis1953, hastings1970} update criterion $\rho_t$ reflecting a ratio of importance ratios,
\begin{equation}
    \rho_t = \frac{\frac{ p\inth(\bar \bx_{t-1}|\tilde{\bz}_t) ~ p\inth(\bar \bz_t )}{p\inth(\bar \bx_{t-1}|\bz_{t-1}) ~ p\inth(\bz_{t-1} )}}{ \frac{ q\inphi(\tilde{\bz}_{t}| \bar{\bx}_{t-1}) }{ q\inphi(\bz_{t-1}| \bar{\bx}_{t-1})} }.
    \label{eqn:met_gibbs_ratio}
\end{equation}
Equation~(\ref{eqn:met_gibbs_ratio}) provides the needed correction when using the approximate latent variable posterior $q\inphi(\bz|\bx)$. When the CV's exact posterior is identified, i.e. when  $D_{KL}\left(q\inphi(\bz|\bx) || p\inth(\bz|\bx) \right) = 0$, all proposals $\bz_t$ in Algorithm~\ref{alg:metropolisgibbs} are accepted with $\rho_t=1$.

\begin{algorithm}[H]
	\caption{\small Metropolis-within-Gibbs Sampler~[\onlinecite{mattei2018}].}
	\begin{algorithmic}
		\small
		\INPUT Trained model $p\inth(\bx|\bz) p\inth(\bz)$ and approximate posterior $q\inphi(\bz|\bx)$. Total steps $T$.
		\Init $(\bz_0, \bar \bx_0)$.
		\FOR{$t=1$ \textbf{to} T }
		\State $\tilde{\bz}_t \sim q\inphi(\bz|\bar \bx_{t-1})$ Draw proposal $\tilde{\bz}_t$ from the approximate posterior $q\inphi(\bz|\bar \bx_{t-1})$.
		\State $\rho_t = \frac{ p\inth(\bar \bx_{t-1}|\tilde{\bz}_t) ~ p\inth(\bar \bz_t )}{p\inth(\bar \bx_{t-1}|\bz_{t-1}) ~ p\inth(\bz_{t-1} )} \frac{q\inphi(\bz_{t-1}| \bar{\bx}_{t-1}) }{q\inphi(\tilde{\bz}_{t}| \bar{\bx}_{t-1})} $ Estimate the Metropolis acceptance ratio, correcting for the use of the  approximate posterior distribution $q\inphi(\bz|\bx)$.
		\State $\bz_t = 
		\begin{cases}
		\tilde{\bz}_t & \text{with probability } \rho_t \\
		\bz_{t-1} & \text{with probability } 1-\rho_t.
		\end{cases} $
		\State $\bar \bx_t \sim p\inth(\bx | \bz_t)$
		\ENDFOR
		\State \textbf{return} $\bar \bx_{1:T}$.
	\end{algorithmic}
	\label{alg:metropolisgibbs}
\end{algorithm}

\subsection{Prior specification}

The recent work of~[\onlinecite{mattei2018}] discusses the pitfalls of overly expressive, deep, latent variable models which can yield infinite likelihoods and ill-posed optimization problems\cite{lecam1990}. 
We address these issues by regularizing  the log-likelihood with functional priors~\cite{west2003, figueiredo2003}. The prior contribution is added as an additional component in the log-likelihood as indicated in~\refeqq{eqn:map_estimate}. In addition to enhanced stability during training~\cite{mattei2018}, sparsity inducing priors alleviate the overparameterized nature of complex neural networks.

We adopt the Automatic Relevance Determination (ARD~\cite{mackay1994}) model which consists of the following distributions:
\be
p(\btheta|\boldsymbol{\tau}) \equiv  \prod_k \mathcal{N}(\theta_{ k}| 0, \tau_k^{-1}) , \quad 
\tau_k \sim  \mathrm{Gamma}( \tau_k |a_0, b_0).
\label{eqn:ardPrior}
\ee
Equation~(\ref{eqn:ardPrior}) implies modeling each $\theta_{k}$ with an independent Gaussian distribution. The Gaussian distribution has zero-mean and an independent precision hyper-parameter $\tau_k$, modeled with a (conjugate) Gamma density.  The resulting prior $p(\theta_k)$ follows (by marginalizing the hyper-parameter $\tau_k$) a heavy-tailed Student's $t-$distribution. This distribution favors a priori sparse solutions with $\theta_k$ close to zero. 
In order to compute derivatives of the log-prior, required for learning the parameters $\btheta$, we treat the $\tau_k$'s  as latent variables in an inner-loop expectation-maximization scheme~\cite{tipping2001} which consists of the following steps:
\begin{itemize}
	\item E-step - evaluate:
	\be \left\langle \tau_k \right\rangle_{p(\tau_k|\theta_{k})} =
	\frac{ a_0 + \frac{1}{2} }{ b_0 + \frac{\theta_{k}^2}{2} }.
	\label{eq:ardPosterior}
	\ee
	\item M-step - evaluate:
	\be
	\frac{\partial \log p(\btheta) }{\partial \theta_{k} } =
	- \mathbb{E}_{p(\tau_k|\theta_{k})} \left[ \tau_k \right]  \theta_{k}.
	\label{eq:priorgrad}
	\ee
\end{itemize}
The second derivative of the log-prior with respect to $\btheta$ is obtained as: 
\be
\frac{\partial^2 \log p(\btheta) }{\partial \theta_{k} \partial  \theta_{l}} 
= \left\{ \begin{array}{cc}
	- \mathbb{E}_{p(\tau_k|\theta_{k})} \left[ \tau_k \right], & \textrm{if $k=l$}\\
	0, & \textrm{ otherwise.}
\end{array}
\right.
\label{eq:priorhessian}
\ee
The ARD choice of the hyper-parameters is $a_0 = b_0=\num{1.0e-5}$.
In similar settings, e.g. coarse-graining of atomistic systems, the ARD prior identified the most salient features~\cite{schoeberl2017}, whereas in this context it improves stability and turns off unnecessary parameters for describing the training data.

\subsection{Approximate Bayesian inference for model parameters - Laplace's approximation}
\label{sec::approxBayInf}
This subsection addresses the calculation of an approximate posterior of the model parameters $\btheta$.
Thus far, we have considered point estimates of the model parameters $\btheta$ (either MLE or MAP). A fully Bayesian treatment however requires the evaluation of the normalization constant of the exact posterior distribution $p(\btheta|\bX)$ of the model parameters $\btheta$, which is computationally impractical. We advocate an approximation to the posterior of $\btheta$ that is based on  Laplace's method~\cite{mackay2003}. The latter has been rediscovered as an efficient approximation for weight uncertainties in the context of neural networks in~[\onlinecite{ritter2018}].

In Laplace's approach, the exact posterior is approximated with a normal distribution with  mean $\boldsymbol{\theta}_{\mathrm{MAP}}$ and  covariance the inverse of the negative Hessian of the log-posterior at $\boldsymbol{\theta}_{\mathrm{MAP}}$.
Here, we assume a Gaussian with  diagonal covariance matrix $\boldsymbol{S}_L = \mathrm{diag}(\bsig_L^2)$  as follows, 
\be
p(\btheta|\bX) \approx \mathcal N \left(\boldsymbol{\mu}_L, \boldsymbol{S}_L = \mathrm{diag}(\bsig_L^2)\right),
\label{eqn:laplaceposterior}
\ee
with,
\be
\boldsymbol{\mu}_L = \btheta_{\text{MAP}},
\label{eqn:post_mean}
\ee
and the diagonal entries of $\boldsymbol{S}_L^{-1}$,
\be
\sigma_{L,k}^{-2} = - \frac{\partial^2 \mathcal L(\bphi, \btheta;\bX)}{\partial \theta_k^2 } \bigg\rvert_{\btheta_{\text{MAP}}, \bphi_{\text{MAP}}} + \mathbb E_{p(\tau_k|\theta_k)} [\tau_k], 
\label{eqn:post_var}
\ee
where the term $\mathbb E_{p(\tau_k|\theta_k)} [\tau_k]$ arises from the prior via~\refeqq{eq:priorhessian}. 
The quantities in Eqs.~(\ref{eqn:post_mean}) and~(\ref{eqn:post_var}) are obtained at the last iteration (upon convergence) of the Auto-Encoding Variational Bayes algorithm. We summarize the procedure in Algorithm~\ref{alg:summary}.

\begin{algorithm}[H]
	\caption{\small Predictive Collective Variable Discovery.}
	\begin{algorithmic}[1]
		%\algsetup{linenosize=\tiny}
		\small
		\INPUT{Dataset $\bX$ with $N$ samples $\bxi \sim p_{\text{target}}(\bx)$.}
		\State $\{\btheta, \bphi\} \leftarrow$ Specify the generative model $p\inth(\bz)$, $p\inth(\bx|\bz)$ \refeqq{eqn:predictive} and the approximate posterior of the latent CVs $q\inphi(\bz|\bx)$ introduced in~\refeqq{eqn:lowerbound} with the corresponding parameters $\btheta$ and $\bphi$, respectively.
		\State $\{\btheta_{\text{MAP}}, \bphi_{\text{MAP}}\} \leftarrow$ Maximize the   lower-bound in~\refeqq{eqn:lowerbound} with stochastic variational inference, see Algorithm~\ref{alg:aevb}, and obtain the MAP estimates of the model parameters $\btheta$ and $\bphi$.
		\State $p(\btheta|\bX) \leftarrow$ Perform approximate Bayesian inference for obtaining the posterior distribution of the parameters of the generative model $\btheta$. See Section~\ref{sec::approxBayInf}.
		\State Predict the atomistic trajectory with Algorithm~\ref{alg:metropolisgibbs} for samples from the approximate posterior of the generative model parameters $\btheta^j \sim p(\btheta|\bX)$.
		\State Estimate credible intervals of observables. This step is summarized in Algorithm~\ref{alg:quantile_estimation}.
	\Return{Probabilistic estimates of observables accounting for epistemic uncertainty.}
	\end{algorithmic}
	\label{alg:summary}
\end{algorithm}

% -------------- NUMERICAL IMPLEMENTATION ----------------

\section{Numerical illustrations}
\label{sec:numericalillustrations}

The following section is devoted to the application of the proposed procedure for identifying collective variables of alanine dipeptide (ALA-2~\cite{smith1999, hermans2011}) as well as of a longer peptide i.e. ALA-15. We discuss the performance and robustness of the proposed  methodology in the presence of a small amount of training data and emphasize the predictive capabilities of the model by the Ramachandran plot\cite{ramachandran1963} and the radius of gyration. The predictions are augmented by error bars capturing epistemic uncertainty. The source code and data needed to reproduce all results presented next are available at \url{https://github.com/cics-nd/predictive-cvs}.
 
\subsection{ALA-2}
\label{sec:numill_sim_ala2}

\subsubsection{Simulation of ALA-2}
\label{sec:numill_sim_ala2_sim_details}

\begin{figure}
	\centering
	\subfigure[~ALA-2 peptide with indicated dihedral angels.]{
	\label{fig:numill_dihedral_angles}
	\includegraphics[width=0.4\textwidth]{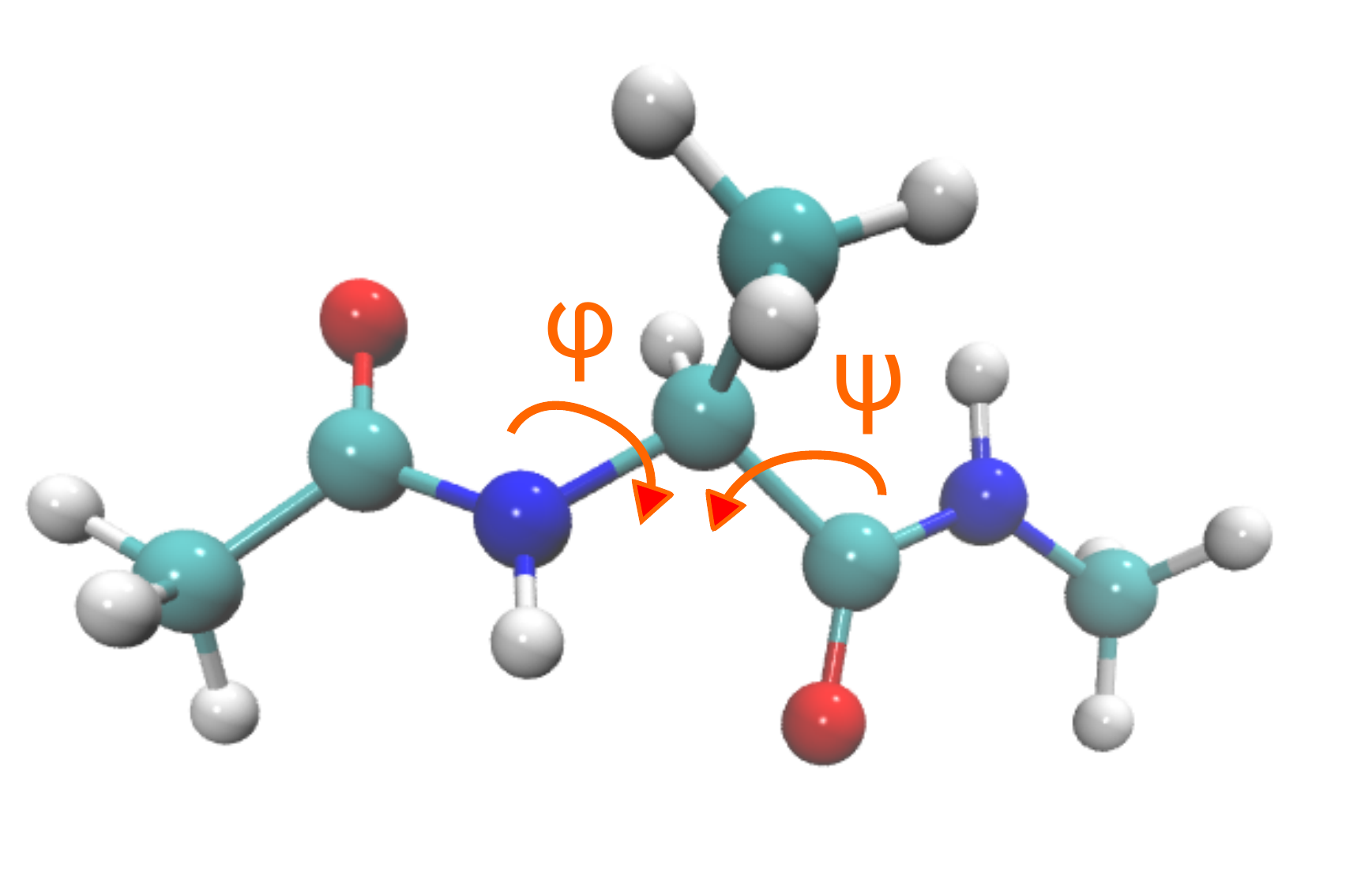}}
	\qquad
	\subfigure[~Characteristic conformations and their labelling as used in the sequel.]{
	\label{fig:numill_modes}
	\includegraphics[width=0.3\textwidth]{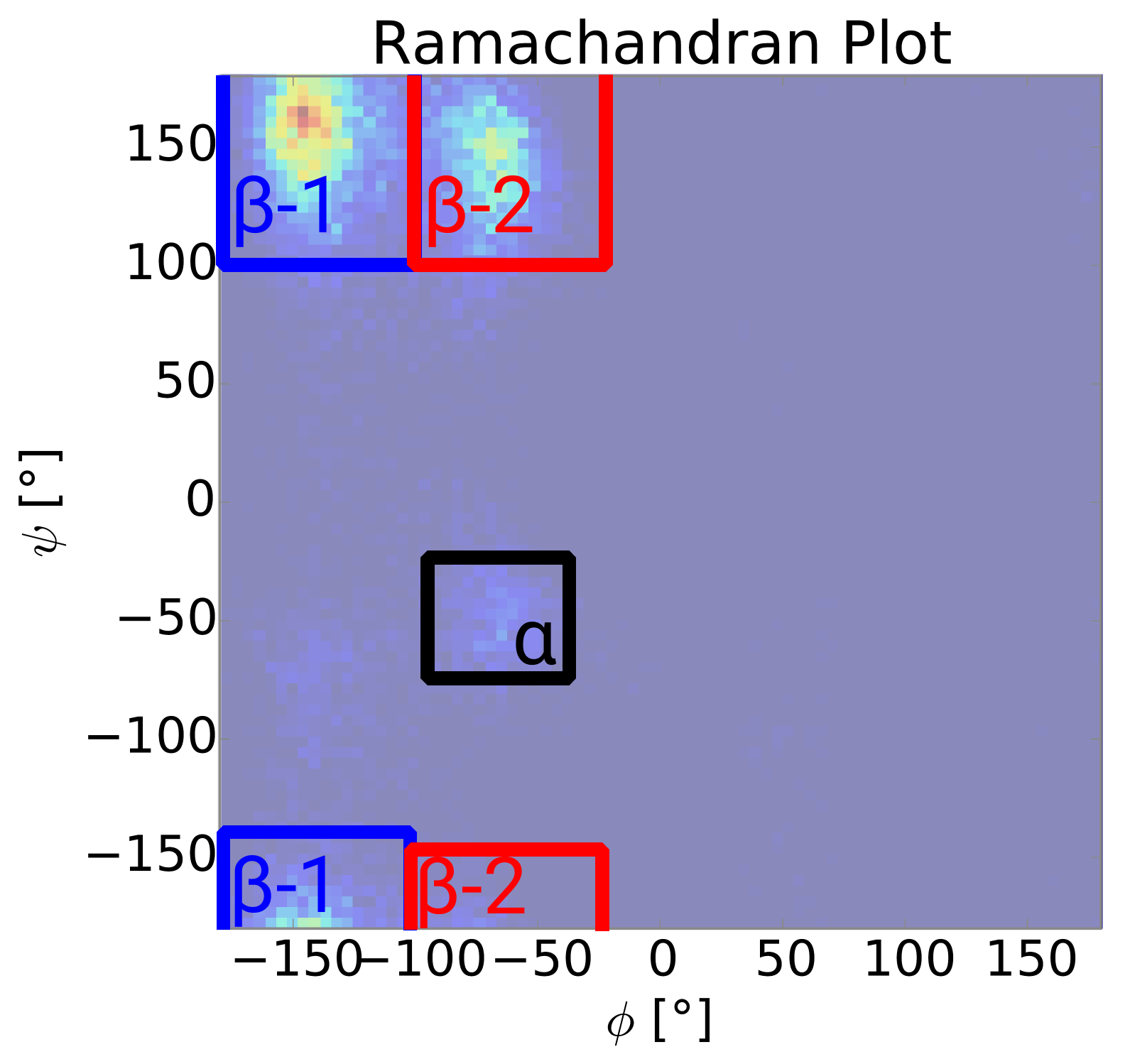}}
	\caption{Definition of the dihedral angles and the labelling of characteristic modes as utilized in this paper.}
	\label{fig:numill_ala_2_ref}
\end{figure}

Alanine dipeptide consists of $22$ atoms leading to $\dim(\bx) = 66$ in a Cartesian representation comprising the coordinates of \emph{all} atoms which we will use later on as the model input. The actual degrees of freedom (DOF) are $60$ after removing rigid-body motion.
It is well-known that ALA-2 exhibits  distinct conformations which are categorized depending on the dihedral angles $(\phi, \psi)$ (as indicated in~\reffig{fig:numill_dihedral_angles}) of the atomistic configuration. We label the three characteristic modes as $\alpha$, $\beta\textnormal{-}1$, and $\beta\textnormal{-}2$ in accordance with~[\onlinecite{vargas2002}] (see~\reffig{fig:numill_modes}).

The procedure for generating the training data for ALA-2 is similar to that in~[\onlinecite{shell2012}]. The atoms of the alanine dipeptide interact via the AMBER ff96~\cite{sorin2005, depaul2010, allen1989} force field and we employ an implicit water model based on generalized Born/solvent accessible surface area model~\cite{onufriev2004, still1990}. However, we note that an explicit water model would better represent an experimental environment.
We employ an Andersen thermostat and the simulations were carried out at constant temperature $T=\SI{330}{K}$ using Gromacs~\cite{berendsen1995, lindahl2001, spoel2005, hess2008, sander2013, prall2015, abraham2015}.
The time step is taken as $\Delta t = \SI{1}{fs}$ with an equilibration phase of $\SI{50}{ns}$.
The training dataset consisted of  snapshots taken  every $\SI{10}{ps}$ after  the equilibration phase. Rigid-body motions have been removed from the dataset.

For demonstrating the encoding into the latent CV space of atomistic 
configurations not contained in the training dataset, 
we used a test dataset selected so that the dihedral angles   $(\phi,\psi)$  had values  belonging to all three modes i.e.   $\alpha$, $\beta\textnormal{-}1$, and $\beta\textnormal{-}2$ (defined in~\reffig{fig:numill_modes}).

\subsubsection{Model specification}
\label{sec:numill_sim_ala2_model_spec}
The model requires the specification of three components. Two components are needed to describe  the generative model $p(\bx | \btheta)$: the probabilistic mapping $p\inth(\bx|\bz)$ and the distribution of the CVs $p\inth(\bz)$. The third component  is the approximate posterior $q\inphi(\bz|\bx)$ of the latent CVs as shown in~\refeqq{eqn:lowerbound}.

Following~[\onlinecite{kingma2014}], 
the distribution of the CVs is taken to be a standard Gaussian,
\be
\label{eqn:cvprior}
p\inth(\bz) = p(\bz) = \mathcal N (\bz; \boldsymbol{0}, \boldsymbol{I}).
\ee
The simplicity in the distribution in~\refeqq{eqn:cvprior} is compensated by a flexible mapping from   $\bz$ to the atomistic coordinates $\bx$. This probabilistic mapping (decoder) is given by a parametrized Gaussian as follows,
\begin{equation}
	\label{eqn:map}
	p\inth(\bx|\bz) = \mathcal N(\bx; \bmu\inth(\bz), \bm{S}\inth),
\end{equation}
where, 
\be
\bmu\inth(\bz) = f\inth^{\bmu}(\bz),
\label{eqn:ftheta}
\ee
is a non-linear  mapping  $\bz \mapsto f\inth^{\bmu}(\bz)$ ($f\inth^{\bmu} : \mathbb R^{n_{\text{CV}}} \mapsto \mathbb R^{n_f}$) parametrized by an expressive multilayer perceptron~\cite{rumelhart1986, malsburg1986, haykin1998}.

We consider a diagonal covariance matrix i.e. $\bm{S}\inth=\mathrm{diag}(\bsig^2_{\btheta})$ \cite{mattei2018} where its entries $\sigma_{\btheta, j}^2$ are treated as model parameters and do not depend on the latent CVs $\bz$.
In order to ensure the non-negativity  of  $\sigma^2_{\btheta,j} > 0$ while performing unconstrained optimization, we operate instead on $\log \sigma^2_{\btheta,j}$.

The approximate posterior $q\inphi(\bzi|\bx^{(i)})$ of the latent variables (encoder, approximating $p\inth(\bzi | \bxi)$) introduced in~\refeqq{eqn:lowerbound} is modeled by a Gaussian with flexible mean and variance represented by a neural network. For each pair of  $\bxi,\bzi$ (for notational simplicity, we drop the index $(i)$):
\begin{equation}
	q\inphi(\bz|\bx) = \mathcal N(\bz; \bmu\inphi(\bx), \bm{S}\inphi(\bx) )%\bsig\inphi^{2}(\bx) \boldsymbol I),
\end{equation}
where the covariance matrix is assumed to be diagonal i.e. $\bm{S}\inphi(\bx)=\mathrm{diag}\left(\bsig_{\bphi}^2(\bx)\right)$. 
Furthermore  $\bmu\inphi(\bx)$ and $\log \bsig_{\bphi}^2(\bx)$ are taken as the outputs of the encoding neural networks $f^{\bmu}\inphi(\bx)$ and $f^{\bsig}\inphi(\bx)$, respectively:
\begin{equation}
	\bmu\inphi(\bx) = f\inphi^{\bmu}(\bx) \quad \text{and} \quad \log \bsig_{\bphi}^2(\bx)= f\inphi^{\bsig}(\bx).
	\label{eqn:fphi}
\end{equation}
We provide further details later in this section along with the structure of the employed networks. In our model, we assume a diagonal Gaussian approximation for $q\inphi(\bz|\bx)$.

We are aware that the actual, but intractable, posterior $p\inth(\bz|\bx)$ could differ from a diagonal Gaussian and even from  a multivariate normal distribution. However, the low variance $\bsig\inphi^{2}$ observed in test cases justifies the assumption of a diagonal Gaussian in this context. An enriched model for the approximate posterior $q\inphi(\bz|\bx)$ over the CVs could rely on e.g. normalizing flows~\cite{mohamed2015}. Recent developments on autoregressive flows~\cite{welling2016} overcome the practical restriction of normalizing flows to low-dimensional latent spaces. This discussion equally holds for the assumption of a Gaussian with diagonal covariance matrix for the generative distribution $p\inth(\bx|\bz)$. In the latter case, the diagonal entries of the covariance matrix $\bm{S}\inth = \mathrm{diag}(\bsig\inth^2)$ were modeled as parameters independent of $\bz$.
Either using $\bm{S}\inth = \mathrm{diag}(\bsig\inth^2)$ or introducing a dependency on the latent CVs, $\bm{S}\inth(\bz) =\mathrm{diag}(\bsig\inth^2(\bz))$ does not influence the predictive quality in terms of observables and predicted atomistic configurations. This statement is particularly valid when an expressive model for the mean $\bmu\inth(\bz)$ in $p\inth(\bx|\bz)$ (as in this work) is considered.  It would be of interest employing more complex noise models for $p\inth(\bx|\bz)$ which e.g. could be achieved by a Cholesky parametrization~\cite{pinheiro1996}. This might reveal structure correlations  while reducing the need for higher complexity in $\bmu\inth(\bz)$.

As noted in~\refeqq{eqn:lowerbound_redvar}, we employ the reparametrization trick by writing each random variable $\bz^{(i,l)} \sim q\inphi(\bzi|\bxi)$ as
\begin{equation}
	\bz^{(i,l)} = g\inphi(\beps^{(l)}; \bxi) = \bmu\inphi(\bxi) + \bsig\inphi(\bxi) \odot \beps^{(l)},
\end{equation}
and
\begin{equation}
	\beps^{(l)} \sim p(\beps) = \mathcal N( \boldsymbol{0}, \boldsymbol{I}),
\end{equation}
where $\odot$ denotes element-wise vector product.

We utilize the following structure for the decoding neural network $f\inth^{\bmu}(\bz)$:
\be
\label{eqn:decoding_net}
f\inth^{\bmu}(\bz) =\left( l\inth^{(4)} \circ \tilde{a}^{(3)} \circ l\inth^{(3)} \circ \tilde{a}^{(2)} \circ l\inth^{(2)} \circ \tilde{a}^{(1)} \circ l\inth^{(1)} \right)(\bz).
\ee

\begin{figure}
	\centering
\newlength\spNet
\spNet=1.cm
\newlength\nodedist
\nodedist=.6cm

\newcommand{\scaletikz}{0.6}

\begin{tikzpicture}[scale=0.7, node distance = .8cm, auto,
    block/.style={
      rectangle,
      draw,%=blue,
      thick,
      fill=orange!10,
      text width=.5cm,
      text centered,
      align=center,
      rounded corners,
      minimum height=6em
    },
    line/.style={black, draw, opacity=0.5}
]

    %%%%%%%%%%%%%%%%%%%%
    %% ENCODER
    %%%%%%%%%%%%%%%%%%%%
    
    % encoder network
    \node [block] (l_1) {$l_{\boldsymbol \phi}^{(1)}$};
    \node [block, right of=l_1] (a_1) {$a^{(1)}$};
    \node [block, right of=a_1, node distance = \spNet] (l_2) {$l_{\boldsymbol \phi}^{(2)}$};
    \node [block, right of=l_2] (a_2) {$a^{(2)}$};
    \node [block, right of=a_2, node distance = \spNet] (l_3) {$l_{\boldsymbol \phi}^{(3)}$};
    %\node [block, right of=l_3] (a_3) {$a_{\boldsymbol \phi}^{(3)}$};
    %\node [block, right of=a_2, node distance = \spNet] (l_3) {$l_{\boldsymbol \phi}^{(1)}$};
    \node [block, right of=l_3] (a_3) {$a^{(3)}$};
    
    % encoder input
    \node [text width=0.5cm, left of=l_1, node distance = 1cm] (input) {$\boldsymbol x^{(i)}$};
    
    % encoder mu
    \node [block, above right=-0.5cm and \nodedist of a_3] (l_4) {$l_{\boldsymbol \phi}^{(4)}$};
    \node [text width=0.5cm, right of=l_4, node distance = 1cm] (mu) {$\boldsymbol \mu\inphi(\boldsymbol x^{(i)})$};
    
    % encoder logvar
    \node [block, below right=-0.5cm and \nodedist of a_3] (l_5) {$l_{\boldsymbol \phi}^{(5)}$};
    \node [text width=0.5cm, right of=l_5, node distance = 1cm] (logvar) {$\log \boldsymbol \sigma\inphi^2(\boldsymbol x^{(i)})$};
    
    % https://tex.stackexchange.com/questions/71478/how-to-center-one-node-exactly-between-two-others-with-tikz
    \path (mu) -- (logvar) node[midway] (post) {$q_{\boldsymbol \phi}(\boldsymbol z^{(i)} | \bxi ) = \mathcal N\left(\boldsymbol z^{(i)}| \boldsymbol \mu\inphi(\bxi),\boldsymbol{S}_{\boldsymbol{\phi}}(\bxi)\right)$};
    %\node (normal) at ($(mu)!0.5!(logvar)$) {$q_{\boldsymbol \phi}(\boldsymbol z^{(i)} ) = \mathcal N(\boldsymbol z^{(i)}| \boldsymbol \mu^{(i)}, \boldsymbol \sigma^{2 (i)} \boldsymbol I)$};
    
    % from https://tex.stackexchange.com/questions/199065/tikz-connecting-nodes-with-multiple-lines
    \foreach \x in {.0,1.} {
    \draw[dashed]($(a_1.north east)!\x!(a_1.south east)$) -- ($(l_2.north west)!\x!(l_2.south west)$);
    \draw[dashed]($(a_2.north east)!\x!(a_2.south east)$) -- ($(l_3.north west)!\x!(l_3.south west)$);
    \draw[dashed]($(a_3.north east)!\x!(a_3.south east)$) -- ($(l_4.north west)!\x!(l_4.south west)$);
    \draw[dashed]($(a_3.north east)!\x!(a_3.south east)$) -- ($(l_5.north west)!\x!(l_5.south west)$);
    }
    
    % mu and logvar to posterior distribution
    \draw[dashed, ->](mu) -> (post);
    \draw[dashed, ->](logvar) -> (post);

    %%%%%%%%%%%
    % DECODER
    %%%%%%%%%%%
    
    % decoder network
    \node [block, below right=3cm and 5.cm of l_5, ] (decl_1) {$l^{(1)}_{\boldsymbol \theta}$};
    \node [block, left of=decl_1] (deca_1) {$\tilde{a}^{(1)}$};
    
    \node [block, left of=deca_1, node distance = \spNet] (decl_2) {$l^{(2)}_{\boldsymbol \theta}$};
    \node [block, left of=decl_2] (deca_2) {$\tilde{a}^{(2)}$};
    
    \node [block, left of=deca_2, node distance = \spNet] (decl_3) {$l^{(3)}_{\boldsymbol \theta}$};
    \node [block, left of=decl_3] (deca_3) {$\tilde{a}^{(3)}$};
    
    % decoder mu
    \node [block, left of=deca_3, node distance = \spNet] (decl_4) {$l^{(4)}_{\boldsymbol \theta}$};
    \node [text width=0.9cm, left of=decl_4, node distance = 1.3cm] (decmu) {$\boldsymbol \mu_{\boldsymbol \theta}(\boldsymbol z)$};
    % decoder logvar
    %\node [block, below left=-0.5cm and \nodedist of deca_3] (decl_5) {$l^5_{\boldsymbol \theta}$};
    %\node [text width=0.5cm, left of=decl_5, node distance = 1cm] (declogvar) {$\log \boldsymbol \sigma^2(\boldsymbol z)$};
    
    % decoder input
    \node [text width=2.cm, right of=decl_1, node distance = 1.5cm] (decinput) {$\boldsymbol z \sim p_{\boldsymbol \theta}(\boldsymbol z)$};
    
    % https://tex.stackexchange.com/questions/71478/how-to-center-one-node-exactly-between-two-others-with-tikz
    %\path (decmu) -- (declogvar) 
    \node[left of=decmu, node distance = 3.3cm] (pred) {$p_{\boldsymbol \theta}(\boldsymbol x | \boldsymbol z ) = \mathcal N\left(\boldsymbol x | \boldsymbol \mu_{\boldsymbol \theta}(\bz), \boldsymbol{S}_{\boldsymbol{\theta}} \right)$};
    %\node (normal) at ($(mu)!0.5!(logvar)$) {$q_{\boldsymbol \phi}(\boldsymbol z^{(i)} ) = \mathcal N(\boldsymbol z^{(i)}| \boldsymbol \mu^{(i)}, \boldsymbol \sigma^{2 (i)} \boldsymbol I)$};
    
    \foreach \x in {.0,1.} {
    \draw[dashed]($(deca_1.north west)!\x!(deca_1.south west)$) -- ($(decl_2.north east)!\x!(decl_2.south east)$);
    \draw[dashed]($(deca_2.north west)!\x!(deca_2.south west)$) -- ($(decl_3.north east)!\x!(decl_3.south east)$);
    \draw[dashed]($(deca_3.north west)!\x!(deca_3.south west)$) -- ($(decl_4.north east)!\x!(decl_4.south east)$);
    %\draw[dashed]($(deca_3.north west)!\x!(deca_3.south west)$) -- ($(decl_5.north east)!\x!(decl_5.south east)$);
    }
    
    % mu and logvar to posterior distribution
    \draw[dashed, ->](decmu) -> (pred);
    \draw[dashed, ->](decl_4) -> (decmu);
    
    %\draw[dashed, ->](declogvar) -> (pred);
     
    \path [line, red, dashed, <->, very thick] ([xshift=0.8cm]post.south west) to [out=330,in=90] %node[above,xshift=1cm,yshift=-0.9cm] {$-\sum_{i=1}^N D_{KL}(q\inphi(\bzi|%\bxi) || p\inth(\bz)) $} 
    (decinput);
    
    \path [line, blue, dashed, <->, very thick] (input) to [out=270,in=145] node[above,xshift=7.cm,yshift=-1.2cm, opacity=1., black] {$\mathcal L(\btheta, \bphi; \bX) =  
    \sum_{i=1}^N
    {\color{blue}    
    \mathbb E_{q\inphi(\bzi|\bxi)} [\log p\inth(\bxi|\bzi) ]
    }
    -\sum_{i=1}^N
    {\color{red}    
    D_{KL}\left( q\inphi(\bzi|\bxi) || p\inth(\bz) \right)
    }
    $
    }  ([xshift=.5cm]pred.north west);

    \draw [black,thick, dotted, opacity=0.5]  ($(l_1.north west)+(-1.5,3.)$)  rectangle ($( post.east)+(0.5,-4.5)$);
    
    \node (encoder) at ($(l_1.north west)+(-0.5, 2.5)$) [minimum width=2cm,minimum height=0.1cm,label={ Encoder},anchor=south] {};

    \draw [black,thick,dotted, opacity=0.5]  ($(pred.north west)+(-0.7, 1.5)$)  rectangle ($( decinput.east)+(0.5,-2)$);
    
    \node (decoder) at ($(pred.north west)+(+0.2, 1.0)$) [minimum width=2cm,minimum height=0.1cm,label={ Decoder},anchor=south] {};

\end{tikzpicture}
	\caption{Schematic of the AEVB depicting the employed network architecture. Fully connected linear layers are denoted with $l^{(i)}$ and non-linear activation functions with $a^{(i)}$. The indices $\bphi$ and $\btheta$ indicate encoding and decoding networks, respectively.
The maximization of the lower-bound on the marginal log-likelihood $\mathcal L (\btheta, \bphi; \bX) $ in~\refeqq{eqn:decomposedlowerbound} simultaneously optimizes the parametrization of the encoder and decoder. The first term in $\mathcal L (\btheta, \bphi; \bX) $ accounts for the reconstruction of the training data $\bxi$ with $\bzi$ distributed according $q\inphi(\bzi|\bxi)$.  The second term, in aggregation of all data $\bxi$, ensures that $q\inphi(\bzi|\bxi)$ is close to $p(\bz)$. }
	\label{fig:aevb_net_structure}
\end{figure}
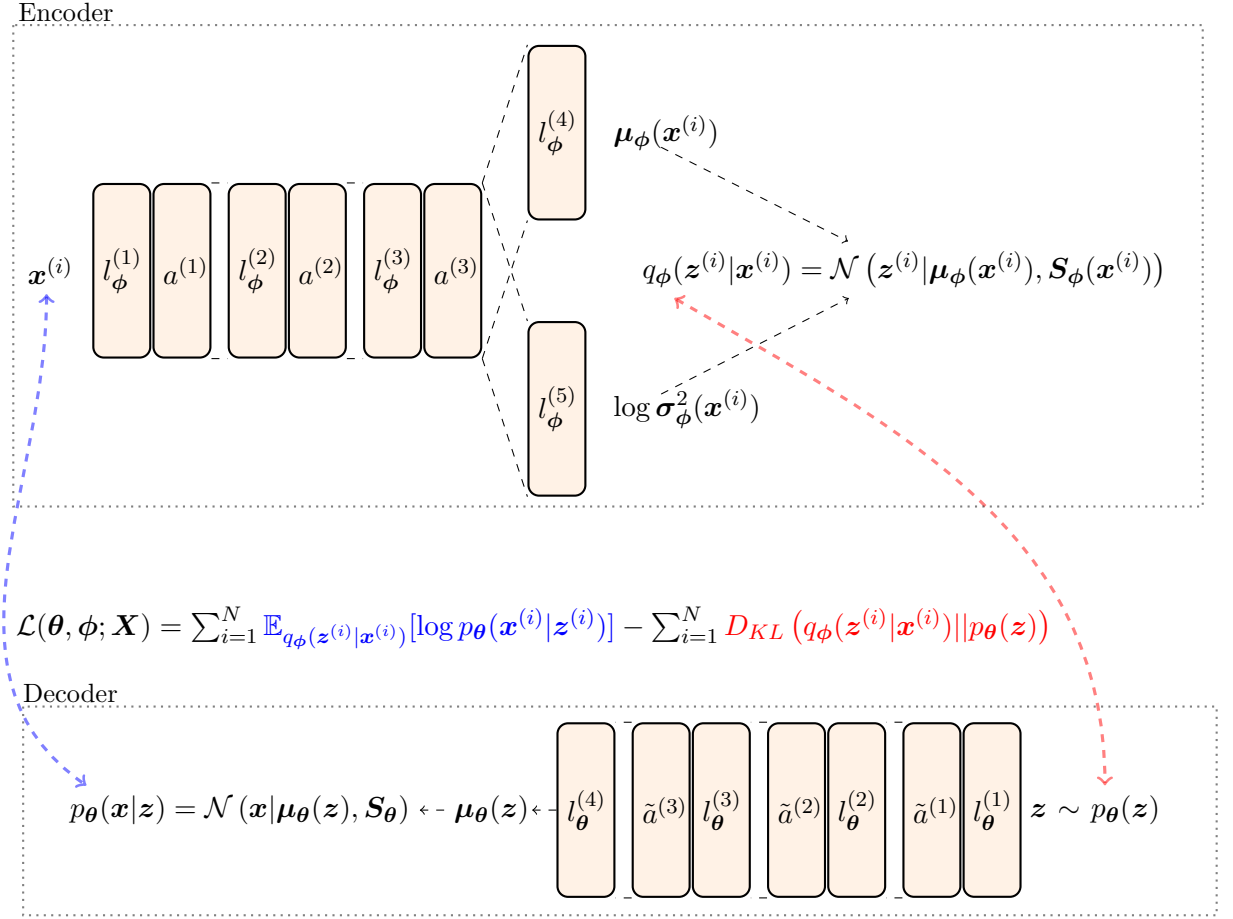

The encoding networks for obtaining $\bmu\inphi(\bx)$ and $\bsig^2\inphi(\bx)$ of the approximate posterior $q\inth(\bz|\bx)$ over the latent CVs share the structure,
\be
\label{eqn:encoding_net_shared}
f\inphi(\bx) = \left( a^{(3)} \circ l\inphi^{(3)} \circ a^{(2)} \circ l\inphi^{(2)} \circ a^{(1)} \circ  l\inphi^{(1)} \right)(\bx),
\ee
which gives rise to $f^{\bmu}\inphi(\bx)$ and $f^{\bsig}\inphi(\bx)$ with,
\be
\label{eqn:encoding_mu_sig}
f^{\bmu}\inphi(\bx) = l\inphi^{(4)}\left(f\inphi(\bx) \right) \quad \text{and} \quad 
f^{\bsig}\inphi(\bx) = l\inphi^{(5)}\left(f\inphi(\bx) \right).
\ee
In Eqs.~(\ref{eqn:decoding_net})-(\ref{eqn:encoding_mu_sig}), we consider linear layers $l^{(i)}$ of a variable $\boldsymbol{y}$ with $l^{(i)}(\boldsymbol{y}) = \boldsymbol{W}^{(i)} \boldsymbol{y} + \boldsymbol{b}^{(i)}$ and non-linear activation functions  denoted   with $a(\cdot)$. The indices $\bphi$ and $\btheta$ of the linear layers $l^{(i)}$ reflect  correspondence to either the encoding or decoding network, respectively. $\bphi$ comprises all parameters of the encoding networks $f\inphi^{\bmu}(\bx)$ and $f\inphi^{\bsig}(\bx)$, $\btheta$ all parameters of the decoding network $f\inth(\bz)$ including the parameters $\bsig^2\inth$ discussed in~\refeqq{eqn:map}. We differentiate the encoding and decoding activation functions by denoting them as  $a^{(i)}$ and $\tilde{a}^{(i)}$, respectively. All  layers considered were fully connected. The general architecture of the neural networks employed and how these affect the objective $\mathcal L (\btheta, \bphi; \bX) $ are depicted in Fig.~\ref{fig:aevb_net_structure}. 
%%%%%%%%%%%%%%%%%%%%
%%% Uncomment the following figure in case the tikz input is not valid. Then comment out the following figure. Both are for Fig. 3.
%%%%%%%%%%%%%%%%%%%%
%\begin{figure}
%    \centering
%    \includegraphics[width=\textwidth]{figures/aevb_scheme.pdf}
%	\caption{Schematic of the AEVB depicting the employed network architecture. Fully connected linear layers are denoted with $l^{(i)}$ and non-linear activation functions with $a^{(i)}$. The indices $\bphi$ and $\btheta$ indicate encoding and decoding networks, respectively. The maximization of the lower-bound on the \textcolor{blue}{marginal} log-likelihood $\mathcal L (\btheta, \bphi; \bX) $ in~\refeqq{eqn:decomposedlowerbound} simultaneously optimizes the parametrization of the encoder and decoder. The first term in $\mathcal L (\btheta, \bphi; \bX) $ accounts for the reconstruction of the training data $\bxi$ with $\bzi$ distributed according $q\inphi(\bzi|\bxi)$.  The second term, in aggregation of all data $\bxi$, ensures that $q\inphi(\bzi|\bxi)$ is close to $p(\bz)$. }
%	\label{fig:aevb_net_structure}
%\end{figure}

The optimization of the objective is carried out by a stochastic gradient ascent algorithm. In our case, we employ ADAM~\cite{kingmaAdam2014} with the parameters chosen as   $\alpha = 0.001,\beta_1= 0.9, \beta_2 =  0.999, \epsilon_{\text{ADAM}}=\num{1.0e-8}$. Gradients of the lower-bound $\mathcal L (\btheta, \bphi; \bX)$ with respect to the model parametrization $\{\bphi, \btheta \}$ are estimated by the backpropagation procedure~\cite{rumelhart1986}.
The required gradients for optimizing the parameters $\bsig^2\inth$ can be computed analytically. For an entry $\sigma_{j,\btheta}^2$, we can write the following:
\begin{align}
\frac{\partial \mathcal L (\btheta, \bphi; \bxi) }{\partial \log \sigma_{j,\btheta}^2} &= \frac{\partial \log p\inth(\bxi|\bz) }{\partial \log \sigma_{j,\btheta}^2} \nonumber \\
&= \frac{\partial }{\partial \log \sigma_{j,\btheta}^2} \left( -\frac{1}{2} \sum_{j=1}^{\dim(\bxi)} \frac{ \left( x^{(i)}_j - \mu_{j, \btheta}(\bz) \right)^2}{\sigma_{j,\btheta}^2} \right) \nonumber \\
&= \frac{\partial }{\partial \log \sigma_{j,\btheta}^2} \left( -\frac{1}{2} \sum_{j=1}^{\dim(\bxi)} \frac{ \left( x^{(i)}_j - \mu_{j, \btheta}(\bz) \right)^2}{\exp\left( \log \left( \sigma_{j,\btheta}^2\right) \right) }  \right) \nonumber \\
&= \frac{1}{2} \frac{ \left( x^{(i)}_j - \mu_{j, \btheta}(\bz) \right)^2}{ \sigma_{j,\btheta}^2 }.
\label{eqn:grad_decoder_sigsq}
\end{align}

\begin{table}[]
	\centering
	\begin{tabular}{|l l l l l |}
		\hline
		Linear layer & Input dimension & Output dimension & Activation layer & Activation function \\
		\hline
		\hline
		$l\inphi^{(1)}$ & $\dim(\bx)$ & $d_1$ &  $a^{(1)}$ & SeLu\footnote{SeLu: $a(x) = 
			\begin{cases}
			\alpha (e^x -1) & \text{if } x<0\\
			x            & \text{otherwise}.
			\end{cases} $ See~[\onlinecite{klambauer2017}] for further details.}\\
		$l\inphi^{(2)}$ & $d_1$ & $d_2$ &  $a^{(2)}$ & SeLu \\
		$l\inphi^{(3)}$ & $d_2$ & $d_3$ &  $a^{(3)}$ & Log Sigmoid \footnote{Log Sigmoid: $a(x) = \log \frac{ 1 }{ 1 + e^{-x}} $}  \\
		$l\inphi^{(4)}$ & $d_3$ & $\dim(\bz)$ &  None & - \\
		$l\inphi^{(5)}$ & $d_3$ & $\dim(\bz)$ &  None & - \\
		\hline
	\end{tabular}
	\caption{Network specification of the encoding neural network with $d_1=50$, $d_2=100$, and $d_3=100$.}
	\label{tab:encoder_actfct}
\end{table}

\begin{table}[]
	\centering
	\begin{tabular}{|l l l l l |}
		\hline
		Linear layer & Input dimension & Output dimension & Activation layer & Activation function \\
		\hline
		\hline
		$l\inth^{(1)}$ & $\dim(\bz)$ & $d_3$ &  $\tilde{a}^{(1)}$ & Tanh \\
		$l\inth^{(2)}$ & $d_3$ & $d_2$ &  $\tilde{a}^{(2)}$ & Tanh \\
		$l\inth^{(3)}$ & $d_2$ & $d_1$ &  $\tilde{a}^{(3)}$ & Tanh \\
		$l\inth^{(4)}$ & $d_1$ & $\dim(\bx)$ &  None & - \\
		\hline
	\end{tabular}
	\caption{Network specification of the decoding neural network with $d_{\{1,2,3\}}$ as defined in Table~\ref{tab:encoder_actfct}.}
	\label{tab:decoder_actfct}
\end{table}

\begin{figure}
	\centering
	\subfigure[~Varying dimensionality of the layers $l^{(i)}_{\{\boldsymbol{\theta},\boldsymbol{\phi}\}}$. The figure's labels represent the dimensionality of the layers in the format $d_1$-$d_2$-$d_3$ as specified in Tables~\ref{tab:encoder_actfct} and~\ref{tab:decoder_actfct}. We use the activation functions as denoted in the tables.]{
	\label{fig:aevb_net_dim}
	\includegraphics[height=0.35\textwidth]{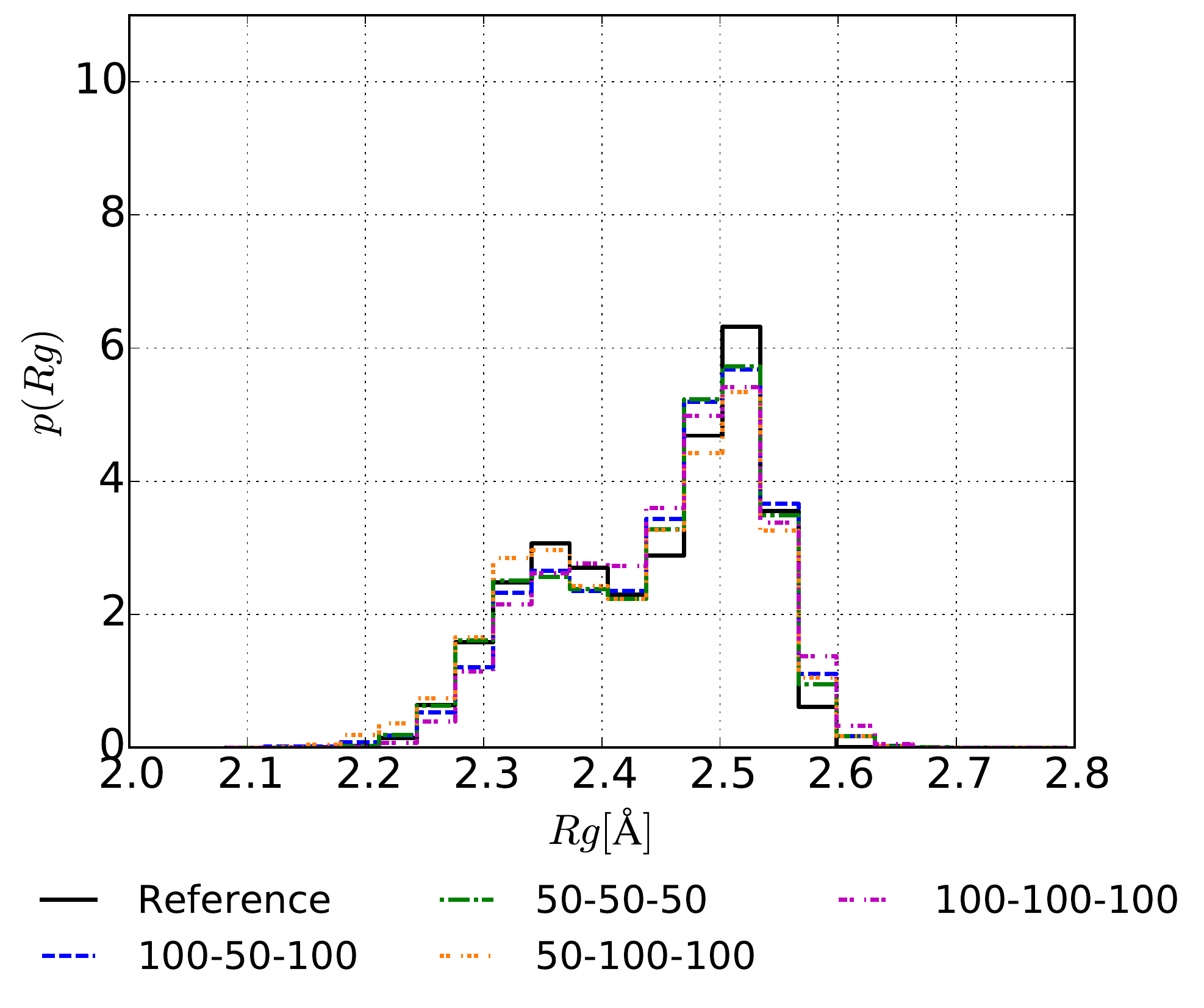}}
	\qquad
	\subfigure[~Testing different activation functions for $a^{(i)}$. The labels specify the utilized activation functions in the following manner: $a^{(1)}$-$a^{(2)}$-$a^{(3)}$-$\tilde{a}^{(1)}$-$\tilde{a}^{(2)}$-$\tilde{a}^{(3)}$. We use the abbreviations:
	t: Tanh, s: SeLu, ls: Log Sigmoid.]{
	\label{fig:aevb_actfct}
	\includegraphics[height=0.35\textwidth]{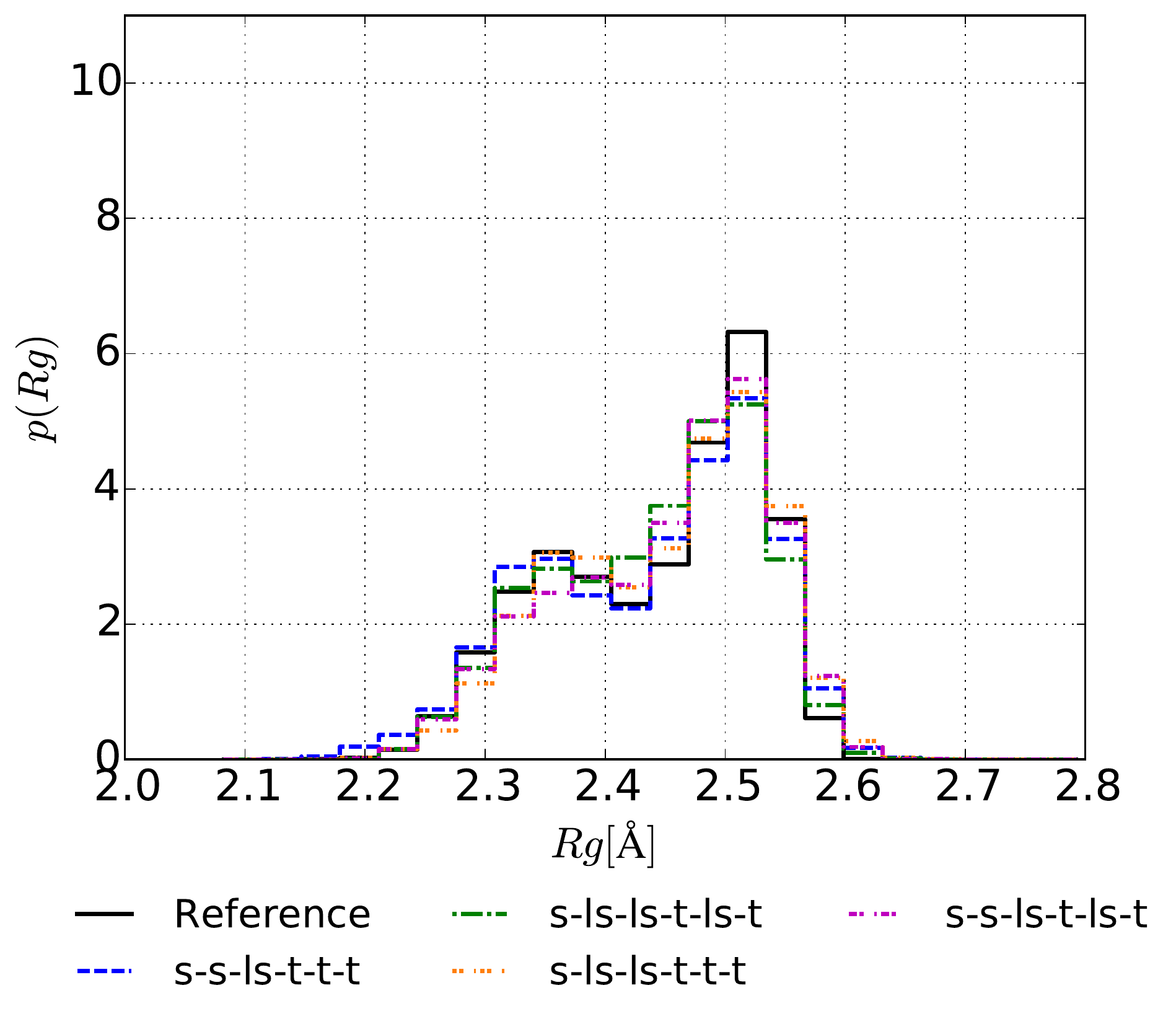}}
	\caption{Prediction of the radius of gyration with differing networks, in terms of (a) the dimensionality of the layers and (b) regarding the type of activation functions used. Changes in the network specification lead to similar predictions. This model has been trained with a dataset of size $N=500$.}
	\label{fig:aevb_net_specification}
\end{figure}

\begin{figure}
    \centering
    \includegraphics[width=0.4\textwidth]{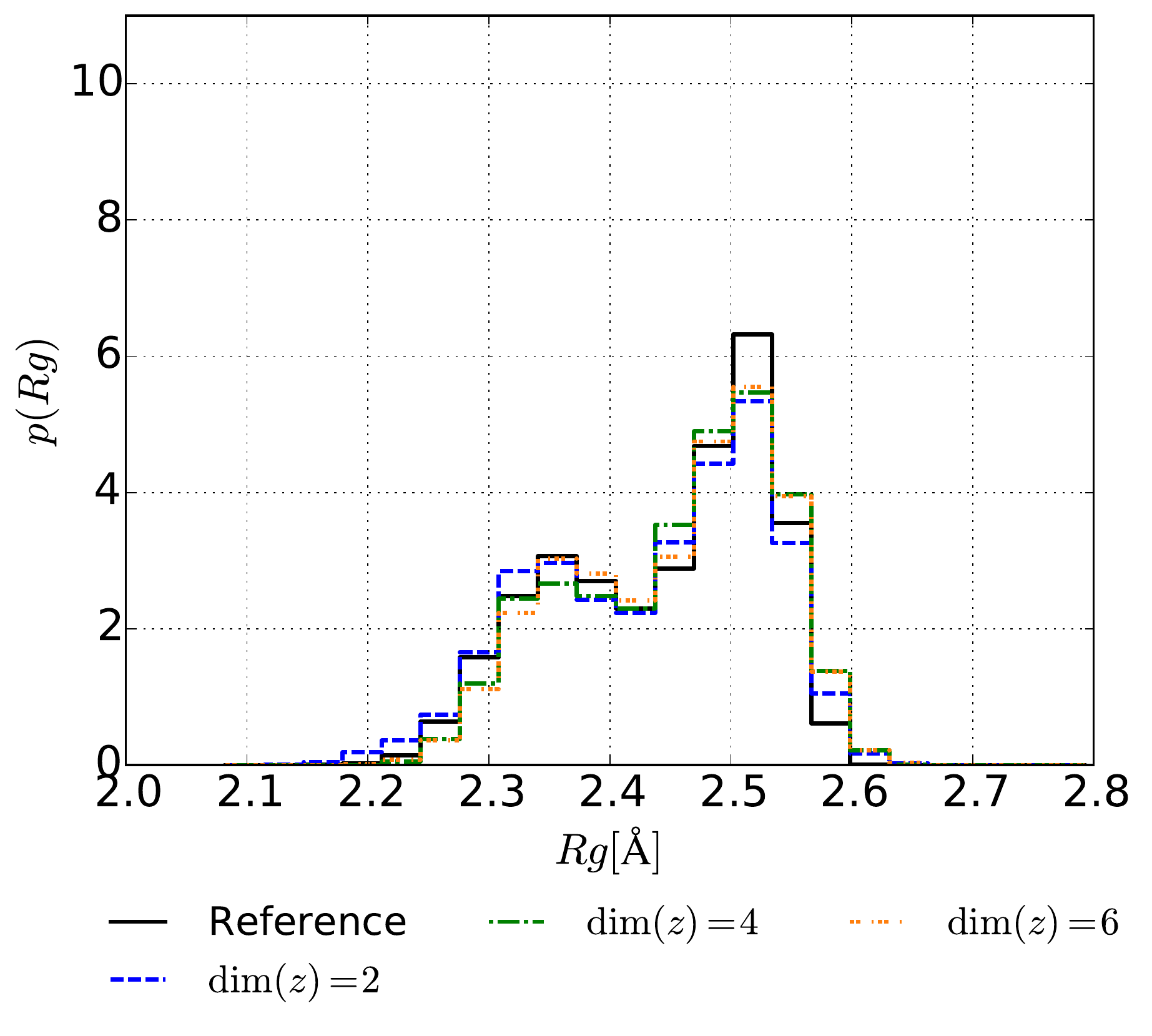}
    \caption{Predicted radius of gyration for models utilizing different $\dim(\bz)$. 
    %Except varying $\dim(\bz)$, 
    The predictions are based on a model as specified in Tables~\ref{tab:encoder_actfct} and~\ref{tab:decoder_actfct} with $N=500$.}
    \label{fig:aevb_ala2_latent_dim}
\end{figure}

Studying different combinations of activation functions and layers for the encoding network $f^{\bmu, \bsig}\inphi(\bx)$ and decoding network $f\inth^{\bmu}(\bz)$  led to the  network architecture depicted in Tables~\ref{tab:encoder_actfct} and~\ref{tab:decoder_actfct}, respectively. This network provided a repeatedly stable optimization during training. Variations of the given network architecture resulted into similar predictive capabilities as shown in Fig.~\ref{fig:aevb_net_specification}.
Stability is not limited to symmetric encoding and decoding activation functions. An automated approach for selecting or learning the best architecture is an active research area~\cite{ramachandran2018}.
Increasing the dimension of $\bz$ did not improve the predictive capabilities as shown in Fig.~\ref{fig:aevb_ala2_latent_dim}. This implies that CVs with $\dim(\bz)=2$ suffice to capture the physics encapsulated in the ALA-2 dataset with $\dim(\bx)=66$ or $60$ DOF.

\subsubsection{Results}

In the following illustrations, we trained the model by varying the  number of snapshots  $N$. We utilized  a sub-sampled batch of size $M=64$ from the dataset of size $N$. In cases where $N<64$, we set $M=N$. The hyper parameters of the ARD prior in~\refeqq{eqn:ardPrior} are set to $a_0 = b_0 = \num{1.0e-5}$. Other values for $a_0, b_0$ in the range of $[\num{1.0e-8}, \num{1.0e-4}]$ were also employed without a significant effect on the obtained sparsity patterns or the predictive accuracy of the model.

\begin{figure}
	\centering
	\includegraphics[width=0.6\textwidth]{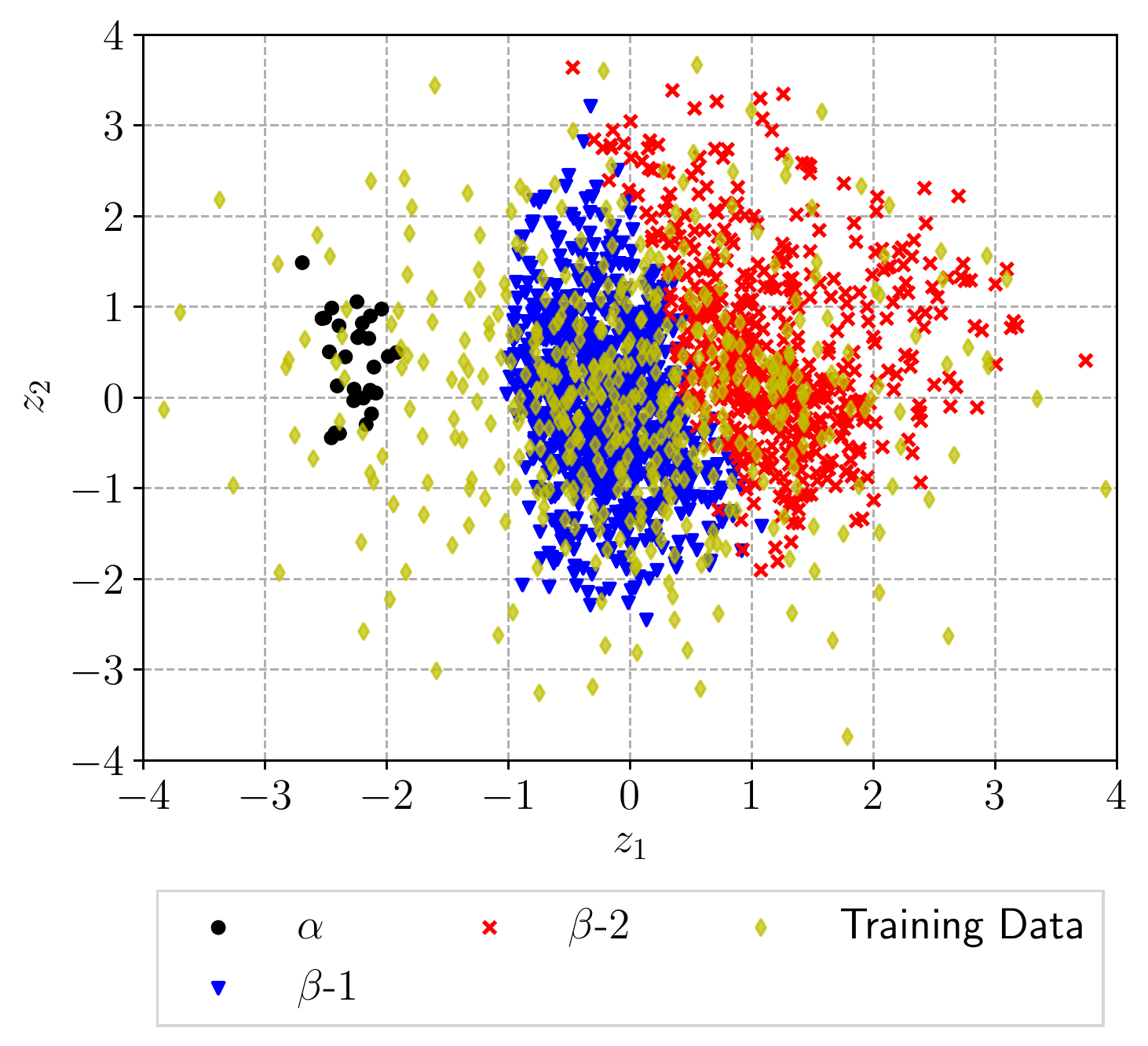}
	\qquad
	\caption{Representation of the $\bz$-coordinates of the training data $\bX$ with $N=500$ in the CV space (yellow diamonds). Using the  trained model and  the mean of $q\inphi(\bz|\bz)$ we computed the $\bz$-coordinates of $1527$ test samples corresponding to  different conformations of the alanine dipeptide  to $\alpha$ (black), $\beta\textnormal{-}1$ (blue), and $\beta\textnormal{-}2$ (red). Without any prior physical information, the encoder yields  three distinct clusters in the CV space.
	}
	\label{fig:aevb_lat_rep}
\end{figure}

Figure~\ref{fig:aevb_lat_rep} depicts the  $\bz$-coordinates of $N=500$ training data as well as those of $1527$ test data which have been classified into the three modes based on the values of the dihedral angles (see \reffig{fig:numill_modes}).
In order to obtain the $\bz$-coordinates of the test data, we made use of the  mean $\bmu\inphi(\bxi)$ of the inferred approximate posterior $q\inphi$ as obtained after training. The resulting picture essentially  provides the pre-images of the atomistic configurations in the CV space.
Interestingly, similar atomistic configurations, i.e. belonging to one of the three   modes, $\alpha, \beta\textnormal{-}1, \beta\textnormal{-}2$, are recognized by $q\inphi(\bz|\bx)$ and mapped to clusters in the identified CV space. 
$\beta\textnormal{-}1$ configurations are encoded by $q\inphi(\bz|\bx)$ to regions with high probability mass in $p\inth(\bz)$, i.e. CVs $\bz$ close to the center of $p\inth(\bz)=N(\boldsymbol{0}, \boldsymbol{I})$ are assigned. This is in accordance with the reference Boltzmann distribution $p(\bx)$ where $\beta\textnormal{-}1$ is the most probable conformation.

Various dimensionality reduction methods are designed in order to keep similar $\bx$ close in their embedding on the lower-dimensional CV manifold, e.g., multidimensional scaling~\cite{troyer1995} or ISOMAP~\cite{tenenbaum2000}. In the presented scheme, the generative model learns that mapping similar $\bx$ to similar $\bz$ leads to an expressive (in terms of the marginal likelihood) lower-dimensional representation. This similarity is revealed by inferring the approximate latent variable posterior $q\inphi(\bz|\bx)$. Therefore, the desired similarity mentioned in~[\onlinecite{rohrdanzclementi2013}] between configurations in the atomistic representation $\bx$ and via $q\inphi(\bz|\bx)$ in the assigned CVs $\bz$ is achieved.

\begin{figure}
	\centering
	\includegraphics[width=0.8\textwidth]{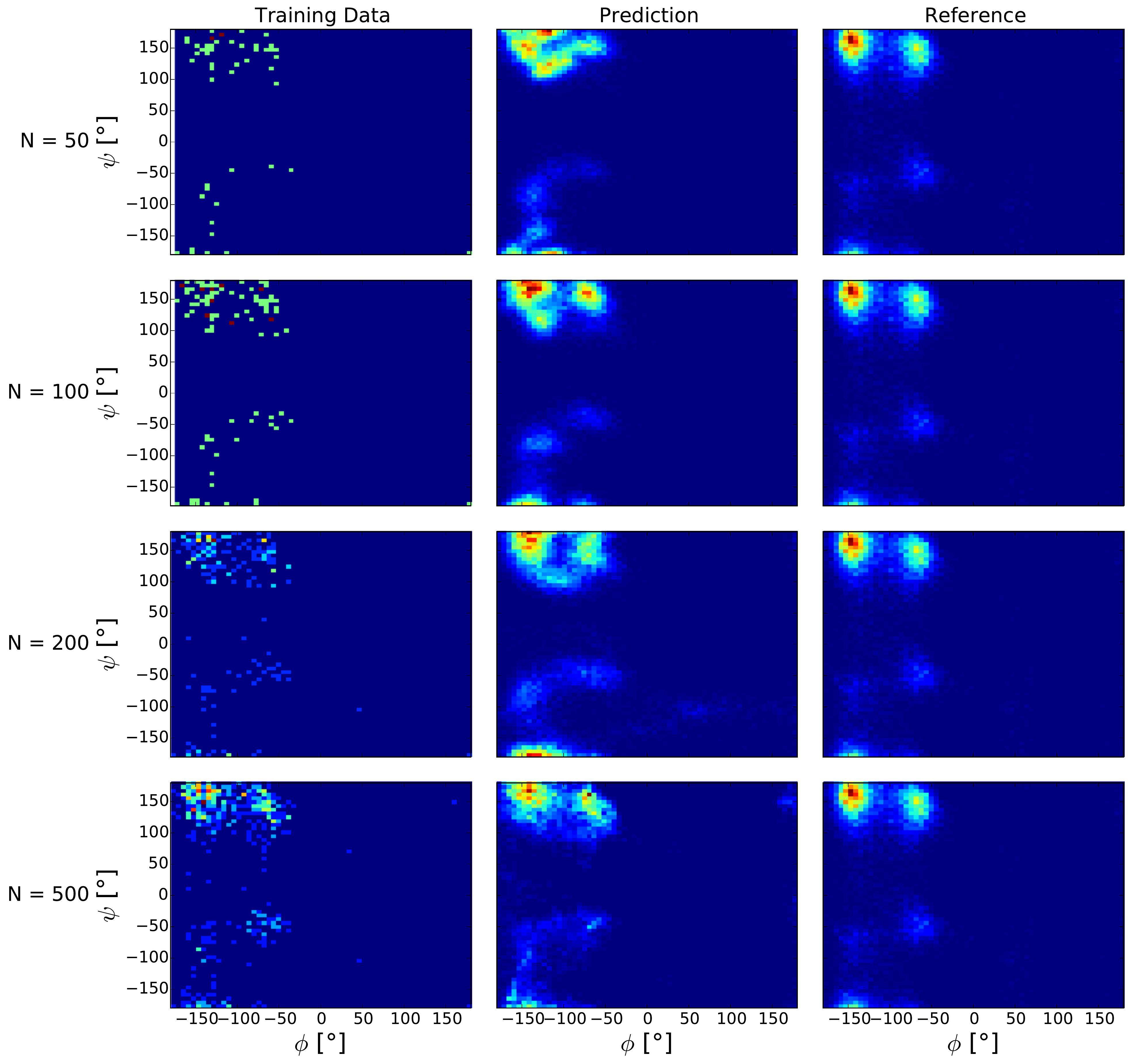}
	\qquad
	\caption{Ramachandran plots estimated with the training data $\bX$ (left column), using predictions of the trained model (middle column), and the reference (right column, estimated with $N=\num{10000}$). Each row refers to different size  $N$ of training datasets (the figure on the right column is repeated to allow easy comparison with the results on the first two columns). The represented predictions are obtained by applying Algorithm~\ref{alg:metropolisgibbs} with $T=\num{10000}$ samples.
	The generative nature of the model allows more accurate estimates than when using the training data alone. In addition,   the Bayesian approach allows for predictions with their   associated uncertainties  as discussed subsequently.}
	\label{fig:aevb_predictive_ability_sparse_low_data}
\end{figure}

In contrast to several other dimensionality reduction techniques (e.g. Isomap~\cite{tenenbaum2000} and Diffusion maps~\cite{coifman2005, coifman2005b, nadler2006, ferguson2011b}), which as mentioned in the introduction require large amounts of training data e.g. $N > \num{10000}$~\cite{rohrdanzclementi2013, duan2013}, the proposed method can perform well in the small data regime, e.g. for $N=50$ as shown in ~\reffig{fig:aevb_predictive_ability_sparse_low_data}. The latter depicts the Ramachandran plot in terms of the dihedral angles based on various amounts of training data $N$ and compares it with the one predicted by the trained model on the same $N$ as well as with the reference (obtained with $N=\num{10000}$).
We note that  the trained model yields Ramachandran plots that more closely resemble the reference as compared to the ones computed by the training data alone. 
The encoder, trained with $N=200$, is capable of generating atomistic configurations leading to $(\phi, \psi)$ tuples which are not included in the training data.

\begin{figure}
    \centering
	\subfigure[~Active ARD prior.]{
	\label{fig:aevb_n_50_latent_no_ard}
	\includegraphics[width=0.4\textwidth]{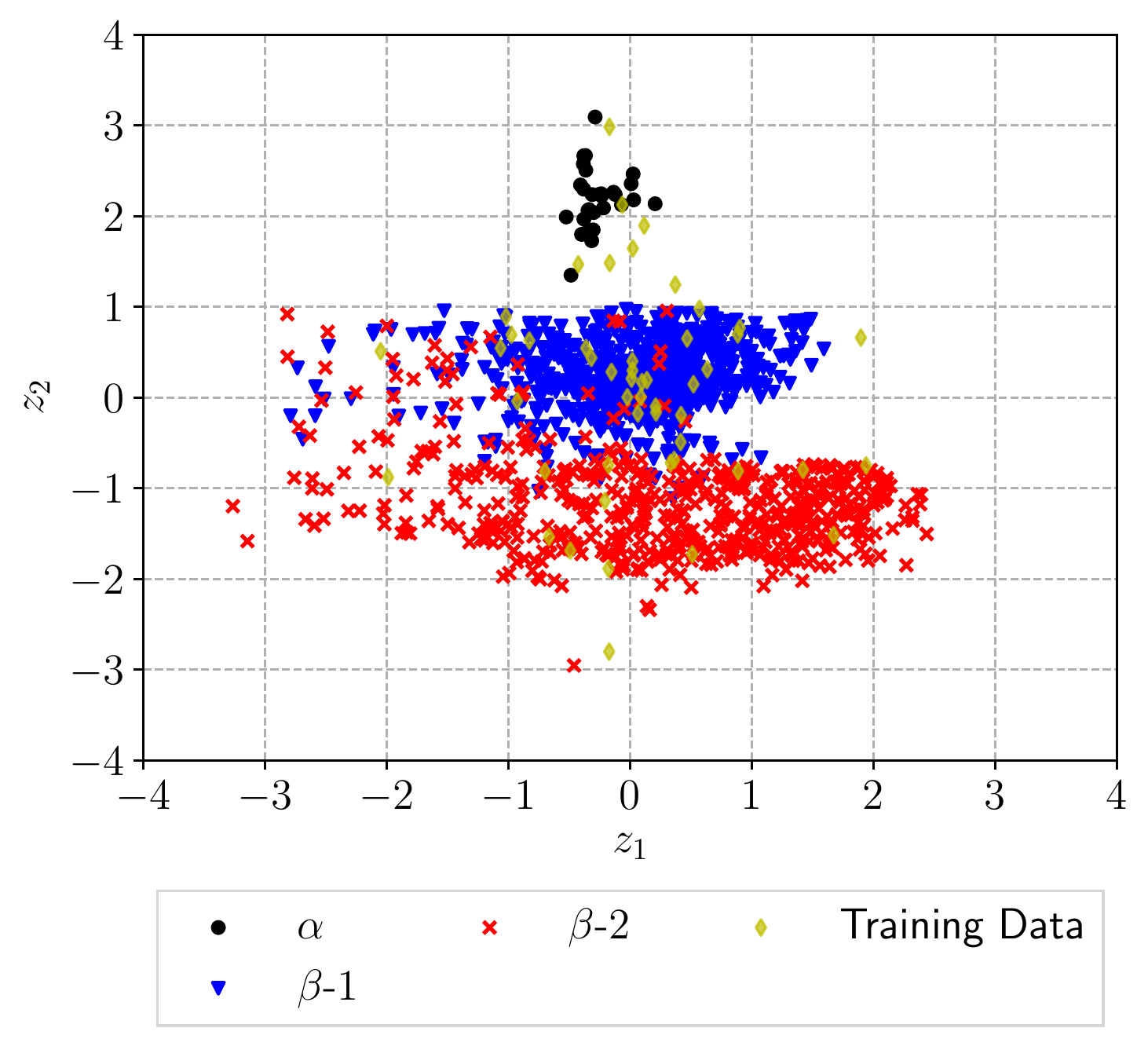}}
	\qquad
	\subfigure[~Without ARD prior.]{
	\label{fig:aevb_n_50_latent_ard}
	\includegraphics[width=0.4\textwidth]{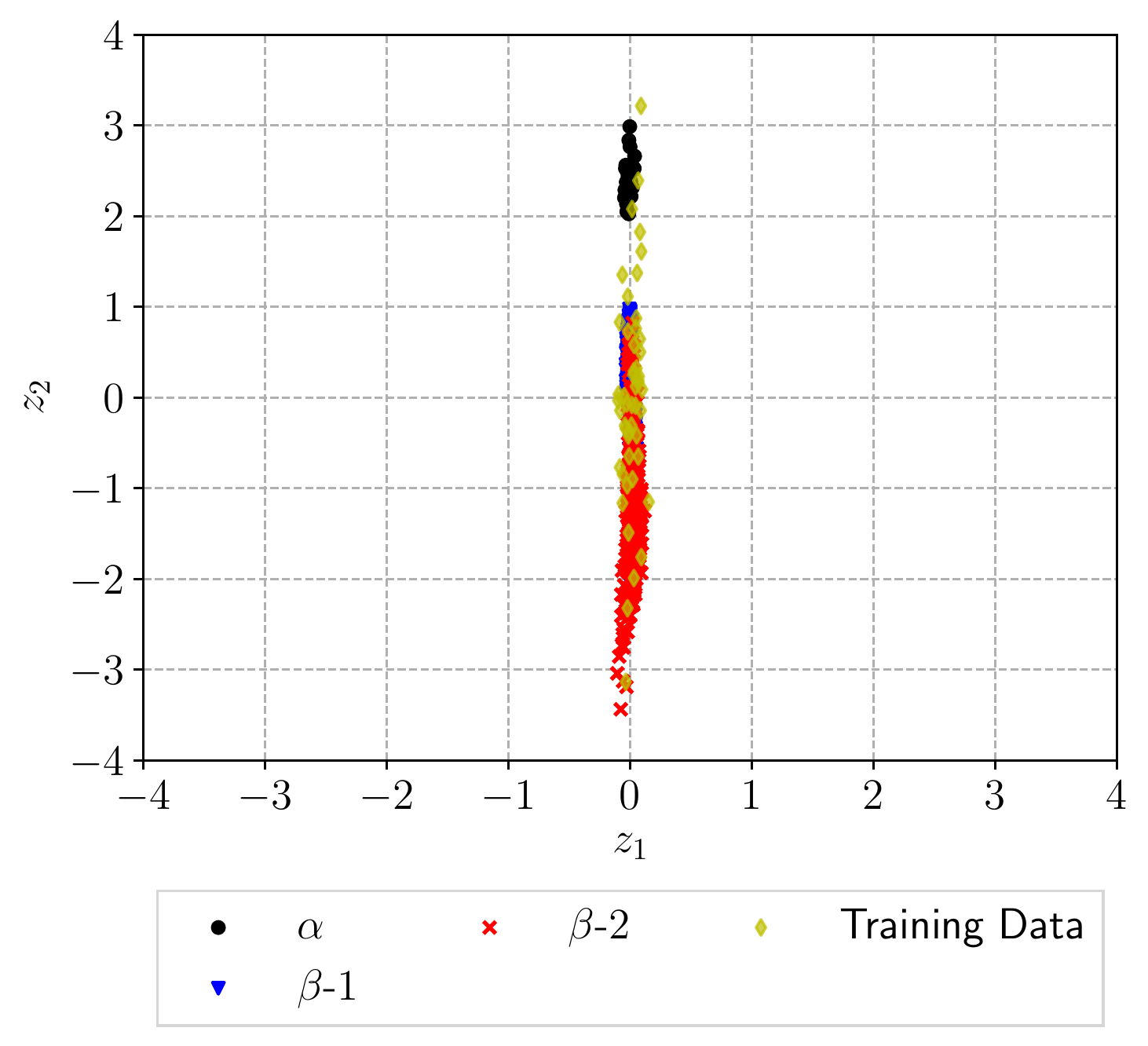}}
    \caption{Representation of the $\bz$-coordinates of the training data $\bX$ with $N=50$ in the CV space (yellow diamonds). Using the  trained model and  the mean of $q\inphi(\bz|\bz)$ we computed the $\bz$-coordinates of $1527$ test samples corresponding to  different conformations of the alanine dipeptide  to $\alpha$ (black), $\beta\textnormal{-}1$ (blue), and $\beta\textnormal{-}2$ (red). % following the labelling in~[\onlinecite{vargas2002}].  
    In the case of limited training data, the ARD prior facilitates the identification of physically meaningful CVs (left) compared to the representation on the right obtained without the ARD prior. Note that the changed positioning of the conformations in the CV space compared to~\reffig{fig:aevb_lat_rep} is due to symmetries in $p\inth(\bz)$.}
    \label{fig:aevb_n_50_compare_latent}
\end{figure}

The  ARD prior  in~\refeqq{eqn:ardPrior}  drives  $58$\% of the parameters $\btheta$ to zero (as a threshold, we consider a parameter to be inactive when its value drops below $\num{1.0e-4}$). In contrast, all network parameters $\btheta$ remain active while optimizing the objective without the ARD prior.
Apart from the qualitative advantage, the sparsity-inducing prior provides a strong regularization in the presence of limited data and yields superior predictive estimates.
In addition to obtaining sparse solutions, the  ARD prior facilitates the identification of physically meaningful latent representations for limited data (e.g. $N=50$) as shown in~\reffig{fig:aevb_n_50_compare_latent}.
Without the ARD prior, the data is encoded in a rather small region of the latent space. 

\begin{figure}[h]
	\centering
	\includegraphics[width=0.9\textwidth]{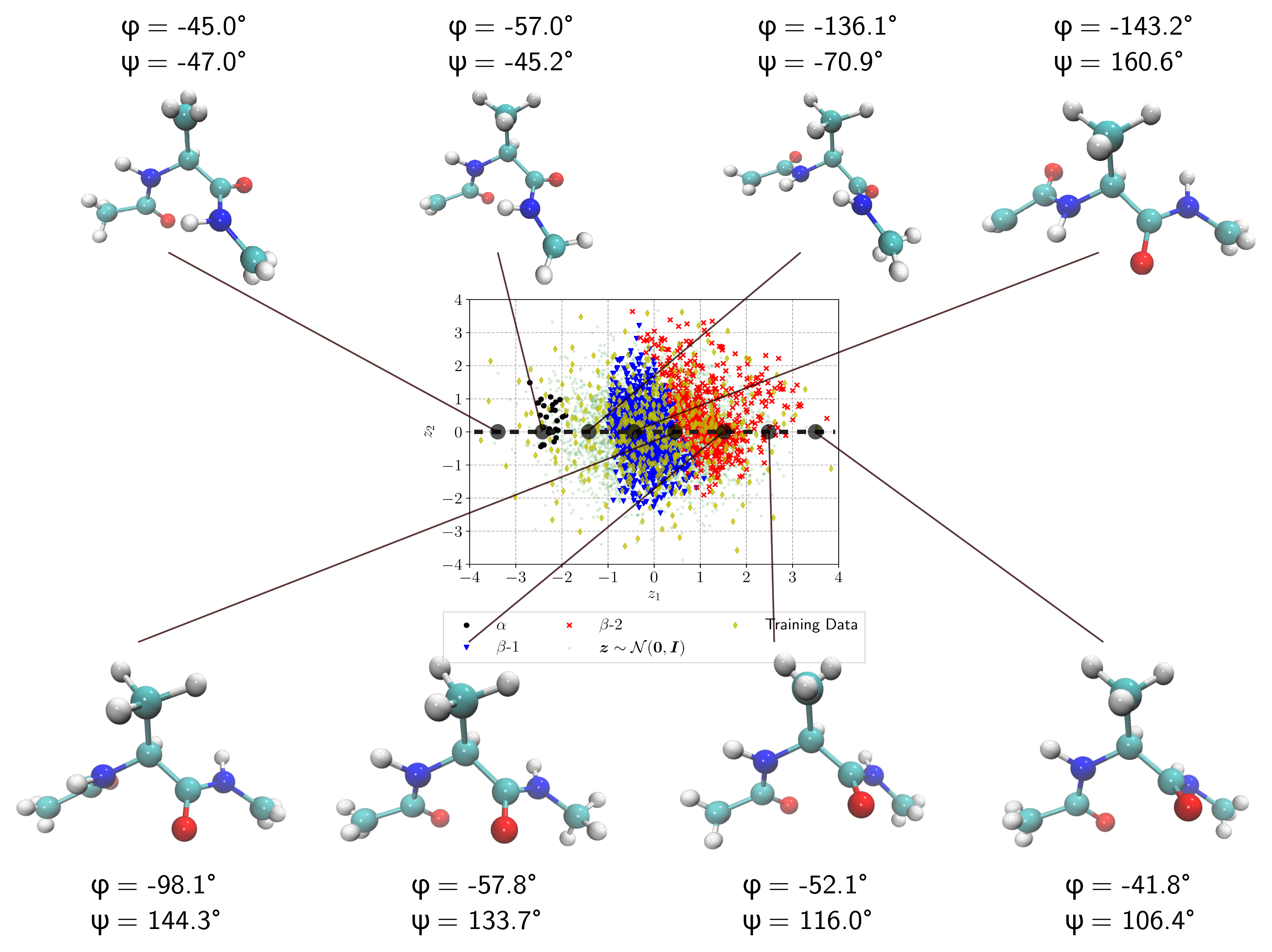}
	\caption{\emph{Predicted} configurations $\bx$ (including dihedral angle values) for $\{\bz|z_1=\{-3.5, -2.5, \dots, 3.5\}, z_2=0\}$ with $\bmu\inth(\bz)$ of $p\inth(\bx|\bz)$. As one moves along the $z_1$ axis, we obtain for the given CVs atomistic configurations $\bx$ reflecting the conformations $\alpha$, $\beta\textnormal{-}1$, and $\beta\textnormal{-}2$.
	All rendered atomistic representations in this work are created by VMD~\cite{humphrey1996}.}
	\label{fig:aevb_lat_rep_prediction}
\end{figure}

\begin{figure}
	\centering
	\subfigure[~Mean prediction $\bmu\inth(\bz^0)$ for a sample $\bz^0 \sim p(\bz)$.]{
		\label{fig:aevb_vis_mean}
		\includegraphics[width=0.3\textwidth]{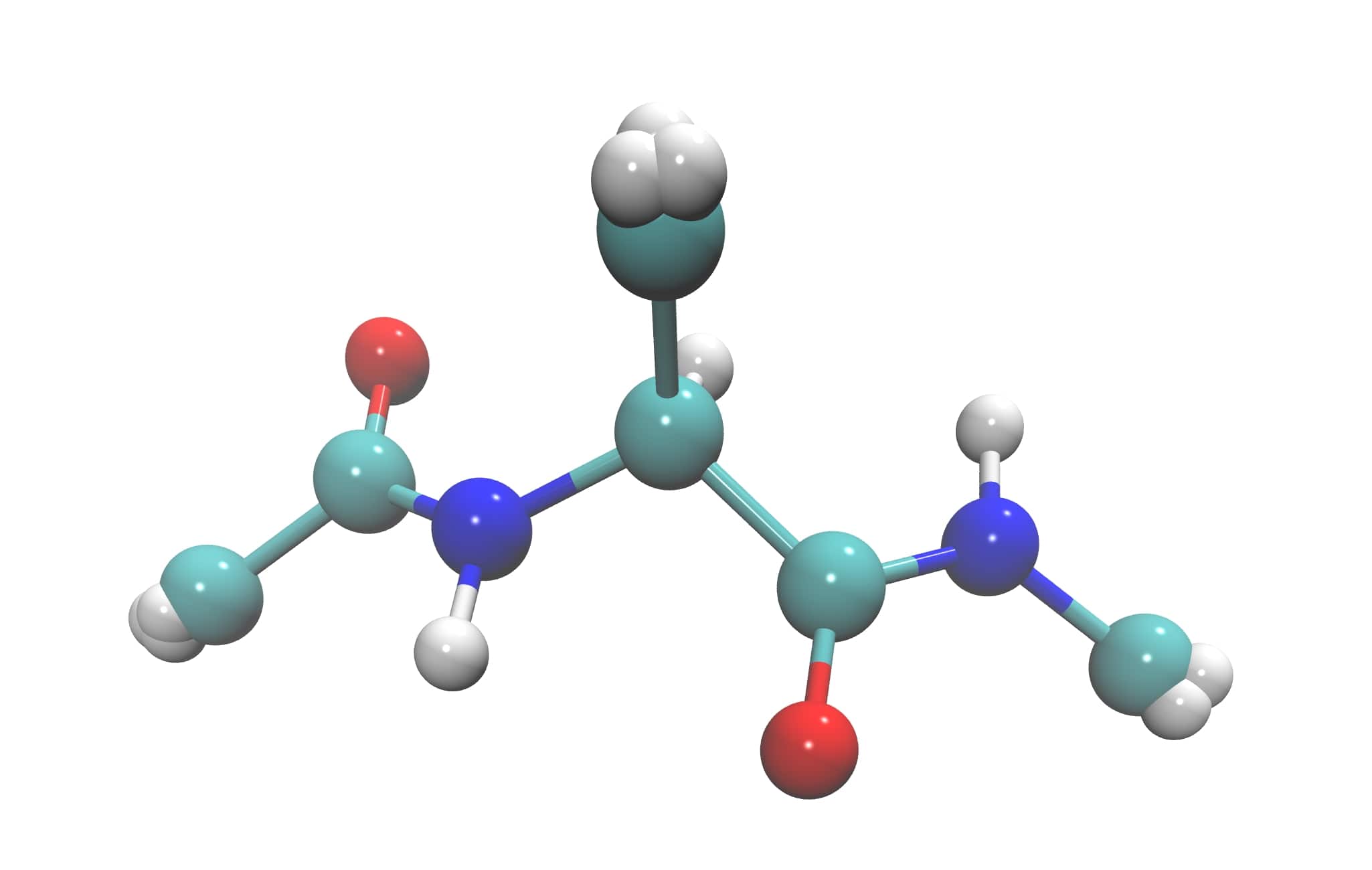}}
	\qquad
	
	\subfigure[~Realization $\bx^{0,0} \sim p\inth(\bx|\bmu\inth(\bz^0), \bm{S}\inth=\mathrm{diag}(\bsig\inth^2 ))$.]{
		\label{fig:aevb_vis_rep_1}
		\includegraphics[width=0.3\textwidth]{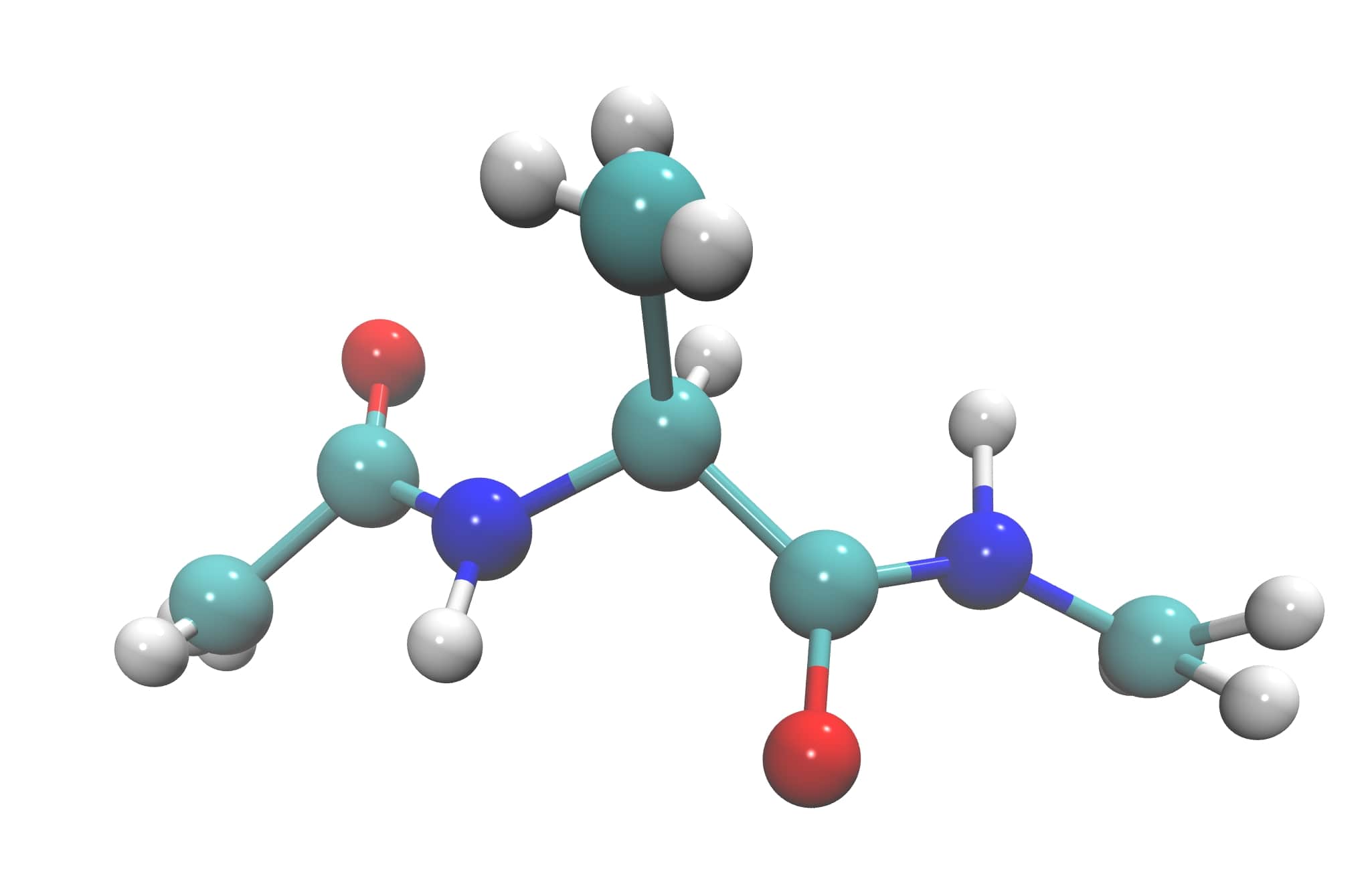}}
	%\qquad
	\subfigure[~Realization $\bx^{1,0} \sim p\inth(\bx|\bmu\inth(\bz^0), \bm{S}\inth=\mathrm{diag}(\bsig\inth^2))$.]{
		\label{fig:aevb_vis_rep_2}
		\includegraphics[width=0.3\textwidth]{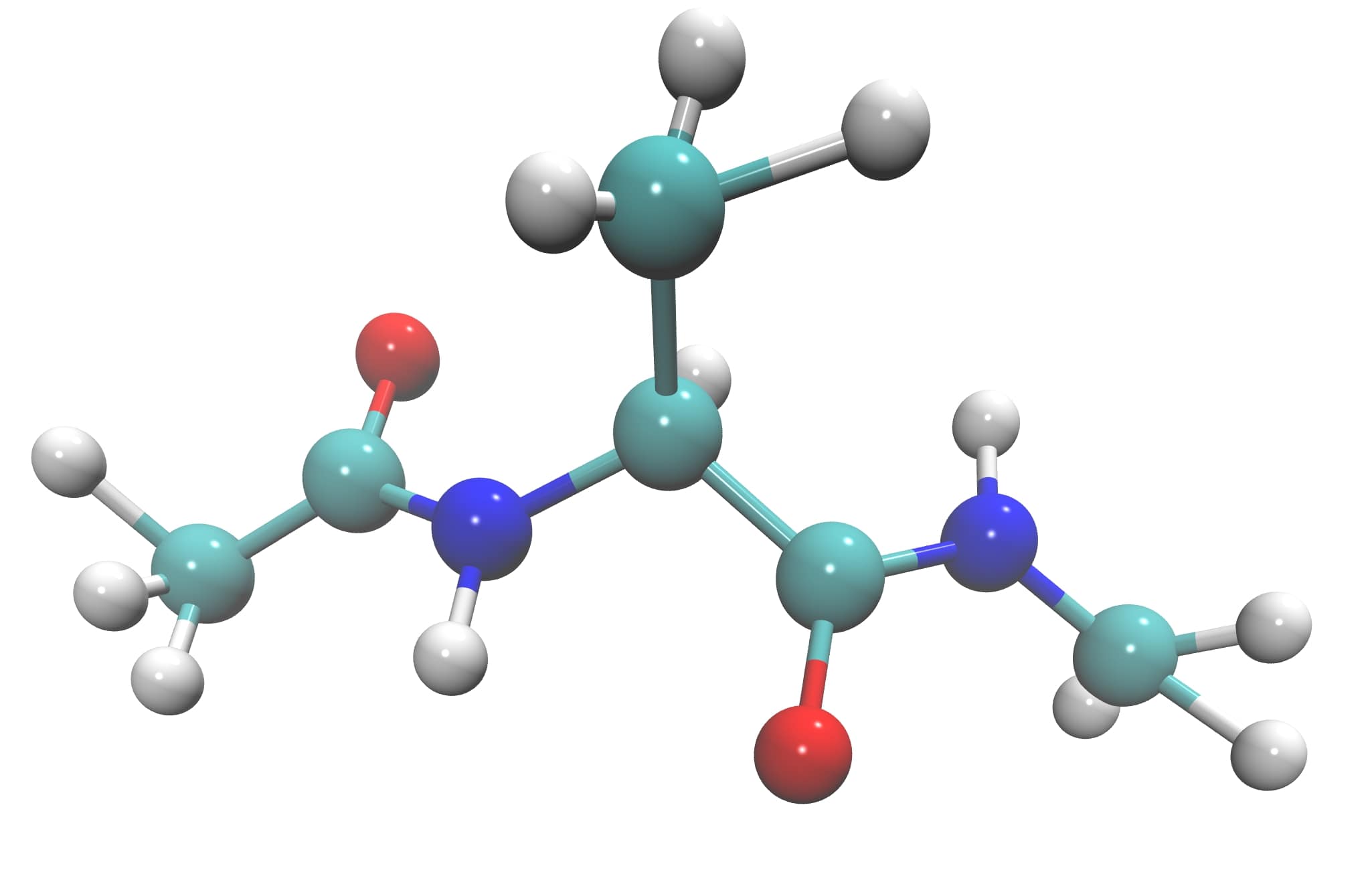}}
	%\qquad
	\subfigure[~Realization $\bx^{2,0} \sim p\inth(\bx|\bmu\inth(\bz^0), \bm{S}\inth=\mathrm{diag}(\bsig\inth^2))$.]{
		\label{fig:aevb_vis_rep_3}
		\includegraphics[width=0.3\textwidth]{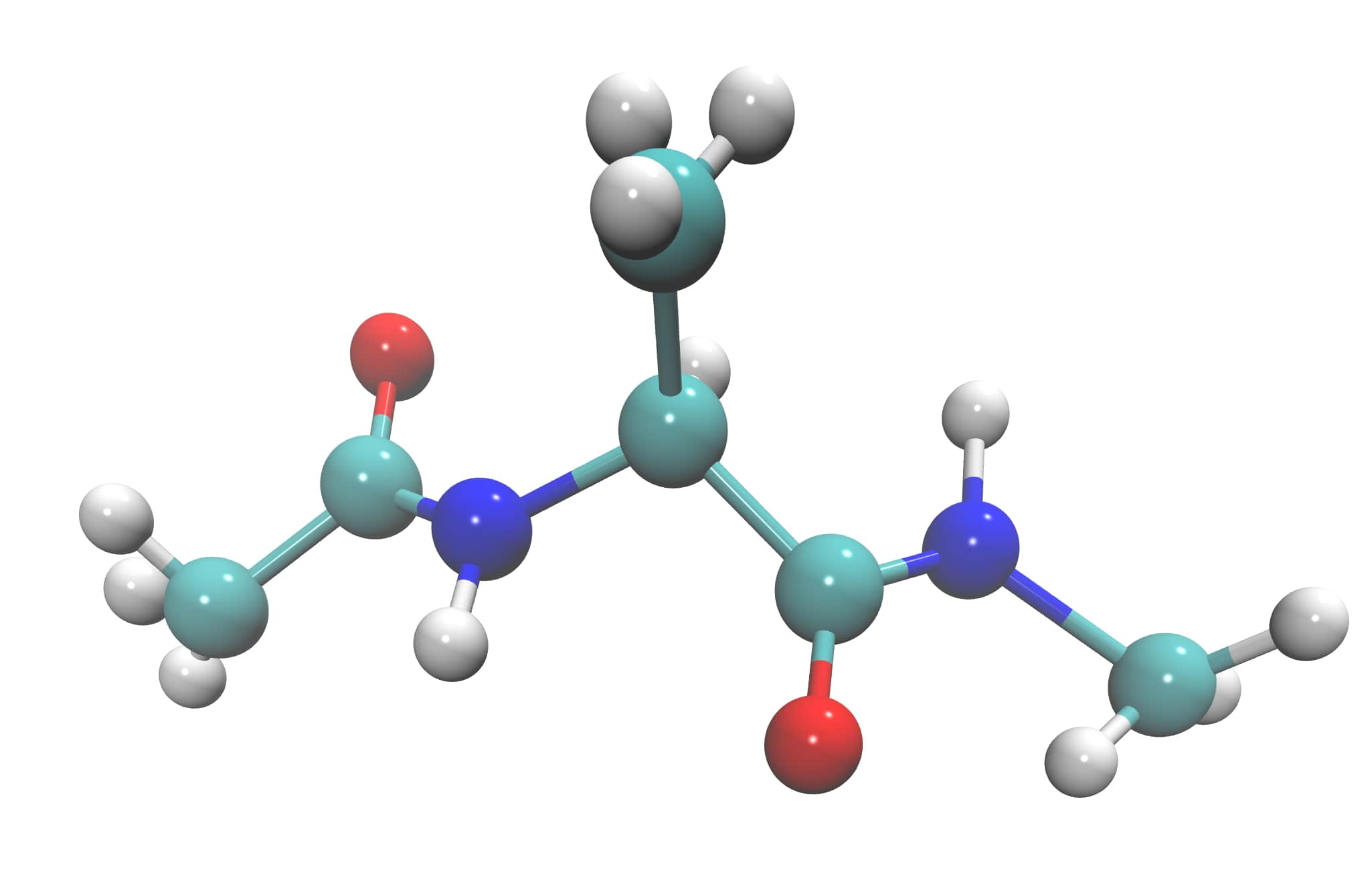}}
	%\qquad
	\caption{Visualization of the mean prediction (a) for a sample $\bz^0 \sim p(\bz)$, obtained from the decoding network $\bmu\inth(\bz^0) = f\inth(\bz^0)$, and realizations (b-d) $\bx^{j,0} \sim p\inth(\bx|\bz^0)$. Less relevant positions of the outer Hydrogen atoms are captured by the noise $\bsig\inth$ of the model $p\inth(\bx|\bz^0) = \mathcal N(\bmu(\bz\inth^0), \bm{S}\inth=\mathrm{diag}(\bsig\inth^2))$.}
	\label{fig:aevb_vis_mean_predictions}
\end{figure}

In~\reffig{fig:aevb_lat_rep_prediction}, we attempt to provide insight on the physical meaning of the CVs $\bz$ identified. In particular, we plot  the  atomistic configurations $\bx$ corresponding to various values of the first CV $z_1$ while keeping $z_2=0$.  The conformational transition in predicted atomistic configurations can be clearly recognized in the peptides of~\reffig{fig:aevb_lat_rep_prediction}.
We note that we start on the left ($z_1<0$) with $\alpha$ configurations, then move towards $\beta\textnormal{-}1$ (starting at $z_1 \approx -1$), and finally obtain $\beta\textnormal{-}2$ configurations.
For illustration purposes, the predictions in ~\reffig{fig:aevb_lat_rep_prediction} are based solely on the mean  $\bmu\inth(\bz)$ of the probabilistic decoder $p\inth(\bx|\bz) = \mathcal N (\bx; \bmu\inth(\bz), \bm{S}\inth=\mathrm{diag}(\bsig\inth^2) )$. We note that for each value of the CVs $\bz$ several atomistic realizations $\bx$ can be drawn from $p\inth(\bx|\bz)$  as depicted in~\reffig{fig:aevb_vis_mean_predictions}. This figure reveals the characteristic and relevant movement of the backbone that is captured by the predictive mean $\bmu\inth(\bz)= f^{\bmu}\inth(\bz)$.
Fluctuations of less relevant outer Hydrogen atoms (see Figs.~\ref{fig:aevb_vis_rep_1}-\ref{fig:aevb_vis_rep_3}) are recognized as noise of the decoder $p\inth(\bx|\bz) = \mathcal N\left(\bmu(\bz), \bm{S}\inth=\mathrm{diag}(\bsig\inth^2) \right)$ denoted in~\refeqq{eqn:map}. We note also that the corresponding entries of $\bsig\inth$ responsible for the outer Hydrogen atoms are five times larger compared to the remaining atoms.  The proposed model can therefore in unsupervised fashion identify the central role of the backbone coordinates whereas this physical insight is pre-assumed in~[\onlinecite{noe2018,chen2017}].

\begin{figure}[h]
	\centering
	\includegraphics[width=0.6\textwidth]{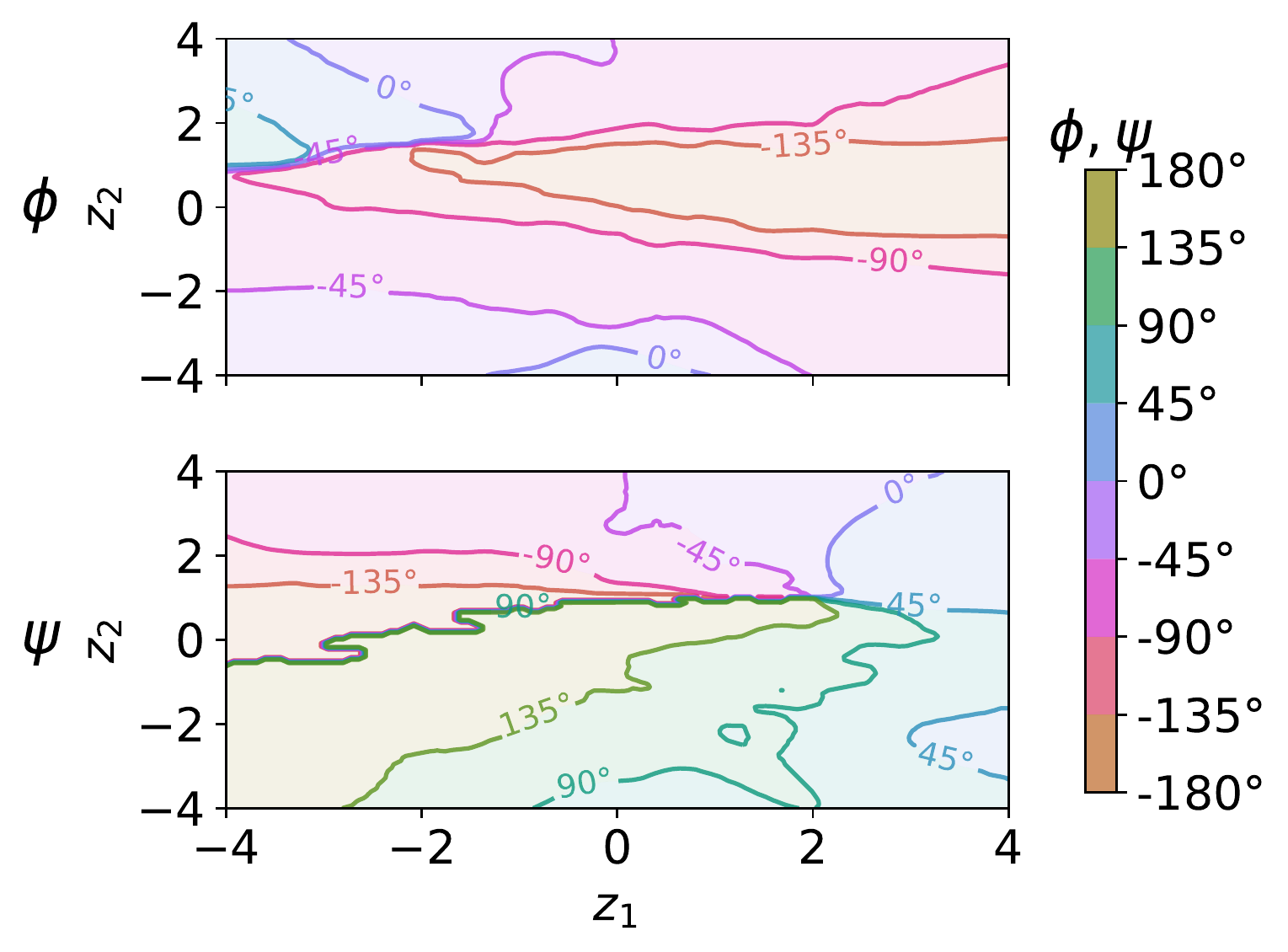}
	\caption{Predicted dihedral angles $(\phi, \psi)$ given the latent variables $\bz \in [ -4, 4 {]^2} $.}
	\label{fig:aevb_phi_psi}
\end{figure}

\begin{figure}[h]
	\centering
	\subfigure[~]{%
		\label{fig:aevb_scatter_phipsi_z1}%
		\includegraphics[width=0.45\textwidth]{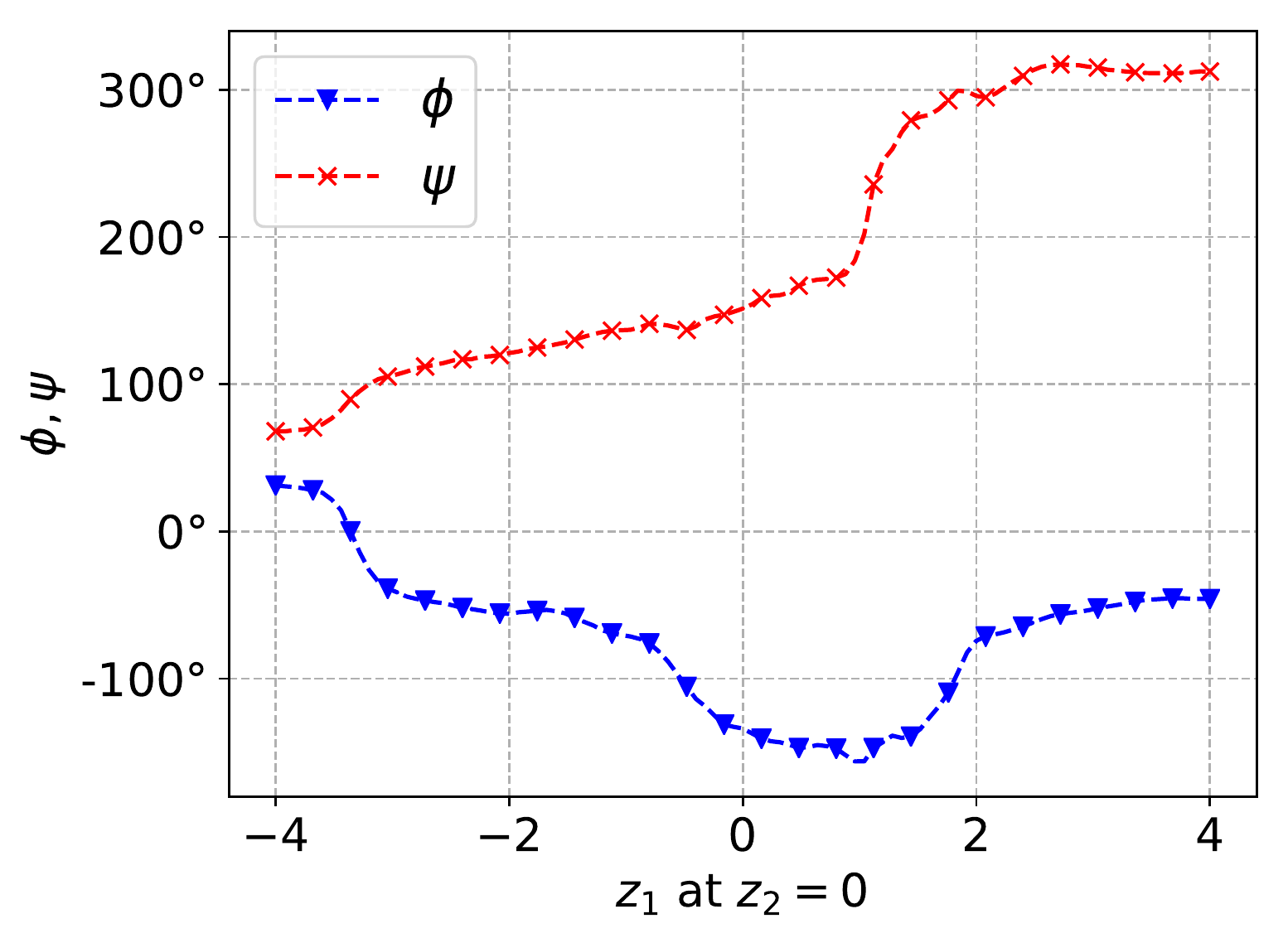}}%
	\qquad
	\subfigure[~]{%
		\label{fig:aevb_scatter_phipsi_z2}%
		\includegraphics[width=0.45\textwidth]{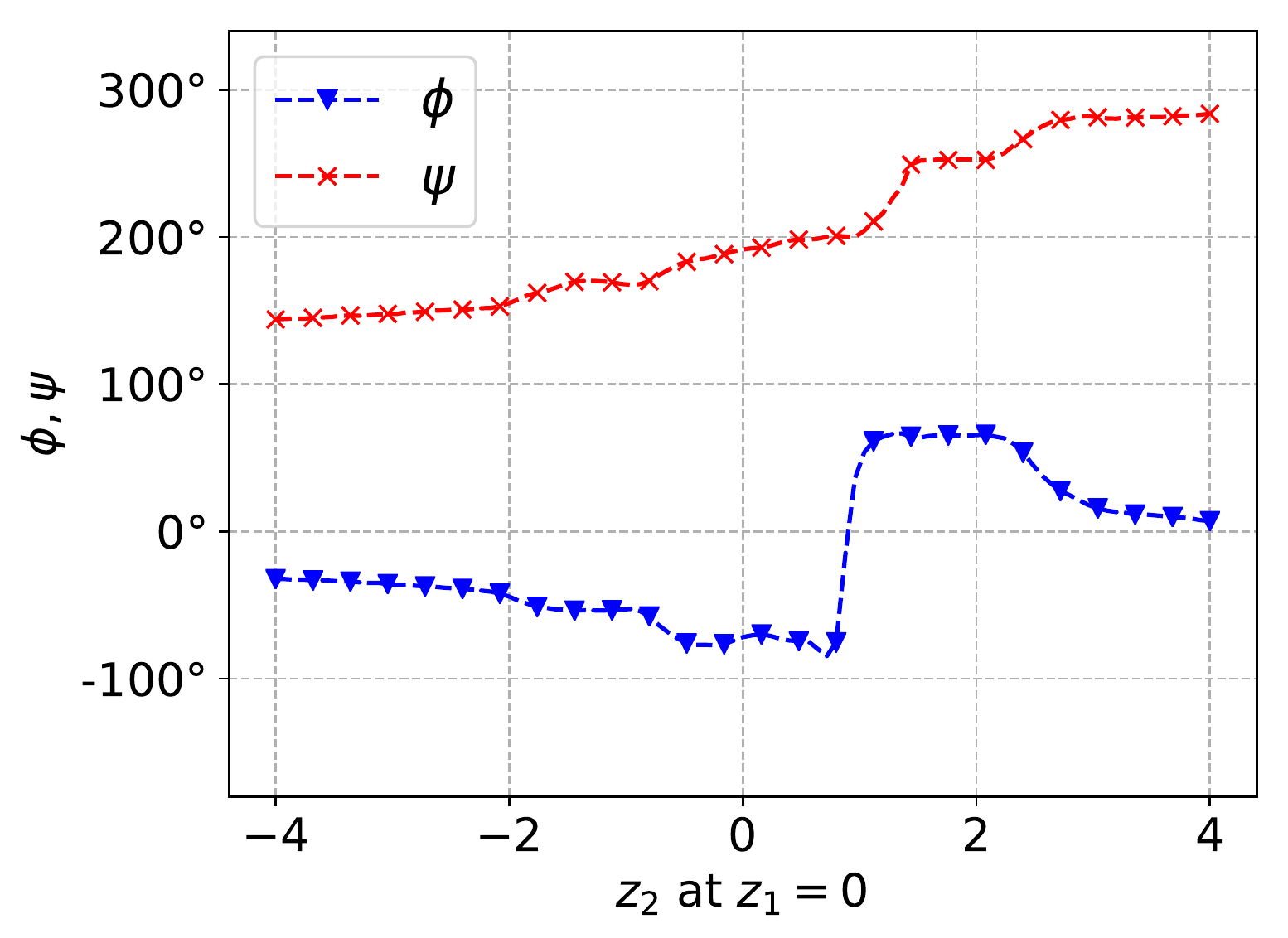}}%
	\qquad
	\caption{Predicted dihedral angles $(\phi, \psi)$ given the latent variables (a) $\{z_1, z_2 | z_1 \in [-4,4], z_2=0\}$ and (b) $\{z_1, z_2 | z_1=0, z_2 \in [-4,4] \}$.}
	\label{fig:aevb_cut_latent_phi_psi}
\end{figure}

In order to gain further insight into the relation between the dihedral angles $\phi,\psi$ and the discovered CVs $\bz$, we plot in Figs.~\ref{fig:aevb_phi_psi} and~\ref{fig:aevb_cut_latent_phi_psi} the corresponding maps for various combinations of $\bz$-values.
While it is clear that the map is not always bijective, the figures reveal the strong correlation between the two sets of variables. It should also be noted that in contrast to the dihedral angles, the $\bz$ value for a given atomistic configuration $\bx$ are not unique but rather  there is a whole distribution   as implied by $q\inphi(\bz|\bx)$. For the aforementioned plots we computed the $\bz$ from the mean of this density, i.e. $\boldsymbol{\mu}\inphi(\bx)$.

The trained model can also be employed in computing predictive estimates of observables $\int a(\bx)~p_{target}(\bx)~d \bx$ by making use of $p\inth(\bx)$ and samples drawn from it as described in Section~\ref{sec:pred}.
We illustrate this by computing the radius of gyration (Rg)~\cite{fluitt2015, shell2012} given as,
\be
a_{\text{Rg}}(\bx) = \sqrt{\frac{\sum_p m_p ||\bx_p - \bx_{\text{COM}}||^2}{\sum_p m_p}}.
\label{eqn:observable_rg}
\ee
The sum in~\refeqq{eqn:observable_rg} considers all atoms $p=1,\ldots, P$ of the peptide, where $m_p$ and $\bx_p$ denote the mass and the coordinates of each atom, respectively. $\bx_{\text{COM}}$ denotes the center of mass of the peptide. The histogram of $a_{\text{Rg}}(\bx)$ reflects the distribution of the size of the peptide and is correlated with the various conformations~\cite{fluitt2015}.

\begin{figure}[h]
	\centering
	\subfigure[~$N=50$.]{
		\label{fig:aevb_rg_526}
		\includegraphics[width=0.3\textwidth]{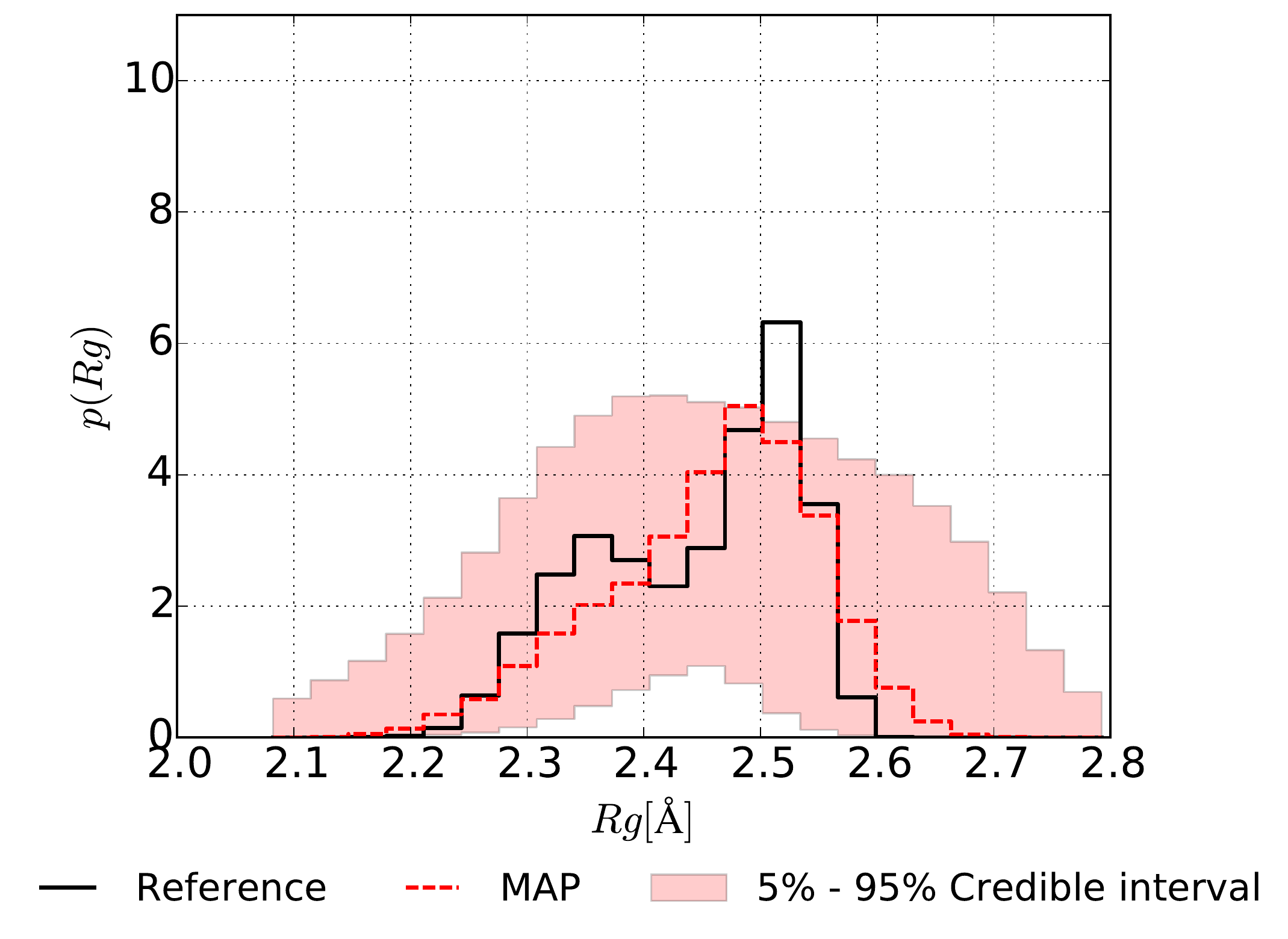}}
	%\qquad
	\subfigure[~$N=200$.]{
		\label{fig:aevb_rg_1527}
		\includegraphics[width=0.3\textwidth]{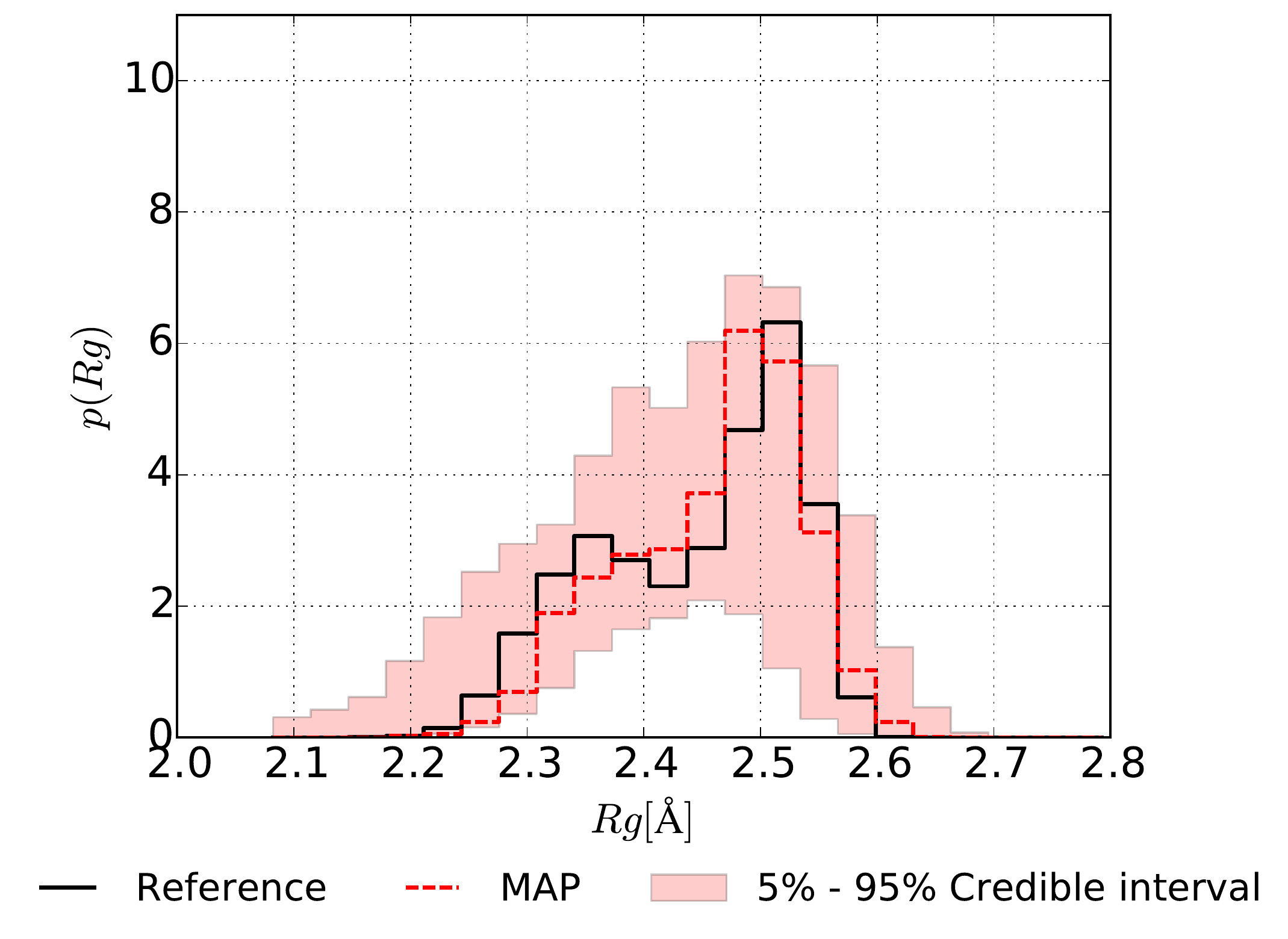}}
	%\qquad
	\subfigure[~$N=500$.]{
		\label{fig:aevb_rg_4004}
		\includegraphics[width=0.3\textwidth]{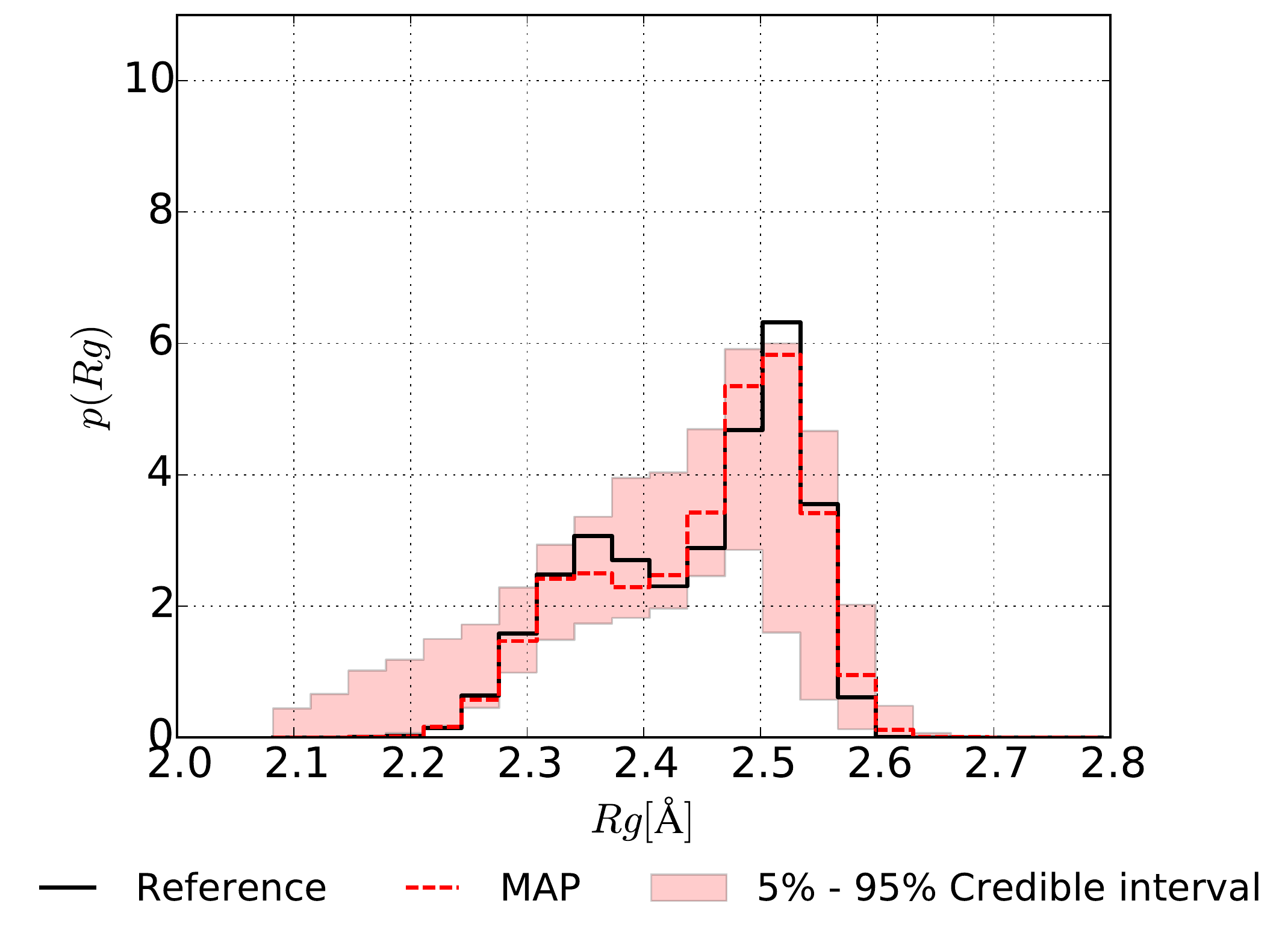}}
	%\qquad
	\caption{Predicted radius of gyration with $\dim(\bz)= 2$ for various sizes $N$ of the training dataset. The MAP estimate indicated in red is compared to the reference (black) solution. The latter is estimated by $N=\num{10000}$. The shaded area represents  the 5\%-95\% credible interval, reflecting the induced epistemic uncertainty from the limited amount of training data.}
	\label{fig:aevb_prediction_rg}
\end{figure}

In the estimates that we depict in~\reffig{fig:aevb_prediction_rg}, we have also employed the posterior approximation of the model parameters $\btheta$ obtained as described in Section~\ref{sec::approxBayInf} in order to compute credible intervals for the observable. These credible intervals are estimated as  described in Algorithm \ref{alg:quantile_estimation} utilizing $J=\num{3000}$ samples.
We observe that the model's predictive confidence increases with the size of the training data. This is reflected in shrinking credible intervals in~\reffig{fig:aevb_prediction_rg} for increasing $N$.

\begin{algorithm}[H]
	\caption{\small Estimating Credible Intervals.}
	\begin{algorithmic}%[1]
	%\SetAlgoLined
		%\algsetup{linenosize=\tiny}
		\small
		\INPUT{$J$ the number of samples to be drawn, optimal values of $\btheta=\btheta_{MAP}$ and $\bphi = \bphi_{MAP}$.}
		\State Compute Laplace's approximation $\mathcal N(\boldsymbol{\mu}_L, \boldsymbol{S}_L= \mathrm{diag}(\bsig_L^2))$ to  the  posterior $p(\btheta|\bX)$ (\refeqq{eqn:laplaceposterior}).
		\FOR{$j=1$ \textbf{to} J }
		\State Draw a posterior sample: $\btheta^j \sim \mathcal N(\boldsymbol{\mu}_L, \boldsymbol{S}_L = \mathrm{diag}(\bsig_L^2))$.
		\State Obtain a predictive trajectory $\bar{\bx}_{1:T}^j$, given the parametrization $\btheta^j$ utilizing Algorithm~\ref{alg:metropolisgibbs}.
		\State Estimate the observable $\hat{a}(\btheta^j) = \frac{1}{T}\sum_{t=1}^T a(\bar{\bx}^j_t)$, given the trajectory $\bar{\bx}_{1:T}^j$.
		\ENDFOR
		\State Estimate the desired quantiles with $\hat{a}(\btheta^{1:J})$.
	\end{algorithmic}
	\label{alg:quantile_estimation}
\end{algorithm}

In summary for ALA-2, we note that the proposed methodology for identifying CVs (\reffig{fig:aevb_lat_rep}) and predicting observables~(Figs.~\ref{fig:aevb_predictive_ability_sparse_low_data} and~\ref{fig:aevb_prediction_rg})  works well with small size datasets, e.g.  $N=\{50, 200, 500\}$. 

\subsection{ALA-15}

\subsubsection{Simulation of ALA-15 and model specification}

The following example considers a larger alanine peptide with $15$ residues, ALA-15 which consists of  $162$ atoms giving rise to  $\dim(\bx)=486$ with $480$ DOF. The reference dataset $\bX$ has been obtained in a similar manner as specified in Section~\ref{sec:numill_sim_ala2_sim_details} with the only difference  being that we  utilize a replica-exchange molecular dynamics~\cite{sugita1999} algorithm with $21$ temperature replicas distributed according to $T_i = T_0 e^{\kappa \cdot i}$ ($T_0 = \SI{270}{K}$, and $\kappa = 0.04$). This leads to an analogous simulation setting as employed in~[\onlinecite{shell2012}]. The datasets are obtained as mentioned in the previous example. We consider here $N=\{300, \num{3000}, \num{5000}\}$. Using the same model specifications as  in Section~\ref{sec:numill_sim_ala2_model_spec}, we present next a summary of the obtained results.
 
\subsubsection{Results}

For visualization purposes of the latent CV space, we assumed $\dim(\bz)=2$ in the following, even though  the presence of 15 residues each requiring a pair of dihedral angles $(\phi, \psi)$ would potentially suggest a higher-dimensional representation.
However, when considering test cases with  $\dim(\bz)=\{15,30\}$, no significant differences were observed in the predictive capabilities. This is in agreement with~[\onlinecite{marini2011}] where it is argued based on density functional theory calculations that not all dihedral angles are equally relevant.
The $(\phi, \psi)$ pairs within a peptide chain show high correlation. Mulitlayer neural networks provide the capability of transforming independent CVs (as considered in this study) to correlated ones by passing them through the subsequent network layers. This explains the reasonable predictive quality of the model using independent and low-dimensional CVs with $\dim(\bz)=2$. Considering more expressive $p\inth(\bz)$ than the standard Gaussian employed, could have accounted (in part) for such correlations. In this example, by employing the  ARD prior, only  $43$\% of the decoder parameters $\btheta$ remained effective.

\begin{figure}[h]
    \centering
    \includegraphics[width=0.9\textwidth]{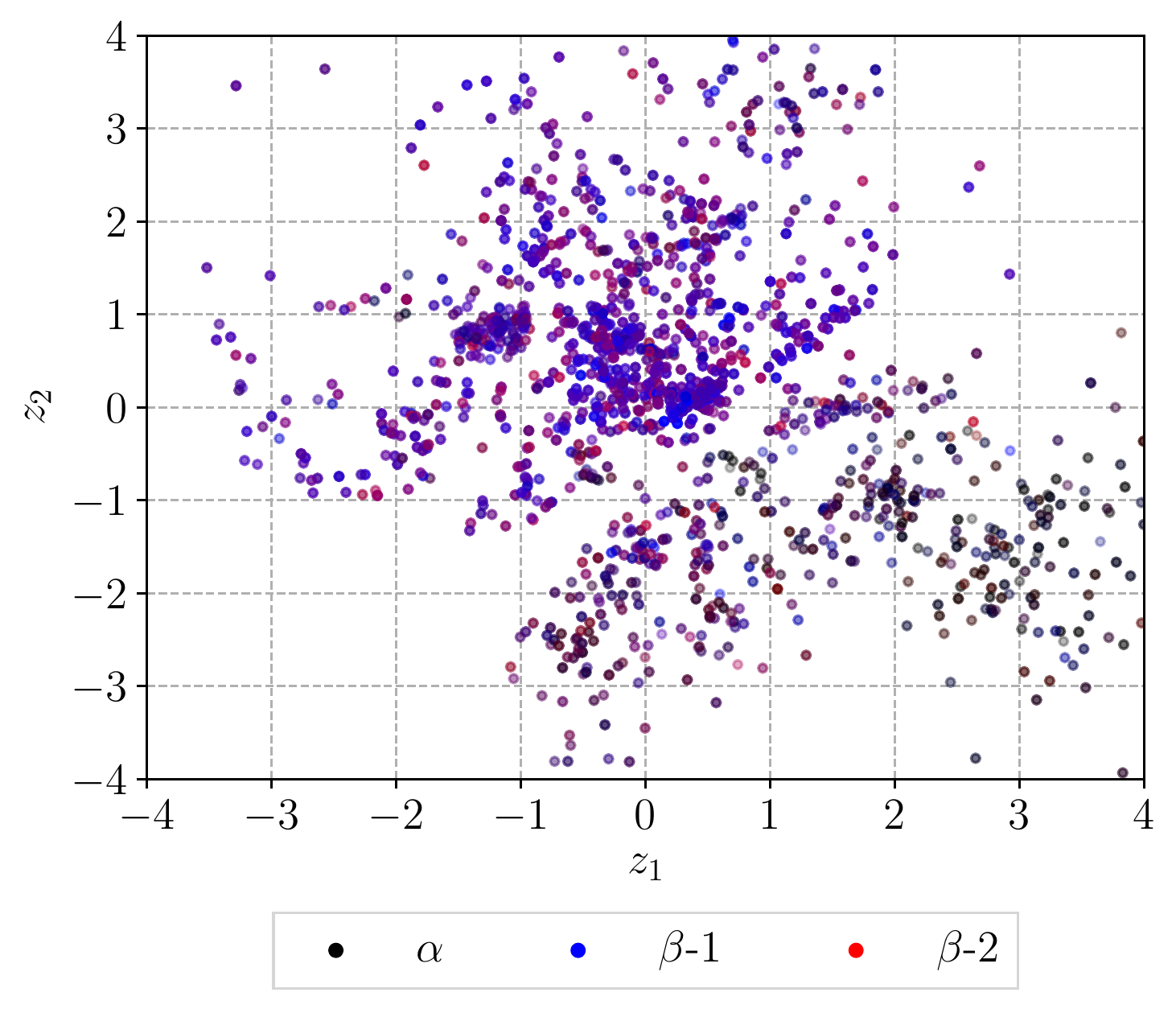}
    \caption{Representation of the training data $\bX$ with $N=\num{3000}$ in the encoded collective variable space.
	The inferred approximate posterior $q\inphi(\bz|\bx)$ of the latent CVs separates residues mostly belonging to the $\beta$ conformations (mixture of red and blue) and peptide configurations containing largely residues in the $\alpha$ configuration (black).
	Here, the mean $\bmu\inphi(\bx)$ of the approximate posterior $q\inphi(\bz|\bx)= \mathcal N(\bx;\bmu\inphi(\bx), \bm{S}\inphi=\mathrm{diag}(\bsig\inphi^2(\bx)))$ is depicted.}
	\label{fig:aevb_ala_15_lat_rep}
\end{figure}

Figure~\ref{fig:aevb_ala_15_lat_rep} depicts the posterior means of the $N=\num{3000}$ training data in the CV space $\bz$.
Given that a peptide configuration contains residues from different conformations labelled here as $\alpha$, $\beta\textnormal{-}1$, and $\beta\textnormal{-}2$ and residues in intermediate $(\phi, \psi)$ states,  we applied the following rule for labelling/coloring each datapoint. The assigned color in~\reffig{fig:aevb_ala_15_lat_rep} is a mixture between the RGB colors black (for $\alpha$), blue (for $\beta\textnormal{-}1$), and red (for $\beta\textnormal{-}2$). The mixture weights of the assigned color are proportional to the number of residues belonging to the $\alpha$ (black), $\beta\textnormal{-}1$ (blue), and $\beta\textnormal{-}2$ (red) conformations, normalized by the total amount of residues which can be clearly assigned to $\alpha$, $\beta\textnormal{-}1$, and $\beta\textnormal{-}2$.
Additionally, we visualize the amount of intermediate $(\phi, \psi)$ states of the residues by the opacity of the scatter points. The opacity reflects the amount of residues which are clearly assigned to the $\alpha$, $\beta\textnormal{-}1$, and $\beta\textnormal{-}2$ conformations compared to the total amount of residues in the peptide. For example, if all residues of a peptide configuration correspond to a specific mode, the opacity is taken as $100$\%. If all residues are in non-classified intermediate states, the opacity is set to the minimal value which is here taken as $20$\%.

We note that  peptide configurations  in which the majority of residues belong to $\beta\textnormal{-}1$  (blue) or in the $\beta\textnormal{-}2$ conformation (red), are clearly separated in the CV space from datapoints with residues   predominantly in the $\alpha$ conformation (black).
Nevertheless, we observe that the encoder has difficulties separating blue ($\beta\textnormal{-}1$) and red ($\beta\textnormal{-}2$) datapoints. 
We remark though that the related secondary structures~\cite{zhou2016} resulting from the assembly of residues in $\beta\textnormal{-}1$ and $\beta\textnormal{-}2$, such as the  $\beta$-sheet and $\beta$-hairpin, share a similar atomistic representation $\bx$ which  explains the similarity in the CV space.

\begin{figure}[h]
	\centering
	\includegraphics[width=0.7\textwidth]{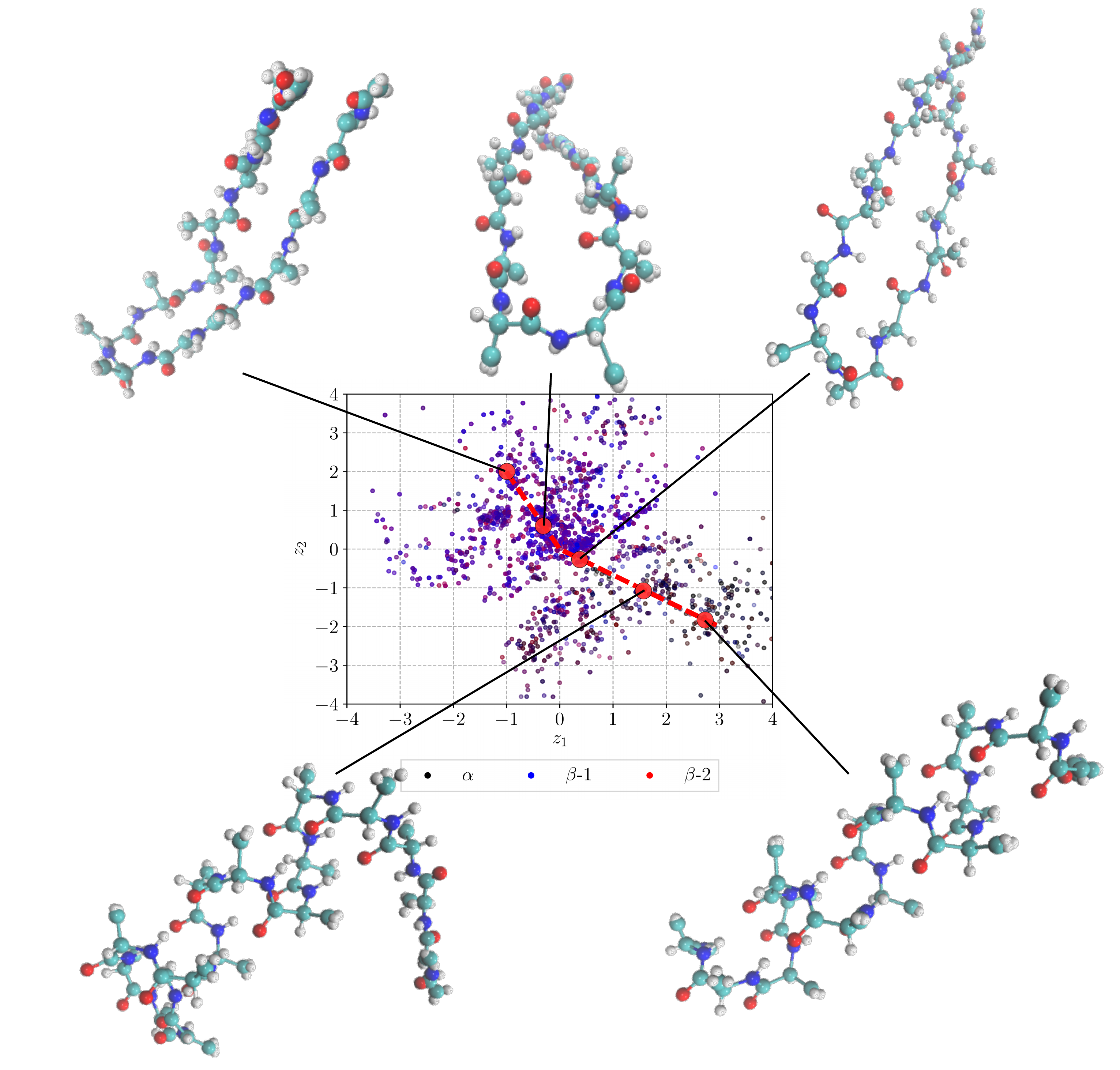}
	%\qquad
	\caption{
	\emph{Predicted} configurations $\bx$ for decoding CVs indicated as red points on the dashed line in the plot. Depicted configurations have been produced by evaluating the mean $\bmu\inth(\bz)$ of $p\inth(\bx|\bz)$.
	Moving along the path, we obtain atomistic configurations $\bx$ partially consisting of the conformations $\alpha$, $\beta\textnormal{-}1$, and $\beta\textnormal{-}2$ in the ALA-15 peptide resulting into peptide secondary structures such as $\beta$-sheet (top left), $\beta$-hairpin (top middle and right), and $\alpha$-helix (bottom row).}
	\label{fig:aevb_ala_15_lat_rep_prediction}
\end{figure}

When one moves in the CV space $\bz$  along the path indicated by a red dashed line in~\reffig{fig:aevb_ala_15_lat_rep_prediction} and reconstructs the corresponding $\bx$ using the mean function of the decoder $p\inth(\bx|\bz)$, we obtain  atomistic configurations of the ALA-15 partially consisting of the conformations $\alpha$, $\beta\textnormal{-}1$, and $\beta\textnormal{-}2$ which correspond to   the aforementioned  secondary structures i.e.  $\beta$-sheet (top left), $\beta$-hairpin (top middle and right), and $\alpha$-helix (bottom row).

\begin{figure}[h]
    \centering
    \includegraphics[width=0.9\textwidth]{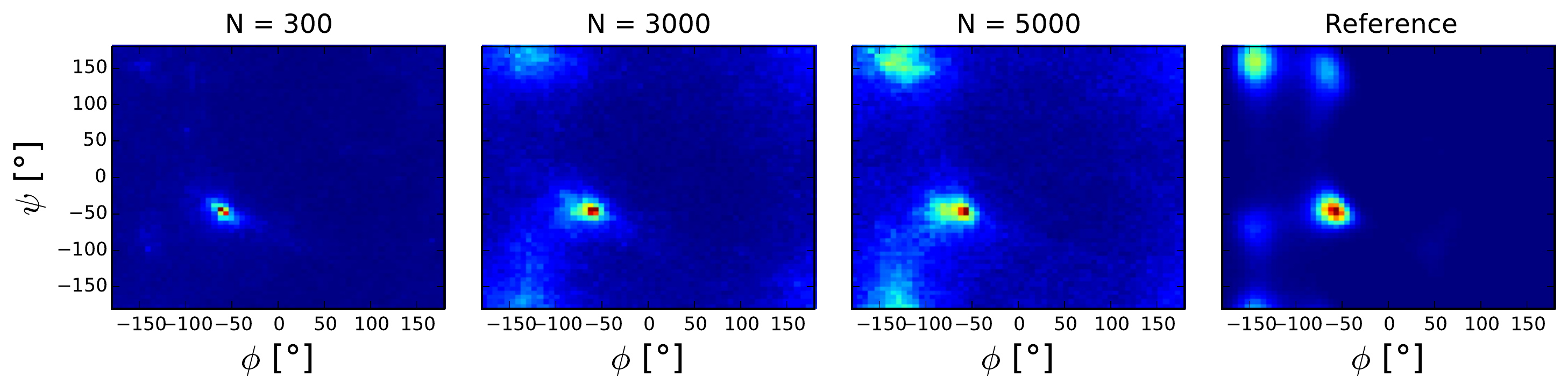}
    \caption{Predicted Ramachandran plots with $\dim(\bz)= 2$ for various sizes $N$ of the training dataset (first three plots from the left). Depicted predictions are MAP estimates based on $T=\num{10000}$ samples. The plot on the right is the reference MD prediction with $N=\num{10000}$ configurations.}
	\label{fig:aevb_ala_15_rama_compare}
\end{figure}

\begin{figure}[h]
	\centering
	\subfigure[~$N=300$.]{
		\label{fig:aevb_rg_ala_15_300}
		\includegraphics[width=0.3\textwidth]{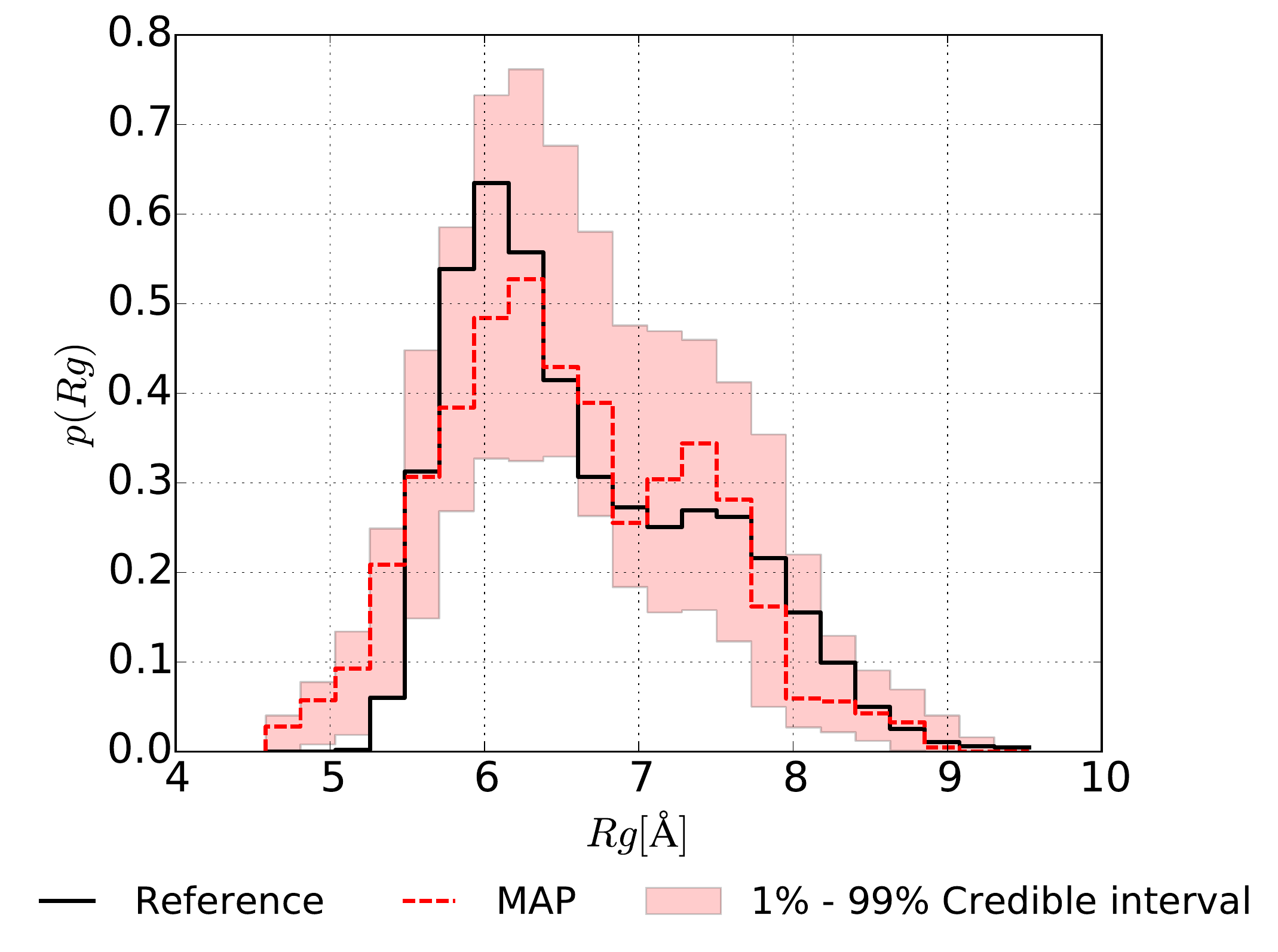}}
	%\qquad
	\subfigure[~$N=\num{3000}$.]{
		\label{fig:aevb_rg_ala_15_3000}
		\includegraphics[width=0.3\textwidth]{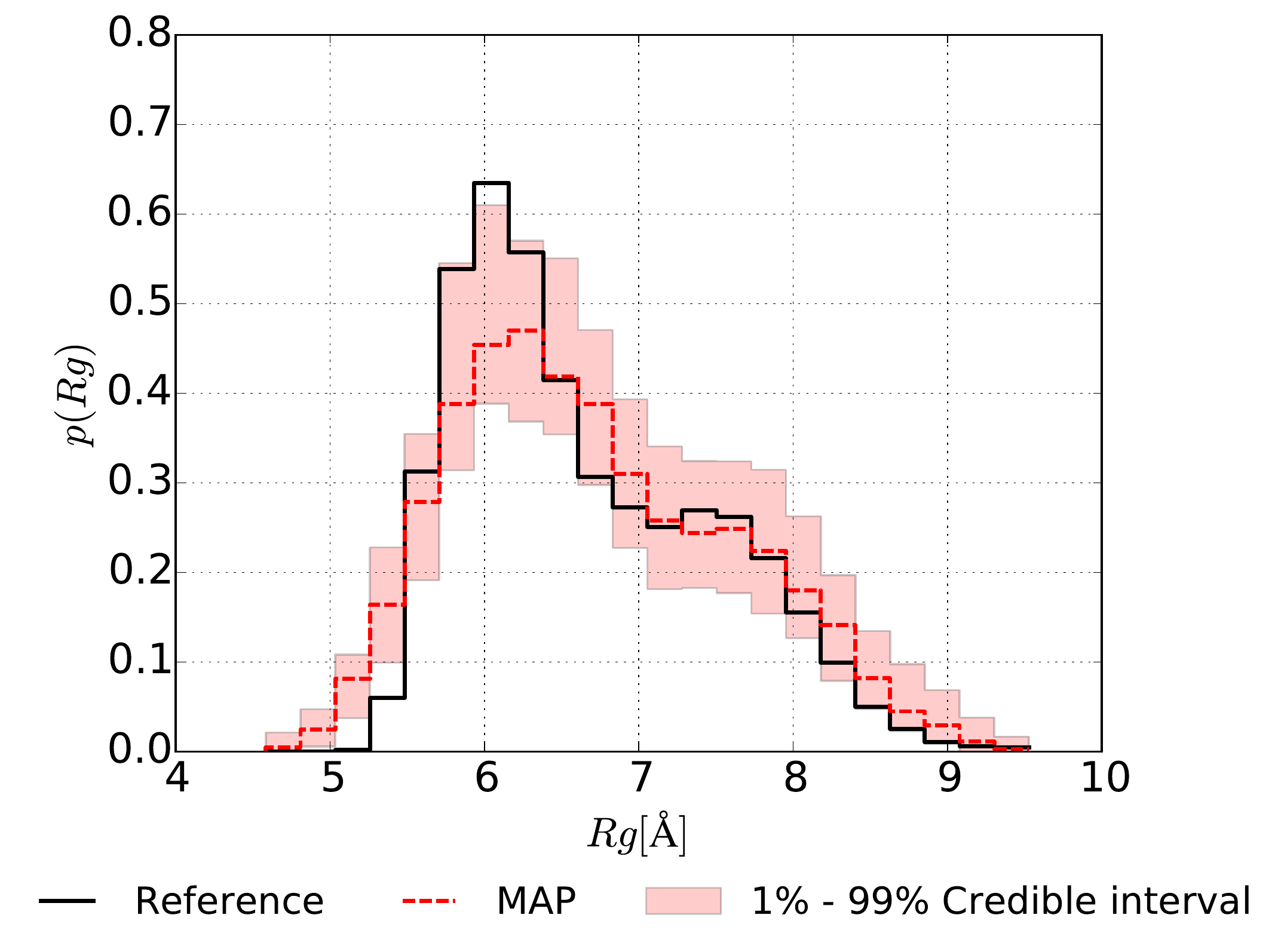}}
	%\qquad
	\subfigure[~$N=\num{5000}$.]{
		\label{fig:aevb_rg_ala_15_5000}
		\includegraphics[width=0.3\textwidth]{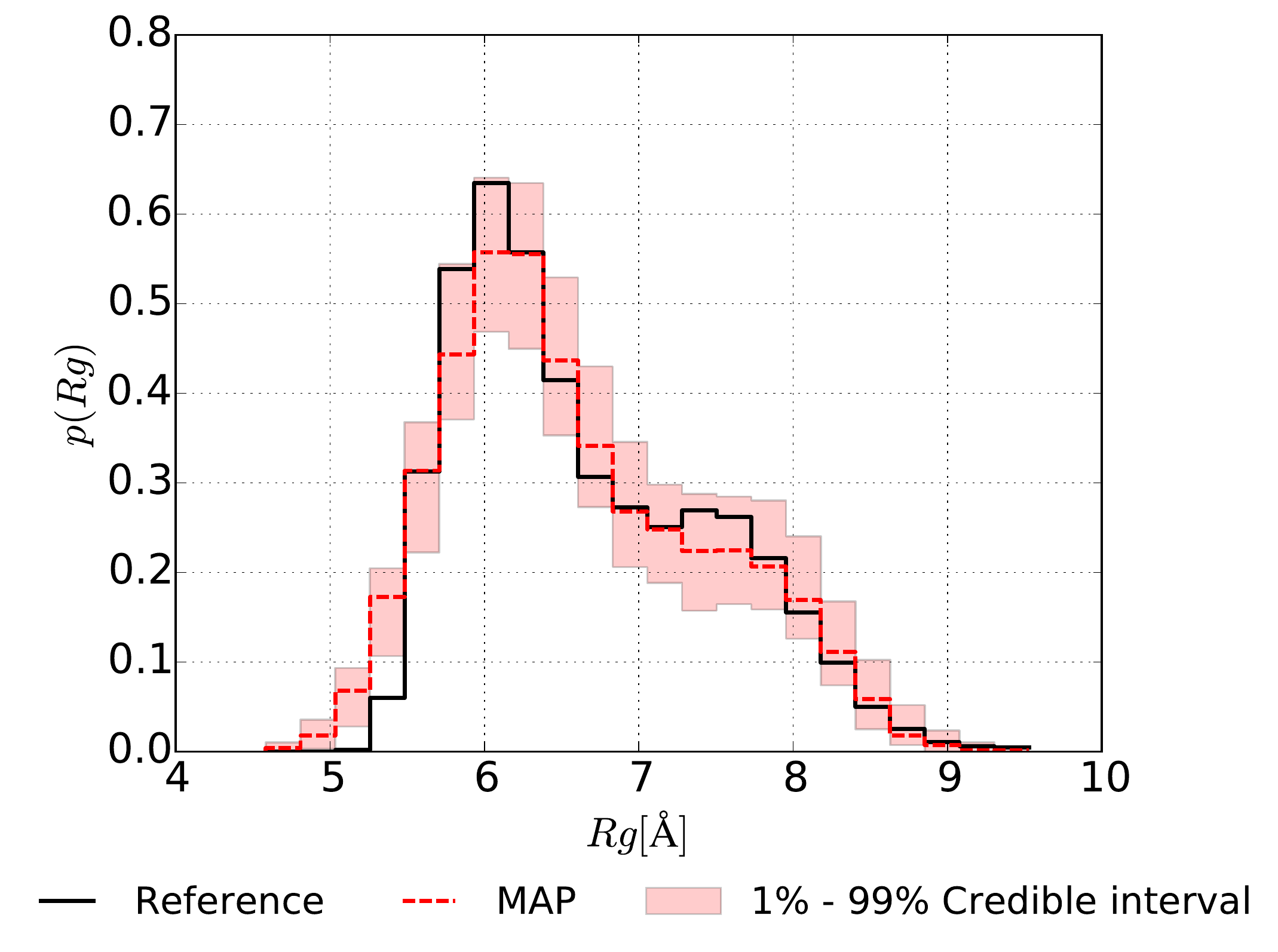}}
	%\qquad
	\caption{Predicted radius of gyration with $\dim(\bz)= 2$ for various sizes $N$ of the training dataset. The MAP estimate indicated in red is compared to the reference (black) solution. The latter is estimated by $N=\num{10000}$. The shaded area represents  the 1\%-99\% credible interval, reflecting the induced epistemic uncertainty from the limited amount of training data.}
	\label{fig:aevb_prediction_rg_ala_15}
\end{figure}

The ambiguity between  $\beta\textnormal{-}1$ and  $\beta\textnormal{-}2$ states  is also reflected in the  predicted Ramachandran plot in~\reffig{fig:aevb_ala_15_rama_compare}.
Nevertheless properties, independent of the explicit separation of configurations predominantly consisting of residues in $\beta\textnormal{-}1$ and $\beta\textnormal{-}2$ states, are  predicted  accurately by the framework. This is demonstrated with the computed radius of gyration  in~\reffig{fig:aevb_prediction_rg_ala_15}. The MAP estimate is complemented by the credible intervals which reflect the epistemic uncertainty and are able to envelop the reference profile. As in the previous example, the breadth of the credible intervals shrinks with increasing training data $N$.

% -------------------- CONCLUSIONS -----------------------

\section{Conclusions}
\label{sec:conclusions}
We presented an unsupervised learning  scheme for discovering CVs of atomistic systems. We defined   the CVs  as  latent generators of the atomistic configurations and formulated their identification as a Bayesian inference task. Inference of the posterior distribution of the latent CVs given the fine-scale atomistic training data identifies a probabilistic mapping from the space of  atomistic configurations to the latent space. This posterior distribution resembles a dictionary translating  atomistic configurations to the lower-dimensional CV space which is inferred during the training procedure. Compared to other dimensionality reduction methods, the proposed scheme is capable of performing well with comparably heterogeneous and small datasets.

We presented the capabilities of the model for the test case of an ALA-2 peptide (Section~\ref{sec:numericalillustrations}). When the dimensionality of the CVs $\dim(\bz)$ was set to $2$, the model discovered variables that correlate strongly with the widely known dihedral angles $(\phi, \psi)$. Other dimensionality reduction methods~\cite{hotelling1933, troyer1995, haerdle2007, tenenbaum2000,coifman2005, coifman2005b, nadler2006} rely on an objective keeping small distances between configurations in the atomistic space also small in the latent space.
Rather than enforcing this requirement directly, the proposed framework identifies a lower-dimensional representation that clusters configurations in the CV space which show similarities in the atomistic space.
The Bayesian formulation presented allows for a rigorous quantification of the unavoidable uncertainties and their propagation in the predicted observables. The ARD prior chosen was shown to lead to on average $45\%$ less parameters compared to the optimization without it.

We presented an approach for approximating the intractable posterior of the decoding model parameters~(\refeqq{eqn:laplaceposterior}) and provided an algorithm (Algorithm~\ref{alg:quantile_estimation}) for estimating credible intervals. The uncertainty propagated to the observables captures the parameter uncertainty of the decoding neural network $f\inth^{\bmu}(\bz)$. 

In addition to discovering CVs, the generative model employed is able to predict atomistic configurations by sampling the CV space with $p\inth(\bz)$ and mapping the CVs probabilistically via $p\inth(\bx|\bz)$ to full atomistic configurations.
We showed that the predictive mapping $p\inth(\bx|\bz)$ recognizes essential backbone behavior of the peptide while it models fluctuations of the outer Hydrogen atoms with the noise of $p\inth(\bx|\bz)$ (see~\reffig{fig:aevb_vis_mean_predictions}). We use the model for predicting observables and quantifying the uncertainty arising from limited training data.

We emphasize that the whole work was based on data represented by Cartesian coordinates $\bx$ of all the atoms of the ALA-2 ($\dim(\bx)=66$, and $60$ DOF adjusted by removing rigid-body motion) and ALA-15 ($\dim(\bx)=486$, and $480$ DOF adjusted by removing rigid-body motion) peptides. Considering a pre-processed dataset e.g. by considering solely coordinates of the backbone atoms, heavy atom positions, or a representation by dihedral angles assumes the availability of tremendous physical insight. The aim of this work was to reveal CVs with physicochemical meaning and the prediction of observables of complex systems without using any domain-specific physical notion.

Besides the framework proposed, generative adversarial networks (GANs)~\cite{goodfellow2014} and its Bayesian reformulation in~[\onlinecite{wilson2017}] open an additional promising avenue in the context of CV discovery and enhanced sampling of atomistic systems. GANs are accompanied by a two player (generator and discriminator) min-max objective which poses known difficulties in training the model. 
The training of GANs is not as robust as the VAE employed here and Bayesian formulations are not well studied. In addition, one needs to address the mode collapse issue (see~[\onlinecite{salimans2016}]) which is critical for atomistic systems.

Future work involves the use of the CVs discovered in the context of enhanced  sampling techniques that can lead to an accelerated exploration of the configurational space.
In addition to identifying good CVs, a crucial step for enhanced sampling methods is the biasing potential for lifting deep free-energy wells. In contrast  to the ideas e.g. presented in~[\onlinecite{chen2017}, \onlinecite{chen2018}, \onlinecite{gavelis}], we would advocate a formulation where the biasing potential is based on the lower-dimensional pre-image of the currently visited free-energy surface. To that end, we envision  using the posterior distribution $q_{\boldsymbol{\phi}}(\boldsymbol{z}|\boldsymbol{x})$ to construct a   locally optimal biasing potential defined in the CV space which gets updated on the fly as the simulations explore the configuration space. %This implicitly answers the critical question of the choice of the biasing potential. 
The biasing potential can be transformed by the probabilistic mapping of the generative model $p_{\boldsymbol{\theta}}(\boldsymbol{x}|\boldsymbol{z})$ to the atomistic  description as follows,
\begin{equation}
U^{\boldsymbol{x}^{(i)}}_{\text{bias}}(\boldsymbol{x}) \propto -\log
\int_{\mathcal M_{CV}}
p_{\boldsymbol{\theta}}(\boldsymbol{x}|\boldsymbol{z})
q_{\boldsymbol{\phi}}(\boldsymbol{z}|\boldsymbol{x}^{(i)})~d\boldsymbol{z}.
\label{eqn:bias_pot}
\end{equation}
Equation~(\ref{eqn:bias_pot}) is differentiable with respect to atomistic coordinates. Subtracting it from the atomistic potential 
could   accelerate the simulation by ``filling-in'' the deep free-energy wells.

% --------------------------------------------------------

\section*{Acknowledgements}

The authors acknowledge support  from the Defense Advanced Research Projects Agency (DARPA) under the Physics of Artificial Intelligence (PAI) program (contract HR$00111890034$). M.S. gratefully acknowledges the non-material and financial support of the Hanns-Seidel-Foundation, Germany funded by the German Federal Ministry of Education and Research. M.S. likewise acknowledges the support of NVIDIA Corporation.

\bibliography{ref}

\end{document}